\documentclass[10pt]{article}
\usepackage[preprint]{tmlr}
\usepackage{hyperref}
\usepackage{url}

\usepackage{array}
\usepackage{tabularx}
\usepackage{multirow}
\usepackage{float}
\usepackage{booktabs}

\usepackage{amsthm}
\usepackage{amsmath}
\usepackage{amsfonts}
\usepackage{graphicx}
\usepackage{arydshln}
\usepackage{algorithm}
\usepackage{algpseudocode}
\usepackage{amssymb}

\newcommand{\utility}{f}
\newcommand{\candidate}{y}
\newcommand{\configuration}{x}
\newcommand{\cost}{c}
\newcommand{\response}{r} 
\newcommand{\responseRND}{R} 

\newcommand{\BibTeX}{B\kern-.05em{\sc i\kern-.025em b}\kern-.08em\TeX}

\DeclareMathOperator*{\argmax}{arg\,max}

\newcommand{\rname}[1]{%
  \raisebox{-4.3ex}[0pt][0pt]{%
    \rotatebox[origin=lb]{90}{%
      \makebox[1.75cm][c]{\small #1}
    }%
  }%
}

\newcommand{\rotlabel}[1]{%
  \raisebox{0ex}[0pt][0pt]{%
    \rotatebox[origin=lb]{90}{%
      \makebox[2cm][c]{\small #1}
    }%
  }%
}

\title{Consecutive Preferential Bayesian Optimization}

\author{%
\name Aras Erarslan \email aras.erarslan@helsinki.fi \\
\addr University of Helsinki
\AND
\name Carlos Sevilla Salcedo \email casevill@ing.uc3m.es \\
\addr Universidad Carlos III de Madrid
\AND
\name Ville Tanskanen \email ville.tanskanen@helsinki.fi \\
\addr University of Helsinki
\AND
\name Anni Nisov \email anni.nisov@vtt.fi \\
\addr VTT Technical Research Centre of Finland
\AND
\name Eero P\"aiv\"akumpu \email eero.paivakumpu@vtt.fi \\
\addr VTT Technical Research Centre of Finland
\AND
\name Heikki Aisala \email heikki.aisala@vtt.fi \\
\addr VTT Technical Research Centre of Finland
\AND
\name Kaisu Honkap\"a\"a \email kaisu.honkapaa@vtt.fi \\
\addr VTT Technical Research Centre of Finland
\AND
\name Arto Klami \email arto.klami@helsinki.fi \\
\addr University of Helsinki
\AND
\name Petrus Mikkola \email petrus.mikkola@helsinki.fi \\
\addr University of Helsinki
}

\begin{document}

\maketitle

\begin{abstract}
Preferential Bayesian optimization allows optimization of objectives that are either expensive or difficult to measure directly, by relying on a minimal number of comparative evaluations done by a human expert. Generating candidate solutions for evaluation is also often expensive, but existing preferential optimization methods do not account for this production cost. We introduce Consecutive Preferential Bayesian Optimization (CPBO) for the consecutive setting, where production cost is reduced by reusing the previous candidate in each comparison. We also account for the perceptual ambiguity of the oracle providing the feedback by incorporating a \emph{Just-Noticeable Difference} threshold into a probabilistic preference model to capture indifference to small utility differences. We adapt an information-theoretic acquisition strategy to this setting, selecting new configurations that are most informative about the unknown optimum under a preference model accounting for the perceptual ambiguity. We empirically demonstrate a notable increase in accuracy in setups with high production costs or indifference feedback, and demonstrate the practical feasibility of CPBO in a real-world food extrusion task.
\end{abstract}

\section{Introduction}
\label{sec:introduction}

\emph{Preferential Bayesian optimization} (PBO) tells how to efficiently find the optimum of an unknown function $\utility(\candidate) \in \mathbb{R}$ when we cannot even evaluate the function itself but can only conduct \emph{preferential queries} \mbox{\citep{chu2005preference, gonzalez2017pbo}}. That is, we only have an oracle (often a human expert) for evaluating the relative quality of two candidates $\candidate_i$ and $\candidate_j$, which must be used to learn a proxy for $\utility(\candidate)$.
As a concrete example, in an algorithmic search for the best possible avocado toast recipe, we want to leverage preferential comparisons since providing absolute scores is difficult \citep{kochanski2017dessert}, and we want to minimize the number of preferential queries that all require eating a toast.

However, most real optimization tasks involve another cost that the PBO literature ignores: Making the toasts is not free either. Methods like the popular qEUBO \citep{astudillo2023qeubo} that query preferences over \emph{multiple} new candidates can be especially wasteful in this common setting.

\begin{figure*}[t]
  \centering
  \includegraphics[width=0.96\textwidth]{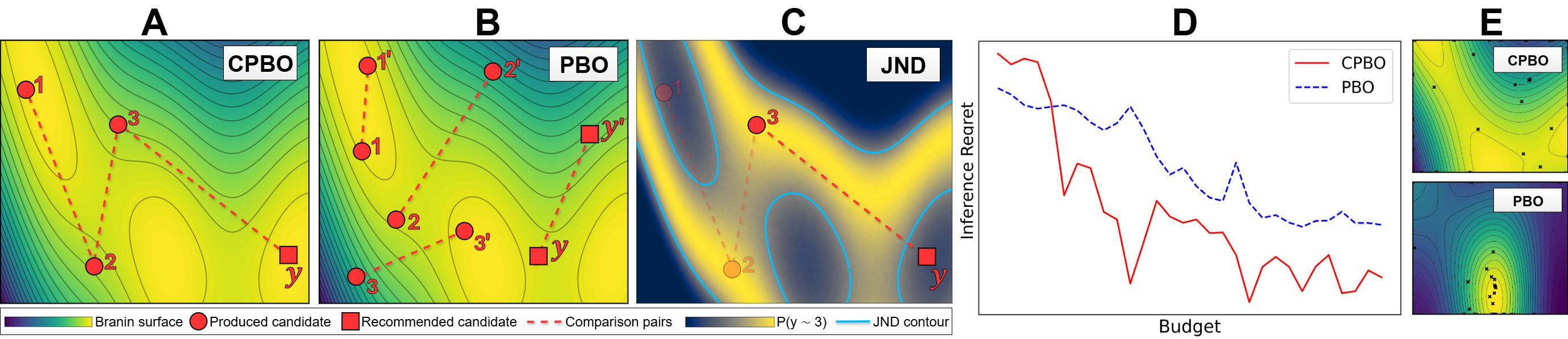}
  \caption{Consecutive Preferential BO (CPBO) operates with consecutive comparisons (A), in contrast to standard PBO where two (or more) new candidates are proposed at each iteration (B). For each comparison we model how likely the expert cannot tell the candidates apart; (C) shows this probability as a heatmap when candidate is compared to candidate “3”, with the contour marking the Just-Noticeable Difference (JND) threshold. Under non-zero production costs and JND, CPBO improves optimization performance for a fixed production/evaluation budget (D) and the quality of the learned utility proxy (E) over standard PBO, here EUBO by \citet{lin2022eubo}.
  } 
  \label{fig:teaser}
\end{figure*}

We introduce \emph{Consecutive Preferential Bayesian Optimization} (CPBO), a new problem formulation for addressing preferential optimization setups where the cost of producing the candidates is not negligible (Fig.~\ref{fig:teaser}A-B). Instead of freely proposing candidates for new preferential queries, we now need to consider as reference points some candidates that have already been produced. Whenever the cost of producing a candidate is sufficiently large, an optimal strategy is to consider \emph{only} preference queries of the type $\candidate_t \succ \candidate_l$, where $\candidate_l$ is a candidate solution produced during an earlier iteration $l<t$. We introduce the general setup, explain how the performance of current PBO methods changes when constrained to operate consecutively, even though they were not originally designed for it, and discuss how the relative costs of production and evaluation influence the optimal choices.

Within the general CPBO setup, we focus on a case with an additional constraint: Instead of all previously produced candidates, we can only compare $\candidate_t$ to $\candidate_{t-1}$, the previous one. This is needed when the candidates deteriorate over time: Avocados need to be consumed fresh, and in our concrete empirical experiment of high-moisture extrusion processing (Section~\ref{sec:experiment_extrusion}) the candidates need to be evaluated while they are still warm after production \citep{lawless2010sensory}. Conditions may also need to be experienced consecutively to provide a reliable assessment, as in thermal comfort optimization \citep{xu2024popbo}. We propose a concrete algorithm that extends Max-value Entropy Search (MES) \citep{wang2017MES} to select the optimal candidate to be produced and compared against the previous candidate, representing the first practical solution for automating optimization for this type of problem. Even though \citet{xu2024popbo} mentioned the setup anecdotally, they did so only to simplify theoretical analysis and did not consider any real consecutive optimization problems. 

Setups integrally relying on human evaluations warrant more careful consideration of the evaluation aspect.
The PBO literature is primarily motivated by tasks where the oracle is a human expert -- for example, in animation design \citep{brochu2010bayesian} or in molecular configuration \citep{mikkola2020ppbo} -- but most works model humans as perfect oracles, assuming they have definite answers to all queries: either $\candidate_i \succ \candidate_j$ or $\candidate_i \prec \candidate_j$. That is, the evaluator is forced to prefer one toast over the other, even when they regard them as equally good or when there are conflicting signals (\emph{``I prefer the spices in this one, but the toast is less crusty."}). This oversimplification has been questioned in other learning setups, with demonstrated improvement from accounting for the indifference in human preferences \citep{liu2024reward}.

We address this shortcoming using the concept of \emph{Just-Noticeable Difference} (JND) \citep{weber1978sense}. In addition to $\candidate_i \succ \candidate_j$ or $\candidate_i \prec \candidate_j$, we also consider $\candidate_i \sim \candidate_j$, where $\sim$ is the commonly used symbol for \emph{indifference} to the preference query. Indifference-aware preference modeling has been considered previously in the context of standard preference-based BO. \citet{nguyen2021top} introduce a learnable indifference threshold within the Plackett--Luce model \citep{plackett1975analysis,luce1959individual} and propose an entropy-based acquisition strategy for top-$k$ ranking problems, which reduces to the Bradley--Terry model \citep{bradley1952rank} in the case of pairwise comparisons. Their setting assumes that arbitrary candidate pairs or subsets can be freely selected at each iteration. In contrast, we focus on a cost-constrained consecutive setting, where only one new candidate is generated per iteration and comparisons must be made against previously produced references, fundamentally changing the acquisition problem and precluding a direct application of their method. We model such ambiguity by extending the Thurstone--Mosteller model \citep{thurstone1927judgment,mosteller1951remarks} with a learnable indifference band in a latent space underlying the assumed utility. We empirically compare the proposed CPBO method against concurrent PBO methods \citep{lin2022eubo, xu2024popbo} in scenarios of varying degrees of indifference, showing that already for cases where $10-20\%$ of comparisons are indicated as indifferent, our method clearly outperforms them (Fig.~\ref{fig:teaser}C-E). Furthermore, having some indifferences actually helps, compared to assuming perfect answers for all queries; Fig.~\ref{fig:teaser}C shows how the JND helps avoid querying candidates with similar expected utility.
In this sense, the consecutive setting and JND are not independent ingredients: the consecutive constraint fixes one side of the comparison, which makes the JND threshold directly actionable when selecting the next candidate. In particular, the acquisition rule can avoid comparisons whose expected utility difference is likely below the JND, while also avoiding overly obvious comparisons that provide little information. Thus, CPBO provides a natural setting in which perceptual ambiguity can influence the optimization loop itself, rather than only being handled retrospectively in the preference model.

We make the following concrete contributions to the PBO literature.
\textbf{Extended cost-awareness:} We explain how PBO needs to account also for the cost of producing the candidates, not just the evaluation, and characterize how the balance influences the choice of the algorithm.
\textbf{Consecutive PBO:} We characterize a new learning setup where comparisons are constrained to previously produced candidates,
designed for scenarios of high production cost and deteriorating candidates.
\textbf{Practical Method:} We extend the Thurstone model to account for JND and adapt MES for conditional comparisons, arriving at a concrete CPBO method for real optimization tasks.
\textbf{Experimental Validation:} 
Quantification of the importance of accounting for indifference evaluations, and a real-world laboratory demonstration in food extrusion optimization using preference data physically collected under real operational constraints.

Source code is available at \href{https://github.com/aerarslan/CPBO}{https://github.com/aerarslan/CPBO}.


\section{Problem Setup}
\label{sec:problem_setup}

\paragraph{Cost-Aware Preferential BO.}
Analogously to the setup presented by \citet{lin2022eubo}, proper treatment of any PBO setup requires three quantities:
\begin{itemize}
   \renewcommand\labelitemi{}
  \item $\configuration$: A configuration to be evaluated (recipe for a toast; settings of an extruder)
  \item $\candidate(\configuration)$: The candidate itself, produced with configuration $\configuration$ (an actual toast; extrudate)
  \item $\utility(\candidate)$: The utility of the user preference, so that 
  $\utility(\candidate) > \utility(\candidate')$, denoted by $\candidate \succ \candidate'$, 
  implies the user prefers candidate $\candidate$ over $\candidate'$ (taste; subjective evaluation of overall quality)
\end{itemize}
Our goal is to find the configuration $\configuration$ that maximizes the utility for the corresponding candidate $\candidate$ as
$
    \argmax_{x} \utility(\candidate(\configuration)),
$
by selecting a sequence of configurations $\configuration_1,..,\configuration_T$ while respecting a budget $B>0$ such that $\sum \cost_p + \sum \cost_e \leq B$. Here, each configuration incurs the \emph{production cost} $\cost_p \geq 0$ to produce $y(x)$, and each pairwise comparison $(x,x')$ incurs the \emph{evaluation cost} $\cost_e \geq 0$ to evaluate $\mathbb{I}(y(x) \succ y(x'))$. The sums are over \textit{unique} productions and judgments: the production cost incurs only once per unique candidate, not when using the candidate again in a later comparison. 

The existing PBO literature does not explicitly distinguish between the production cost $\cost_p$ and the evaluation cost $\cost_e$, lacking tools to account for them. Our formulation makes three separate scenarios apparent:
\begin{enumerate}
  \item For $\cost_p \ll \cost_e$, we can ignore the production cost, as implicitly done by previous PBO methods tracking progress in terms of preference evaluations \citep{gonzalez2017pbo}. qEUBO \citep{astudillo2023qeubo} takes this to the extreme, proposing $q \ge 2$ new candidates for a single evaluation.
  \item For $\cost_p \approx \cost_e$, we should typically produce only one new candidate for each comparison, selecting the other candidate from those already produced to avoid a new production cost.
  \item For $\cost_p \gg \cost_e$, each newly produced candidate should be compared against \emph{all} previous ones. For instance, in the duel bandit context, \citet{agarwal2022duel} adds every new candidate to a candidate set and uses all pairwise comparisons.
\end{enumerate}

\paragraph{Consecutive Preferential BO.}
The first case is well covered by current methods and addressing the third one requires only a minor extension using multiple comparisons. Our main focus is in the unexplored middle case. We leave the general case of cost-aware preferential BO, in which the costs $\cost_p$ and $\cost_e$ enter the acquisition policy explicitly, for future work, but see \citep{swersky2013multitask, lee2020costawarebo, luong2021costawarebo, garnett_bayesoptbook_2023} for the standard formulation of cost-aware BO. Instead, we study how to select an optimal $\configuration_j$ conditional on an already fixed $\configuration_i$.

To focus on this novel aspect, the concrete CPBO algorithm assumes that each iteration produces one new candidate and performs one comparison, so every candidate considered by the acquisition function has the same cost $\cost_p+\cost_e$. Normalizing the acquisition by this constant cost would therefore not change the selected configuration. We further restrict the reference point to the immediately previous candidate $\candidate_{t-1}$. In scenarios where the candidates deteriorate over time, it is the only candidate that can be assessed without incurring yet a third possible cost, the \emph{retrieval cost} $\cost_r$ for storing (e.g., freezing) and retrieving (de-frosting) historical candidates. This special case is a direct answer to a concrete need, as demonstrated in Section~\ref{sec:experiment_extrusion}.

The formulation can be extended to more general cases, but we do not evaluate such extensions in this work. For instance, if $L$ previous candidates can be assessed without the retrieval cost, we can compute the information gains for all $L$ choices in parallel. More generally, a cost-aware policy could choose between a standard query involving two newly produced candidates, with cost $2\cost_p+\cost_e$, and a conditional query involving one new candidate and one previously produced reference, with cost $\cost_p+\cost_e$, by comparing their expected information gain per unit cost.


\section{Method}
\label{sec:method}

To solve the CPBO problem, we propose a preferential BO algorithm with three key components. First, it uses consecutive comparisons as defined above. Second, it models preference feedback using a Random Utility Model (RUM) \citep{mcfadden1974rum, manski1977structure} with a JND threshold to capture indifference. Third, it selects the next candidate $\candidate$ by maximizing the mutual information between the preference response and the unknown maximum of the latent utility proxy $\utility(\configuration)$, using an adapted variant of MES. As a proxy, we make the common choice of a Gaussian Process (GP) \citep{williams2006gaussian}, with analytic expressions for many required quantities. 

We first specify the observation model conditional on latent preferences, followed by description of how we model the latent preferences using GPs. Finally, we introduce the acquisition function and its computational details. Additional details complementing the presentation and a high-level pseudo-code are provided in Appendix~\ref{app:method_details}.

\subsection{Observation Model}
\label{sec:observation_model}

We model user feedback using a RUM, where preferences arise from latent utility values corrupted by perceptual noise. At each iteration, a new candidate $\candidate_t = \candidate(\configuration_t)$ is produced with a configuration $\configuration_t \in \mathcal{X}$. The expert compares this to the previous candidate $\candidate_{t-1}$ and provides a discrete preference label
$\responseRND_t \in \mathcal{R} = \{-1, 0, +1\}$,
indicating the preferred candidate or indifference. We denote its observed realization by $\response_t$.

After $t$ iterations, the preference data is
$ 
\mathcal{D}_t = \left\{ \left( \left[ \configuration_{i-1}, \configuration_i \right], \response_i \right) \right\}_{i=2}^{t}
$ 
with $\candidate_i = \candidate(\configuration_i)$. Although preferences are given for the physical outputs $\candidate(\configuration)$, we define the latent utility function $\utility(\configuration)$ to be the utility of $\candidate(\configuration)$, treating the production as deterministic. This lets us learn $\utility(\configuration)$ over $\mathcal{X}$ without an explicit model of $\candidate$. The current formulation therefore does not model stochastic variation in production. We also do not separately model systematic presentation-order or carry-over effects in the preference responses. Such effects would require extending the observation model.

Specifically, we form our preference likelihood by extending the classical Thurstone--Mosteller model \citep{thurstone1927judgment, mosteller1951remarks} to a three-outcome setting with an explicit indifference region.
Indifference-aware extensions of random utility models have been studied in the preference learning literature \citep{davidson1970extension, zhou2008learning}, and \citet{benavoli2024tutorial} also discusses JND in GP preference learning.
Our use of JND differs in that we combine the threshold with perceptual noise to obtain smooth response probabilities for the three possible outcomes, and use these probabilities inside the acquisition function for consecutive Bayesian optimization, introducing modifications required for stable inference and effective acquisition.

We assume that the expert implicitly evaluates a latent utility difference
\begin{equation}
\Delta \utility_t = \utility(\configuration_t) - \utility(\configuration_{t-1}),
\label{eq:utility_difference}
\end{equation}
but perceives only noisy utilities. Following the Thurstone-Mosteller model, we assume additive Gaussian noise on each candidate,
$
\varepsilon_t, \varepsilon_{t-1} \sim \mathcal{N}(0, \sigma^2),
$
making the perceived utility difference
\begin{equation}
\Delta \utility_t + \delta_t, \quad \text{where} \quad \delta_t = \varepsilon_t - \varepsilon_{t-1} \sim \mathcal{N}(0, 2\sigma^2).
\label{eq:noisy_utility_diff}
\end{equation}

The user's response compares the noisy utility difference within the JND threshold $\gamma > 0$
\begin{equation}
\label{eq:preference_model}
\responseRND_t =
\begin{cases}
+1, & \text{if } \; \Delta \utility_t + \delta_t > \gamma, \\
\;\,0, & \text{if } \; |\Delta \utility_t + \delta_t| \le \gamma, \\
-1, & \text{if } \; \Delta \utility_t + \delta_t < -\gamma.
\end{cases}
\end{equation}
This threshold $\gamma$ models perceptual ambiguity, mapping small utility differences more likely to indifference. For $\gamma = 0$, it reduces to standard binary preferences \citep{chu2005preference}.

Given the latent utility function $\utility$, the likelihood of a single preference observation can be written as
\begin{equation}
\begin{array}{l}
P(\responseRND_t = +1 \mid \utility) =
  \Phi\!\left( \tfrac{\Delta \utility_t - \gamma}{\sqrt{2}\sigma} \right), \\[6pt]
P(\responseRND_t = 0 \mid \utility) =
  \Phi\!\left( \tfrac{\gamma - \Delta \utility_t}{\sqrt{2}\sigma} \right)
  - \Phi\!\left( \tfrac{-\gamma - \Delta \utility_t}{\sqrt{2}\sigma} \right), \\[6pt]
P(\responseRND_t = -1 \mid \utility) =
  \Phi\!\left( \tfrac{-\Delta \utility_t - \gamma}{\sqrt{2}\sigma} \right),
\end{array}
\label{eq:three_outcome_likelihood}
\end{equation}
where $\Phi$ denotes the standard normal CDF.
The joint likelihood of the data is hence
\begin{equation}
P(\mathcal{D}_t \mid \utility) = \prod_{i=2}^{t} P(\response_i \mid \utility).
\label{eq:dataset_likelihood}
\end{equation}

\subsection{Latent Utility Model and Posterior Inference}
\label{sec:latent_utility_model}

The above observation model was defined conditional on the latent utility function $\utility$.
To model uncertainty over the latent utility, we place a GP prior
$
\utility \sim \mathcal{GP}(\mu, k),
$
where the prior mean is constant, $\mu(\configuration)=m$, and the covariance is a radial basis function (RBF) kernel with automatic relevance determination (ARD), where each $\ell_d$ is a separate lengthscale for input dimension $d$:
\begin{equation}
k(\configuration,\configuration')
= \sigma_f^2 \exp\Bigl(-\tfrac12\sum_{d=1}^D
\frac{(\configuration_d-\configuration'_d)^2}{\ell_d^2}\Bigr).
\label{eq:gp_kernel}
\end{equation}
The GP output scale $\sigma_f^2$ is fixed to remove the scale ambiguity discussed below. We follow the standard preferential GP inference framework \citep{chu2005preference}, but replace Laplace approximation with variational inference. Combined with the observation likelihood in Eq.~\eqref{eq:dataset_likelihood}, the posterior is
$
p(\utility \mid \mathcal{D}_t)
\propto
P(\mathcal{D}_t \mid \utility)p(\utility).
$

We use variational approximation to estimate the posterior $p(\utility | \mathcal{D}_t)$ \citep{titsias2009variational}, implemented with the \texttt{GPyTorch} library \citep{gardner2018gpytorch}. At iteration $t$, let
$
\mathbf{u} = [\utility(\configuration_1), \dots, \utility(\configuration_t)]^\top
$
denote the latent utility values at all configurations observed so far. We define a multivariate Gaussian variational distribution
$
q(\mathbf{u}) = \mathcal{N}(\mathbf{m}_u, \mathbf{L}\mathbf{L}^{\top}),
$
where $\mathbf{m}_u \in \mathbb{R}^t$ and $\mathbf{L} \in \mathbb{R}^{t \times t}$ is a Cholesky factor. The approximate posterior over the latent utility function is then obtained by integrating the GP conditional,
\begin{equation}
q\bigl(\utility(\cdot)\bigr)
= \int p\bigl(\utility(\cdot)\mid \mathbf{u}\bigr) q(\mathbf{u})\,d\mathbf{u}.
\label{eq:variational_posterior}
\end{equation}
We jointly optimize the variational parameters, the constant prior mean, the ARD lengthscales, and the JND threshold $\gamma$ by maximizing the Evidence Lower Bound (ELBO) as an approximate marginal likelihood. The expected log-likelihood term in the ELBO is estimated by Monte Carlo samples of latent utilities drawn from $q(\utility)$, and the resulting objective is optimized with gradient-based methods.

Due to the latent structure of the model, where feedback depends only on utility differences, the solution is not identifiable. The likelihood is invariant under joint rescaling of the latent function $\utility(\configuration)$ and the perceptual noise $\sigma$; $\Delta \utility / \sigma$ and $\gamma / \sigma$ are invariant to scaling. To remove this invariance, we fix $\sigma$ and the GP output scale to constant reference values. The sensitivity analysis of the perceptual-noise scale $\sigma$ in Appendix~\ref{cpbo_benchmarks} empirically validates this choice, showing that optimization performance remains similar across all seven benchmark functions when $\sigma$ is varied from $0.25\times$ to $4\times$ its default value. Further details on the model and computation, including numerical stabilization and optimization settings, are provided in Appendix~\ref{app:gp_model_details}.

\subsection{Acquisition Function}
\label{sec:acquisition}

We adapt Max-value Entropy Search (MES)~\citep{wang2017MES} to the CPBO setting, where each new candidate $\candidate(\configuration)$ is compared against the previous one $\candidate_t$. This adaptation differs from standard MES in two key respects. First, the consecutive constraint fixes one side of each comparison to the previous candidate, so the information provided by a proposed configuration depends on its relation to this reference. Second, the observation is a three-outcome preference response rather than a direct function evaluation, so the acquisition is defined using the JND-aware probabilities that the new candidate is preferred, the candidates are judged indifferent, or the previous candidate is preferred. The goal is to select the next configuration $\configuration$ that maximizes the mutual information with the unknown maximum value
\begin{equation}
\utility^* = \max_{\configuration \in \mathcal{X}} \utility(\configuration),
\label{eq:argmax_configuration}
\end{equation}
based on the expert's next preference feedback $\responseRND_{t+1}$. We define the acquisition function as
\begin{equation}
\begin{aligned}
\alpha(\configuration) &= 
  I(\utility^*; \responseRND_{t+1} \mid \mathcal{D}_t) \\[6pt]
&= H(\responseRND_{t+1} \mid \mathcal{D}_t) 
   - \mathbb{E}_{\utility^*} \!\left[
        H(\responseRND_{t+1} \mid \utility^*, \mathcal{D}_t)
      \right],
\end{aligned}
\label{eq:acquisition_info_gain}
\end{equation}
where the expectation is over $p(\utility^* \mid \mathcal{D}_t)$ and $H$ denotes entropy. In our setting, this entropy is over the three possible responses $\responseRND_{t+1}\in\{-1,0,+1\}$, with response probabilities computed from the JND likelihood in Eq.~\eqref{eq:three_outcome_likelihood}. Direct evaluation of Eq.~\eqref{eq:acquisition_info_gain} is not available in closed form, as it requires integrating over the posterior distribution of $\utility^*$ and evaluating the response distribution conditional on possible values of $\utility^*$. We therefore approximate these terms as described in the following subsection.

\subsection{Approximating the Acquisition Function}
\label{sec:acquisition_approximation}

Following the original MES \citep{wang2017MES}, we discretize $\mathcal{X}$ into a candidate set $\hat{\mathcal{X}}$ and approximate the maximum utility $\utility^* = \max_{\configuration \in \hat{\mathcal{X}}} \utility(\configuration)$ by fitting a Gumbel distribution to quantiles estimated from the GP posterior. The unconditional entropy term $H(\responseRND_{t+1}\mid\mathcal{D}_t)$ is estimated using Monte Carlo samples from the GP posterior, with predictive response probabilities computed using Eq.~\eqref{eq:three_outcome_likelihood}. For each $\utility^*_j \sim \text{Gumbel}$, we construct a truncated GP posterior $p_{\leq \utility^*_j}$ conditioned on $\mathcal{D}_t$, where the truncation enforces $\max\{\utility(\configuration_t), \utility(\configuration)\} \leq \utility^*_j$.

The conditional feedback distribution is then estimated as
\begin{equation}
P(\responseRND_{t+1} = r \mid \utility^*_j, \mathcal{D}_t) 
= \mathbb{E}_{\utility \sim p_{\leq \utility^*_j}} 
\left[ P(r \mid \Delta\utility) \right],
\label{eq:truncated_preference}
\end{equation}
where $\Delta\utility = \utility(\configuration) - \utility(\configuration_t)$ and $P(r \mid \Delta\utility)$ is given by Eq.~\eqref{eq:three_outcome_likelihood}. Acquisition values are computed via Monte Carlo integration over Gumbel samples. In practice, directly evaluating the expectation over $\utility^*$ would require constructing a separate truncated posterior for many sampled maximum values. To reduce this cost, we sample maximum values from the fitted Gumbel
distribution, group them into equally spaced bins, and use the midpoint of
each bin as a representative truncation level. The contribution of each
representative is weighted by the fraction of sampled maximum values falling
in its bin. This reduces the number of truncated posterior computations while retaining a Monte Carlo approximation of the expectation over $\utility^*$.

The acquisition function can be optimized in several ways. To solve the stochastic optimization problem caused by the Monte Carlo estimate of the conditional entropy, we use a gradient-free optimizer, specifically Bayesian optimization for this inner acquisition function optimization loop. Although one could employ a gradient-based optimizer utilizing automatic differentiation and the reparameterization trick (e.g., as in the BoTorch framework \cite{balandat2020botorch}), such methods frequently necessitate multi-start gradient optimization. Using Bayesian optimization as a global optimizer avoids this issue.

For numerical stability, we apply zero-mean normalization before computing entropy terms: for each posterior utility sample $\utility^{(s)}$ and evaluation batch $\mathcal{X}_{\mathrm{eval}}=\{\configuration^{(1)},\ldots,\configuration^{(N)}\}$, we subtract the empirical mean $N^{-1}\sum_{j=1}^{N}\utility^{(s)}(\configuration^{(j)})$ from every value of $\utility^{(s)}$ in that batch. This does not change the preference probabilities, since the likelihood in Eq.~\eqref{eq:three_outcome_likelihood} depends only on utility differences.

From a technical perspective, the approximation closely follows standard MES. The main additional element required by the consecutive preference setting is local dual truncation, where the GP posterior is conditioned jointly on the proposed configuration $\configuration$ and the fixed reference $\configuration_t$ using the sampled maximum $\utility^*_j$.

Appendix~\ref{app:acquisition_details} provides additional details, including the Gumbel fitting equations and the use of the posterior median for centering the fitted distribution, as well as binning parameters, zero-mean normalization of posterior samples, and numerical stabilization.


\section{Experiments}
\label{sec:experiments}

We empirically verify three main claims. First, we show how the balance between the production and evaluation costs influences the general pattern of PBO, using the EUBO method \citep{lin2022eubo} in different ways. We perform this experiment with an established standard PBO method, rather than our proposal, to draw attention to the general findings. Second, we show strong overall performance of our method in the CPBO setup in the presence of JND, with detailed comparison against the baselines. Third, we show that CPBO outperforms baselines accounting for JND only in construction of the latent utility proxy, confirming the need for the proposed acquisition design.
{In addition to the quantitative evaluations, we demonstrate how the method can be applied in real use cases, within a recurring optimization task in food science where consecutive comparisons are the only information source.}

Across all synthetic benchmarks, we use a fixed algorithmic configuration without manual hyperparameter tuning. Within each benchmark and run, all compared methods use the same random seed and start from the same initial configurations, while different runs use different seeds, enabling paired statistical comparisons. GP hyperparameters are learned during the algorithm or fixed to reference values, and the only adjustable setting is the acquisition optimization budget. The same configuration is used in the extrusion demonstration, with a weak prior on GP lengthscales for stabilization (see Appendix~\ref{app:vtt_technical_details}).

\paragraph{Baselines.} We compare against two recent PBO methods: EUBO by \citet{lin2022eubo} as implemented in the \texttt{BoTorch} library \citep{balandat2020botorch}, and POP-BO by \citet{xu2024popbo} as implemented in the authors' code release. For EUBO, we adapt the query strategy by forcing the first alternative to be the candidate proposed in the previous query, and optimizing only over the second alternative. 
Note that EUBO is similar but not identical to qEUBO with $q=2$ \citep{astudillo2023qeubo}; in Appendix~\ref{cpbo_benchmarks} we show additional results with qEUBO.

POP-BO natively supports consecutive comparisons but requires direct evaluations of $\utility(\candidate)$ to determine various hyperparameters and, hence, it can not be used in pure PBO tasks. To side-step this limitation, we re-use the hyperparameter choices \citet{xu2024popbo} provided using such evaluations, and exclude POP-BO for functions they did not consider (NAs in Table~\ref{tab:combined_results}). Nonetheless, we emphasize that this gives POP-BO an unfair advantage over both CPBO and EUBO.

For evaluating the importance of accounting for JND also in the acquisition function, we construct two additional baselines that combine our observation model with UCB acquisitions. We defer their description to Section~\ref{sec:jnd_acquisition_ablation} since they are only considered in that experiment.

\paragraph{Metrics.} We use the standard BO evaluation metrics: \emph{inference regret} $\utility(\configuration^*) - \utility(\hat{\configuration}_n)$, where $\hat{\configuration}_n = \arg\max_{\configuration \in \mathcal{X}} \mu_n(\configuration)$ is the model's final recommendation, and \emph{simple regret}, which measures the quality of the best solution thus far as $\utility(\configuration^*) - \max_{1 \leq i \leq n} \utility(\configuration_i)$. Both regret values require access to the true utility and are therefore available only as evaluation metrics on synthetic benchmarks. However, the recommendation $\hat{\configuration}_n$ used by inference regret is directly available from the learned model, whereas identifying the best sampled candidate for simple regret requires access to the true utilities of the sampled configurations. For this reason, inference regret is our primary metric for evaluating the final deployable recommendation.
To assess the level of exploration, we measure the global quality of the utility proxy $\utility(\configuration)$ after optimization using \emph{ordinal accuracy}, which calculates binary ranking consistency over uniformly random configuration pairs by ignoring the indifference threshold.

\paragraph{Data.}
We use seven standard optimization benchmarks from \citep{surjanovic2013virtual}, grouped into two sets. The first set consists of Branin, Six-Hump, and Bohachevsky, which have non-spiky landscapes and are more likely to be representative of utility functions expected in human studies. The second set consists of Levy13, Bukin6, Cross-Tray, and Ackley, which have spiky landscapes designed as challenging targets for standard optimization problems, used here to demonstrate the robustness and generality of the method. Additional results and full details are provided in the Appendix~\ref{app:experiment_details}. 

The data for the real-world case study is explained in 
Section~\ref{sec:experiment_extrusion}.

\begin{figure*}[t]
    \centering
    \includegraphics[width=1.0\textwidth]{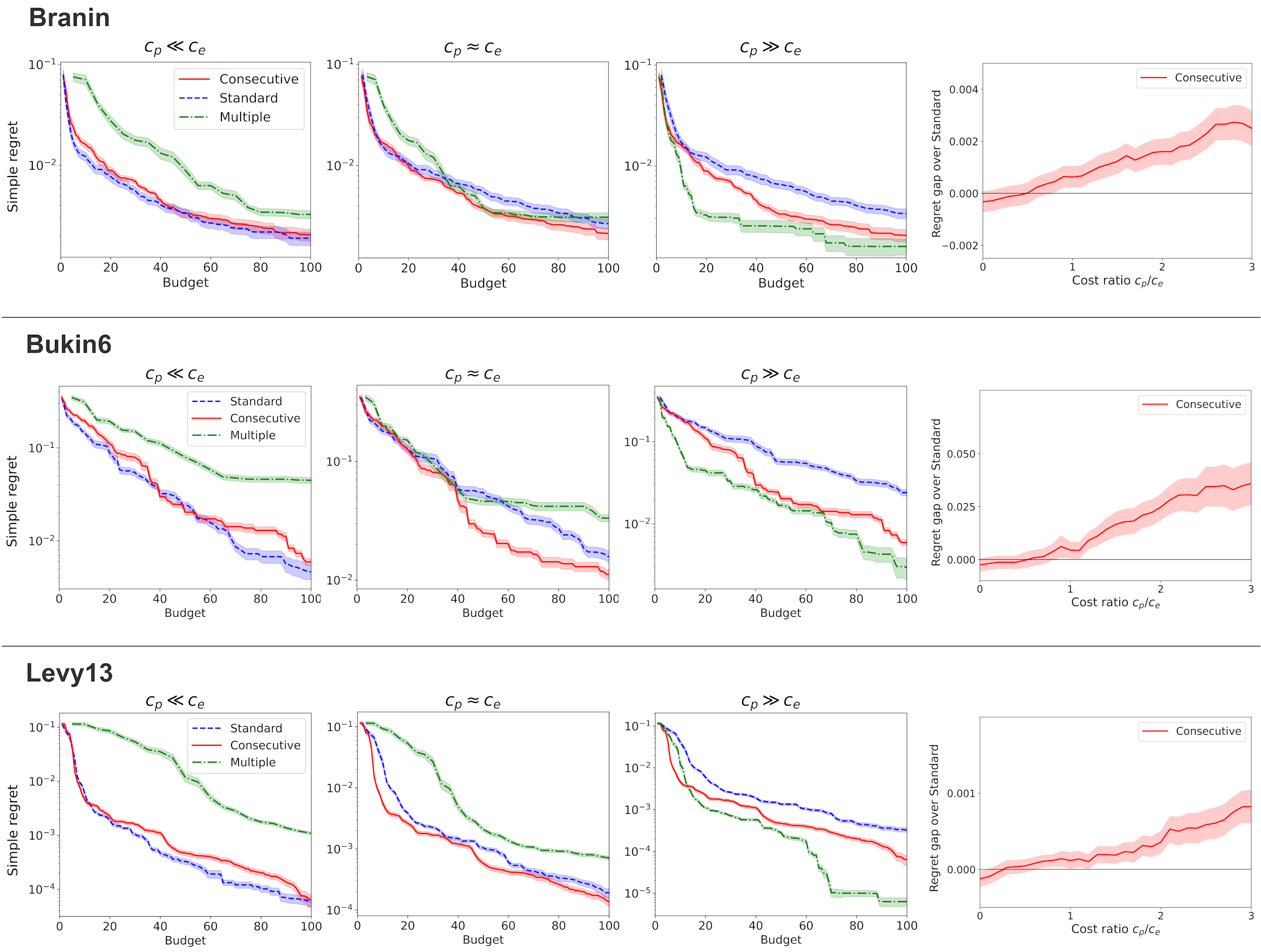}
    \caption{Convergence of PBO variants on three synthetic benchmark functions (Branin, Bukin6, and Levy13) across three cost-balance scenarios.
    \textbf{Left three columns:} Simple regret as a function of the total budget $B$ when $\cost_p \ll \cost_e$, $\cost_p \approx \cost_e$, and $\cost_p \gg \cost_e$.
    \textbf{Right column:} Difference in simple regret between Standard and Consecutive at fixed budget $B=100$ as the cost ratio $\cost_p/\cost_e$ varies. Values above zero indicate that Consecutive achieves lower regret than Standard.
    Solid lines show the mean and shaded regions show the standard error of the mean over 30 runs.
    }
    \label{fig:branin_cost_analysis}
\end{figure*}

\subsection{When Is CPBO Needed?}
\label{sec:experiment_cost_aware}

Standard PBO works track progress as a function of the number of human evaluations, which is most appropriate when $\cost_p=0$. To empirically study how $\cost_p>0$ affects PBO performance,  we evaluate three alternative variants of EUBO:

\begin{enumerate}
    \item {\bf Standard}: Every iteration proposes a pair $(\candidate,\candidate')$ to be compared, with cost of $2\cost_p + \cost_e$.
    \item {\bf Consecutive}: Every iteration proposes one new candidate $y_t$, to be compared against the previous $y_{t-1}$, with cost of $\cost_p + \cost_e$.
    \item {\bf Multiple}: Every iteration proposes one new candidate as in the consecutive setting, which is compared against $L$ previous candidates, with cost of $\cost_p + L\cost_e$. We use $L=5$ for illustration purposes.
    
\end{enumerate}
Fig.~\ref{fig:branin_cost_analysis} shows the simple regret as a function of the total budget $B$ for three benchmark functions and three cost-balance scenarios: $\cost_p \ll \cost_e$, $\cost_p \approx \cost_e$, and $\cost_p \gg \cost_e$; see Appendix~\ref{additional_cost_results} for details and results for other target functions and inference regret.
The results exactly match the expectations:
\begin{enumerate}
    \item When only evaluations are costly, standard PBO performs best. However, the penalty for limiting to consecutive comparisons is small, with confidence intervals overlapping in some cases.
    \item When $\cost_p \approx \cost_e$, the consecutive comparison strategy is consistently the best.
    \item When production cost dominates, we observe a clear benefit from using more comparisons. Standard PBO is clearly the worst.
\end{enumerate}
That is, for extreme cases, standard PBO or a simple extension that uses all possible evaluations is the most effective option, but for balanced scenarios, we need consecutive comparisons.
Fig.~\ref{fig:branin_cost_analysis} (right most panels) quantifies the improvement from using consecutive evaluations, with roughly linear growth as a function of the cost ratio $c_p/c_e$.

\subsection{CPBO with Just-Noticeable Differences}
\label{sec:experiment_jnd}

This section studies the overall performance of PBO methods in the CPBO setup and the importance of accounting for JND. We benchmark the methods on standard BO targets by simulating oracle responses using our user model (Eq.~\ref{eq:preference_model}) with controlled JND threshold $\gamma_{\text{true}}$. To better understand the degree of indifference, we also report the expected ratio of indifferences over random comparison pairs for given $\gamma_{\text{true}}$ for each target, denoted by $\mathbb{E}[\sim]$.

Table~\ref{tab:combined_results} reports the performance of the methods for $\gamma_{\text{true}}=0.04$, corresponding to $10-35\%$ expected fraction of indifferences depending on the function. To ablate the importance of accounting for indifferences, we separately report the results for CPBO with the JND threshold fixed to zero ($\text{CPBO}_{\hat \gamma = 0}$) and with a trainable threshold ($\text{CPBO}_{\hat \gamma}$). The former includes all other elements of our method but specifically avoids accounting for the JND.

CPBO$_{\hat{\gamma}}$ achieves the lowest mean inference regret and the highest mean ordinal accuracy on all seven benchmark functions in Table~\ref{tab:combined_results}. For both metrics, the differences to all available comparison methods are statistically significant on every benchmark except Cross-Tray (benchmark-level intersection-union tests based on paired two-sided Wilcoxon signed-rank tests, $p_{\mathrm{IUT}}<0.05$). Paired mean differences, rank-biserial effect sizes, 95\% bootstrap confidence intervals, and statistical test results are reported in Appendix~\ref{app:statistical_analysis}.

\begin{figure*}[t]
  \centering
  \includegraphics[width=0.85\linewidth]{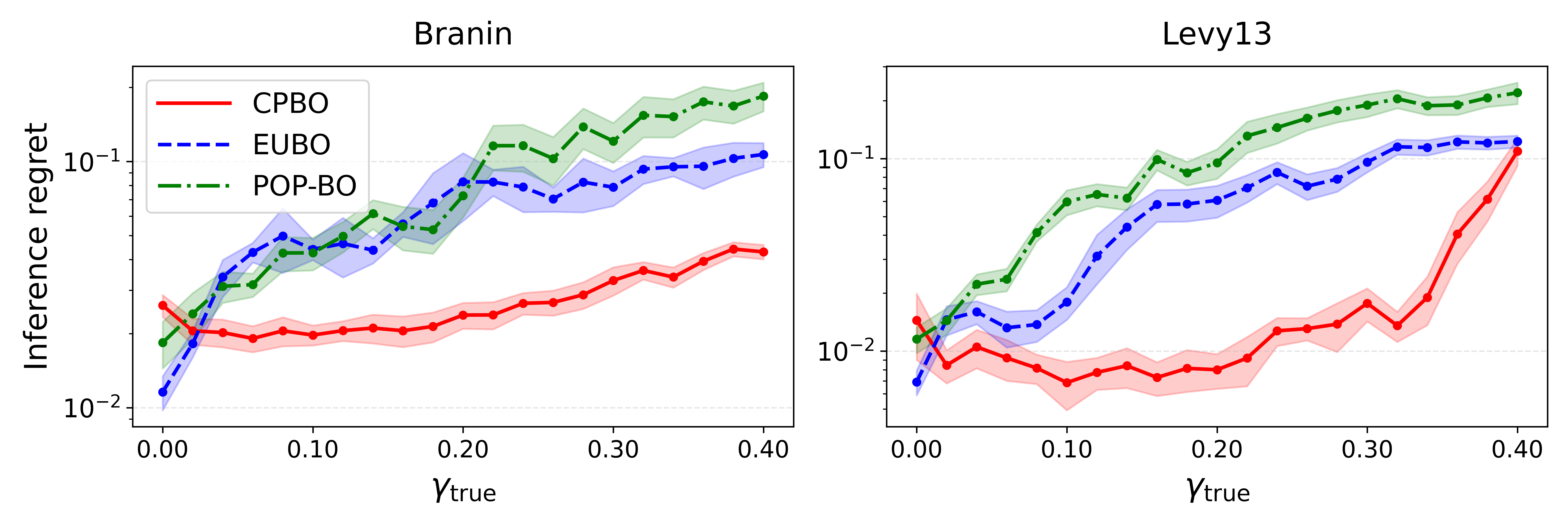}
  \caption{%
    Inference regret on Branin and Levy13 at iteration 30 for varying degrees of JND~$\gamma_{\text{true}}$. Solid line shows the mean and shaded regions the standard error of the mean over 30 runs.%
  }
  \label{fig:varying_gamma_branin_levy}
\end{figure*}

An interesting question is how $\gamma_{\text{true}}$, the true degree of indifference of the oracle, influences the performance. 
Fig.~\ref{fig:varying_gamma_branin_levy} shows that when all preference comparisons are definite, the state-of-the-art EUBO is the best in terms of regret, and CPBO behaves like a standard explorative method under the binary preference model. However, already around $\gamma_{\text{true}} \approx 0.02$, when $\mathbb{E}[\sim] \approx 11\%$ for Branin and $\mathbb{E}[\sim] \approx 9\%$ for Levy13, modeling the perceptual ambiguity becomes critical. For all higher thresholds, CPBO is consistently the best, only losing its advantage when the task becomes effectively impossible (for Levy13 with $\gamma_{\text{true}}=0.4$, we have $r_t=0$ for $94\%$ of comparisons). 

For a broad range of $\gamma_{\text{true}}>0$, CPBO achieves lower mean inference regret than for $\gamma_{\text{true}}=0$. This seemingly counter-intuitive observation has a natural explanation. The information gain Eq.~\eqref{eq:acquisition_info_gain} tends to zero for both $\Delta f_t \ll \gamma$ and $\Delta f_t \gg \gamma$ and hence the algorithm prefers candidates with $\Delta f_t \approx \gamma$. This helps prune out large parts of irrelevant configuration space and assists in proposing high-utility areas, as illustrated in Fig.~\ref{fig:teaser}C and other examples in the Appendix~\ref{cpbo_benchmarks}.

\begin{table*}[t]
\centering
\caption{Optimization and exploration performance measured by inference regret (Regret) and global ordinal accuracy of the utility proxy (Ord.) after 30 iterations, with JND of $\gamma_{\text{true}} = 0.04$. Inference regret is reported as mean $\pm$ sd over 30 runs, and ordinal accuracy as the mean over the same runs. Bold indicates the best mean performance for each benchmark and metric; corresponding statistical comparisons are reported in Appendix Tables~\ref{tab:statistical_regret} and~\ref{tab:statistical_ordinal}. The three functions above the dashed line best represent typical utility functions.}
\vspace{0.5em}
\setlength{\tabcolsep}{5.7pt}
\renewcommand{\arraystretch}{1.1}
\begin{tabular}{lccccccccc}
\toprule
\multirow{2}{*}{$\utility(\candidate)$} 
  & \multirow{2}{*}{$\mathbb{E}[\sim]$} 
  & \multicolumn{2}{c}{\textbf{POP-BO}} 
  & \multicolumn{2}{c}{\textbf{EUBO}} 
  & \multicolumn{2}{c}{\textbf{CPBO$_{\hat \gamma=0}$}} 
  & \multicolumn{2}{c}{\textbf{CPBO$_{\hat \gamma}$}} \\
\cmidrule(lr){3-4} \cmidrule(lr){5-6} \cmidrule(lr){7-8} \cmidrule(lr){9-10}
 &  & Regret\,$\downarrow$ & Ord.$\uparrow$ & Regret\,$\downarrow$ & Ord.$\uparrow$ & Regret\,$\downarrow$ & Ord.$\uparrow$ & Regret\,$\downarrow$ & Ord.$\uparrow$ \\
\midrule
Branin         
 & $21\%$ & $.030 \pm {\scriptstyle .024}$ & 81.5 & $.034 \pm {\scriptstyle .032}$ & 69.1 & $.031 \pm {\scriptstyle .019}$ & 82.0 & $\mathbf{.020} \pm {\scriptstyle \mathbf{.017}}$ & \textbf{90.2} \\
Six-hump 
 & $34\%$ & NA & NA & $.012 \pm {\scriptstyle .010}$ & 72.4 & $.014 \pm {\scriptstyle .011}$ & 78.1 & $\mathbf{.009} \pm {\scriptstyle \mathbf{.004}}$ & \textbf{84.3} \\
Bohachevsky    
 & $11\%$ & NA & NA & $.004 \pm {\scriptstyle .004}$ & 77.2 & $.004 \pm {\scriptstyle .005}$ & 91.7 & $\mathbf{.001} \pm {\scriptstyle \mathbf{.001}}$ & \textbf{98.8} \\
\hdashline
Levy13         
 & $17\%$ & $.022 \pm {\scriptstyle .015}$ & 70.5 & $.016 \pm {\scriptstyle .012}$ & 77.0 & $.029 \pm {\scriptstyle .031}$ & 81.9 & $\mathbf{.011} \pm {\scriptstyle \mathbf{.016}}$ & \textbf{87.4} \\
Bukin6         
 & $10\%$ & $.182 \pm {\scriptstyle .116}$ & 79.4 & $.126 \pm {\scriptstyle .078}$ & 78.3 & $.155 \pm {\scriptstyle .080}$ & 91.5 & $\mathbf{.115} \pm {\scriptstyle \mathbf{.053}}$ & \textbf{97.0} \\
Cross-Tray  
 & $20\%$ 
 & $.216 \pm {\scriptstyle .093}$ & 66.9
 & $.177 \pm {\scriptstyle .112}$ & 60.2 
 & $.177 \pm {\scriptstyle .107}$ & 64.1 
 & $\mathbf{.150} \pm {\scriptstyle \mathbf{.064}}$ & \textbf{68.3} \\
Ackley         
 & $21\%$ & NA & NA & $.149 \pm {\scriptstyle .118}$ & 72.9 & $.163 \pm {\scriptstyle .084}$ & 79.9 & $\mathbf{.093} \pm {\scriptstyle \mathbf{.052}}$ & \textbf{88.7} \\
\bottomrule
\end{tabular}
\label{tab:combined_results}
\end{table*}

\subsection{Is Modeling JND in the Observation Model Sufficient?}
\label{sec:jnd_acquisition_ablation}

An important question is whether modeling indifference in the surrogate is sufficient, or whether the JND should also influence candidate selection. To isolate the effect of the acquisition function, we compare CPBO$_{\hat{\gamma}}$ against two UCB-based alternatives while keeping the surrogate model and inference procedure fixed. All three methods use the same GP prior with an ARD-RBF kernel, three-outcome Thurstone--Mosteller likelihood with a trainable JND threshold $\gamma$, and variational posterior inference via ELBO maximization, as described in Section~\ref{sec:latent_utility_model} and Appendix~\ref{app:gp_model_details}.

In other words, these baselines account for indifferences in user responses and when learning the latent proxy, but not in selecting the next configuration, in contrast to the CPBO acquisition Eq.~\eqref{eq:acquisition_info_gain} that incorporates the responses also into the expected information gain.

To construct the baselines, we consider standard GP-UCB and a consecutive variant, C-UCB, motivated by the dueling formulation of preferential BO \citep{gonzalez2017pbo}. C-UCB applies the same UCB principle to the posterior utility difference relative to the previous candidate, using the GP preference-difference construction of \citet{houlsby2012collaborative}.
Standard UCB selects the next configuration using
$\alpha_{\mathrm{UCB}}(\configuration)
= \mu_t(\configuration) + \beta s_t(\configuration)$,
where $\mu_t(\configuration)$ and $s_t(\configuration)$ are the posterior mean and standard deviation of the latent utility. To account for the fixed consecutive reference, C-UCB instead applies the same principle to the posterior utility difference,
$\alpha_{\mathrm{C\text{-}UCB}}(\configuration;\configuration_t)
= \mu_{\Delta,t}(\configuration) + \beta s_{\Delta,t}(\configuration)$,
where $\mu_{\Delta,t}(\configuration)
= \mu_t(\configuration)-\mu_t(\configuration_t)$ and
$s_{\Delta,t}(\configuration)$ is the corresponding posterior standard deviation. We use a fixed exploration coefficient $\beta=2$
for both UCB variants. Full definitions and implementation details are provided in Appendix~\ref{app:simple_acquisition_baselines}.

\begin{table}[t]
\centering
\caption{Comparison of acquisition functions using the same JND-based surrogate model and consecutive three-outcome feedback under $\gamma_{\text{true}}=0.04$. Inference regret is reported as mean $\pm$ sd over 30 runs, and ordinal accuracy as the mean over the same runs. Bold indicates the best mean performance for each benchmark and metric.}
\vspace{0.5em}
\setlength{\tabcolsep}{5.7pt}
\renewcommand{\arraystretch}{1.1}

\begin{tabular}{lcccccc}
\toprule
\multirow{2}{*}{$\utility(\candidate)$}
  & \multicolumn{2}{c}{\textbf{UCB}}
  & \multicolumn{2}{c}{\textbf{C-UCB}}
  & \multicolumn{2}{c}{\textbf{CPBO$_{\hat{\gamma}}$}} \\
\cmidrule(lr){2-3} \cmidrule(lr){4-5} \cmidrule(lr){6-7}
  & Regret$\downarrow$ & Ord.$\uparrow$
  & Regret$\downarrow$ & Ord.$\uparrow$
  & Regret$\downarrow$ & Ord.$\uparrow$ \\
\midrule

Branin
& $.047 \pm {\scriptstyle .061}$ & 54.1
& $.039 \pm {\scriptstyle .042}$ & 65.3
& $\mathbf{.020} \pm {\scriptstyle \mathbf{.017}}$ & \textbf{90.2} \\

Six-hump
& $.018 \pm {\scriptstyle .028}$ & 53.7
& $.016 \pm {\scriptstyle .022}$ & 67.8
& $\mathbf{.009} \pm {\scriptstyle \mathbf{.004}}$ & \textbf{84.3} \\

Bohachevsky
& $.008 \pm {\scriptstyle .011}$ & 68.6
& $.004 \pm {\scriptstyle .002}$ & 78.0
& $\mathbf{.001} \pm {\scriptstyle \mathbf{.001}}$ & \textbf{98.8} \\

\bottomrule
\end{tabular}
\label{tab:ucb_acquisition_ablation_main}
\end{table}

Table~\ref{tab:ucb_acquisition_ablation_main} shows that CPBO$_{\hat{\gamma}}$ achieves lower mean inference regret and substantially higher mean ordinal accuracy than these baselines on all three benchmark functions.
C-UCB achieves better mean performance than standard UCB, suggesting that conditioning the acquisition on the fixed consecutive reference is beneficial, but it
is worth noting that both tend to perform worse than EUBO (Table~\ref{tab:combined_results}) that is limited to binary responses. Overall, the results indicate that incorporating indifference into the surrogate alone does not provide the full benefit of the JND model, but rather it should influence also the acquisitions.


\subsection{Case Study Demonstration: Extrusion Optimization}
\label{sec:experiment_extrusion}

\paragraph{Task.} 
We demonstrate CPBO in high-moisture extrusion processing (HMEP), a central step in a typical research process for developing new plant-based meat analogues.
HMEP enables the formation of fibrous meat-like structures at an industrial scale through continuous operation, offering a sustainable alternative to conventional animal meat, which is a major contributor to greenhouse gas emissions \citep{dekkers2018meat}.
 
A twin-screw extruder processes powdery ingredients as a continuous stream, and the properties of the extrudate (final outcome) at any given time are influenced by its settings, here $\configuration \in \mathbb{R}^3$ indicating the temperature, water feed rate, and screw speed. The goal is to find $\hat{\configuration}$ such that the outcome is sufficiently good for subsequent research stages, such as instrumental texture analyses and sensory evaluation by a trained panel \citep{nisov2024peptide}.
The preliminary optimization for a new ingredient mixture may take even a full working day and needs to be done repeatedly for alternative ingredients, making it a practical bottleneck. Moreover, the time and quality of the outcome heavily rely on the level of experience of the operator. 
Algorithmic support improves reproducibility across operators by reducing reliance on current subjective trial-and-error approach, especially for less experienced operators.

\noindent\textbf{Extrusion Optimization as CPBO.}~
The HMEP optimization task matches \emph{exactly} the CPBO setup with JND.
After changing the settings, we only observe the outcome once the material has propagated through the extruder. This incurs a production cost $\cost_p$, approximately 5-10 minutes per iteration. 
Each evaluation has a time cost $\cost_e$ of the same magnitude due to the simultaneous assessment of multiple characteristics (e.g., fibrousness and gumminess). This requires a preferential comparison of overall quality rather than a numerical evaluation of the physical characteristics \citep{mertens2021difficulty}, especially when needing to pay simultaneous attention to the extruder \citep{allred2016memory}.
Often the operator, especially an inexperienced one, cannot distinguish between the candidates, e.g., when both are poor in different ways, making modeling of JND critical.
Finally, the candidates need to be evaluated before they cool down. This happens on time-scales comparable to the production delay,  and hence only the latest candidate is suitable for reliable comparisons. Use of older samples would require re-heating, with significant retrieval cost $\cost_r$.

\paragraph{Experimental Setup.} We conducted a study involving three operators with varying level of experience, each solving the same optimization problem by following CPBO recommendations and only indicating the subjective preferences for consecutive comparisons while operating the extruder. Because the optimization target is the operator-specific subjective overall quality of the extrudate, there is no independent scalar ground-truth objective for this study. After 30 iterations (first 15 with a static design but still asking for preferential judgments for learning $\utility(\configuration)$), we produced the candidate for $\argmax \utility(\configuration)$ as the final solution. Afterwards, each operator assessed whether the final candidate was satisfactory, and they were given the opportunity to manually explore the configuration space to find a better solution. See Appendix~\ref{app:extrusion_optimization} for all details.

\begin{figure*}[t]
    \centering
        \includegraphics[width=0.9\linewidth]{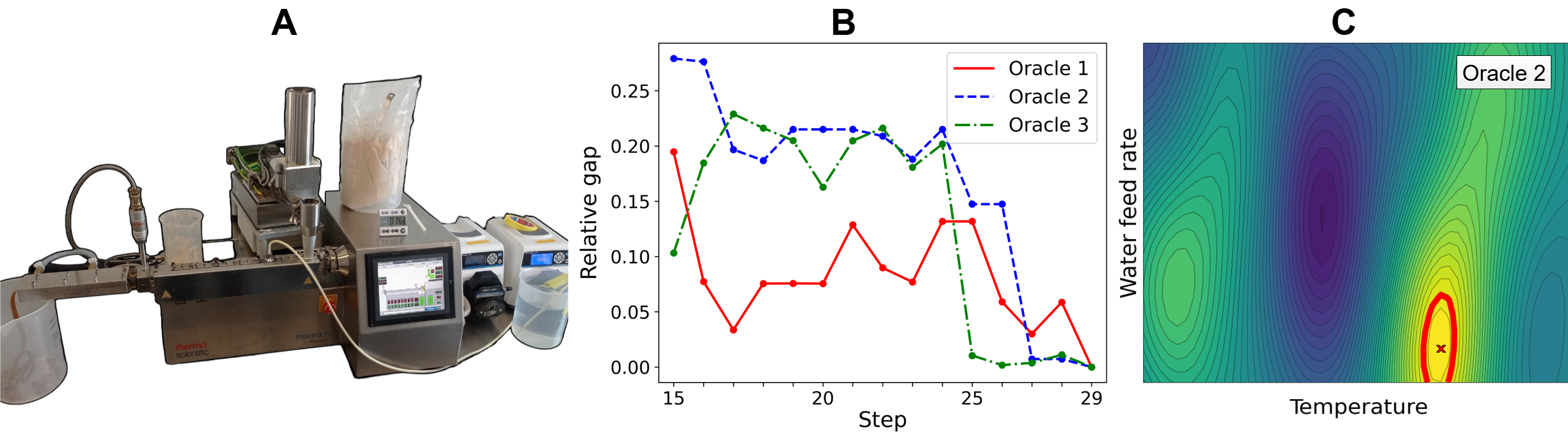}
    \caption{
        (A): High-moisture extruder, producing a candidate $\candidate$ based on current configuration $\configuration$, for sensory comparison against the previous candidate. 
        (B): Model-based utility gap relative to the final recommendation for three operators during optimization.
        (C): 2D slice of the operator's (Oracle 2) learned utility: points on or inside the red contour are predicted to be indifferent to the posterior maximizer (``x").
    }
    \label{fig:experiment_extrusion}
\end{figure*}

\noindent\textbf{Results.}
Fig.~\ref{fig:experiment_extrusion} shows the optimization trajectories for the three operators during the CPBO iterations. Since the true optimum is not known, Fig.~\ref{fig:experiment_extrusion}B reports a model-based utility gap relative to each operator's final recommendation rather than inference regret to a known optimum. The gaps become small toward the end of the optimization, but this should not be interpreted as convergence to the global optimum. The learned $\gamma$ ($0.02, 0.04$ and $0.05$) were consistent for the three operators, each indicating indifference $2$ to $4$ times; the choice of $\gamma_{\text{true}}=0.04$ for the earlier experiment was in part motivated by the desire to make the ratio of indifferences similar to this demonstration.

All operators expressed satisfaction with the result, judging the extrudate to be suitable for subsequent stages. After carrying out 1 to 4 attempts at finding better configurations by any means possible, none of the operators preferred these additional configurations over the final CPBO recommendation. This is indication of the method's usefulness, but is not to be considered as a controlled comparison that would establish global optimality. Rather, the results support the practical feasibility of CPBO for the application task: all operators found settings that produced satisfactory extrudates.
Fig.~\ref{fig:experiment_extrusion}C shows the learned utility proxy for one operator.  Even though the utility landscape is multimodal, the other mode is predicted to be worse than the posterior maximizer, revealed by the JND only covering configurations around the mode. The operator confirmed this by indicating a difference between the modes.

\section{Discussion and Conclusion}
\label{sec:discussion_conclusion}

Despite active efforts to support AI-assisted optimization \citep{rainforth2024modern, lofgren2020boss, klami2024virtual}, complex R\&D optimization tasks that do not fit the standard BO setup remain unsupported in practice. Our main arguments and contributions relating to better accounting for the costs and operational constraints may seem obvious in hindsight, also because of the massive empirical improvement obtained with relatively simple modifications. However, the current PBO literature indeed ignores both the cost of producing the candidates and the obvious need for experts to indicate indifference. We speculate this is in part due to the community focusing on clean benchmarks and computational BO tasks, such as hyperparameter optimization \citep{bergstra2013optim, snoek2012practical}, often neglecting empirical sciences that may have the biggest need for these tools.

Our main contribution is the task formulation and the identification of its necessary components, rather than the specific algorithm. However, the proposed method is already highly robust and easy to use in practice, not requiring any adaptation or hyperparameter tuning for the different target functions (including the real use case). It could, however, be made more computationally efficient, for example by amortizing the information gain computation  \citep{volpp2020meta, swersky2020amortized}. Our current objective also does not account for the quality of intermediate candidates. In applications where poor intermediate outcomes are themselves costly, the objective could be extended to include cumulative regret or quality-dependent production costs. Empirical validation could also be extended to higher dimensions. Here we considered only problems of at most twelve dimensions (see Appendix~\ref{cpbo_benchmarks}) to align with the typical use-cases in empirical science. For higher dimensionalities, CPBO should scale like any PBO method relying on pairwise comparison.

We demonstrated CPBO in empirical research in food science, aiming to speed-up food extrusion processing. However, the range of potential applications is broader than food extrusion. CPBO is relevant when candidate production is costly and comparisons are naturally or operationally restricted to the immediately previous candidate, for example because candidates deteriorate or retrieving earlier candidates is costly. Additionally, in human preference settings, a strictly zero JND is generally unrealistic. For example, limitations of human visual, auditory, and tactile perception alone make it impossible to distinguish tiny differences in stimuli \citep{johnson2002neural}, which makes this work directly relevant to various human-in-the-loop formulations \citep{catkin2023haptic, adachi2024looping}.   


\subsubsection*{Acknowledgements}

University of Helsinki (UH) researchers were supported by the
Research Council of Finland Flagship programme: Finnish Center for Artificial Intelligence FCAI, and by Business Finland Grant No. 3476/31/2023. VTT and UH researchers were supported by the Global Center for Food Innovation and Diversification to Advance the Bioeconomy (FoodID; Business Finland Grant No. 3545/31/2024; Research Council of Finland Grants No. 366905 and 366906). UH researchers acknowledge the research environment provided by the ELLIS Institute Finland.

We thank the extrusion experiment operators at VTT Technical Research Centre of Finland Ltd for their expert assistance. Computational resources were provided by CSC – IT Center for Science. 

\bibliography{references}

@article{thurstone1927judgment,
  author  = {Thurstone, L. L.},
  title   = {A Law of Comparative Judgment},
  journal = {Psychological Review},
  volume  = {34},
  number  = {4},
  pages   = {273--286},
  year    = {1927}
}

@article{mosteller1951remarks,
  author  = {Mosteller, Frederick},
  title   = {Remarks on the Method of Paired Comparisons: I. The Least Squares Solution Assuming Equal Standard Deviations and Equal Correlations},
  journal = {Psychometrika},
  volume  = {16},
  number  = {1},
  pages   = {3--9},
  year    = {1951}
}

@article{davidson1970extension,
  author  = {Davidson, Roger R.},
  title   = {On Extending the {B}radley--{T}erry Model to Accommodate Ties in Paired Comparison Experiments},
  journal = {Journal of the American Statistical Association},
  volume  = {65},
  number  = {329},
  pages   = {317--328},
  year    = {1970}
}

@incollection{mcfadden1974rum,
  author    = {McFadden, Daniel},
  title     = {Conditional Logit Analysis of Qualitative Choice Behavior},
  booktitle = {Frontiers in Econometrics},
  publisher = {Academic Press},
  address   = {New York},
  year      = {1974},
  pages     = {105--142}
}

@article{manski1977structure,
  author  = {Manski, Charles F.},
  title   = {The Structure of Random Utility Models},
  journal = {Theory and Decision},
  volume  = {8},
  number  = {3},
  pages   = {229--254},
  year    = {1977}
}

@incollection{weber1978sense,
  author    = {Weber, Ernst Heinrich},
  title     = {The Sense of {T}ouch},
  booktitle = {The Sense of {T}ouch},
  editor    = {Ross, H. E. and Murray, D. J.},
  publisher = {Academic Press},
  address   = {New York, NY},
  year      = {1978}
}

@article{mckay1979lhs,
  author  = {McKay, M. D. and Beckman, R. J. and Conover, W. J.},
  title   = {A Comparison of Three Methods for Selecting Values of Input Variables in the Analysis of Output from a Computer Code},
  journal = {Technometrics},
  volume  = {21},
  number  = {2},
  pages   = {239--245},
  year    = {1979}
}

@inproceedings{chu2005preference,
  author    = {Chu, Wei and Ghahramani, Zoubin},
  title     = {Preference Learning with {G}aussian Processes},
  booktitle = {Proceedings of the 22nd International Conference on Machine Learning},
  pages     = {137--144},
  year      = {2005}
}

@book{williams2006gaussian,
  author    = {Williams, Christopher K. I. and Rasmussen, Carl Edward},
  title     = {{G}aussian Processes for Machine Learning},
  publisher = {MIT Press},
  year      = {2006}
}

@inproceedings{zhou2008learning,
  author    = {Zhou, Ke and Xue, Gui-Rong and Zha, Hongyuan and Yu, Yong},
  title     = {Learning to Rank with Ties},
  booktitle = {Proceedings of the 31st Annual International ACM SIGIR Conference on Research and Development in Information Retrieval},
  pages     = {275--282},
  year      = {2008}
}

@inproceedings{gardner2018gpytorch,
  author    = {Gardner, Jacob and Pleiss, Geoff and Bindel, David and Weinberger, Kilian Q. and Wilson, Andrew Gordon},
  title     = {GPyTorch: Blackbox Matrix-Matrix Gaussian Process Inference with {GPU} Acceleration},
  booktitle = {Advances in Neural Information Processing Systems},
  volume    = {31},
  year      = {2018}
}

@inproceedings{titsias2009variational,
  author    = {Titsias, Michalis},
  title     = {Variational Learning of Inducing Variables in Sparse {G}aussian Processes},
  booktitle = {Proceedings of the Twelfth International Conference on Artificial Intelligence and Statistics},
  pages     = {567--574},
  year      = {2009}
}

@inproceedings{houlsby2012collaborative,
  author    = {Houlsby, Neil and Husz{\'a}r, Ferenc and Ghahramani, Zoubin and Hern{\'a}ndez-Lobato, Jos{\'e} Miguel},
  title     = {Collaborative Gaussian Processes for Preference Learning},
  booktitle = {Advances in Neural Information Processing Systems},
  volume    = {25},
  year      = {2012}
}

@inproceedings{snoek2012practical,
  author    = {Snoek, Jasper and Larochelle, Hugo and Adams, Ryan P.},
  title     = {Practical {B}ayesian Optimization of Machine Learning Algorithms},
  booktitle = {Advances in Neural Information Processing Systems},
  volume    = {25},
  year      = {2012}
}

@misc{surjanovic2013virtual,
  author       = {Surjanovic, S. and Bingham, D.},
  title        = {Virtual Library of Simulation Experiments: Test Functions and Datasets},
  howpublished = {\url{https://www.sfu.ca/\string~ssurjano/}},
  note         = {Accessed: 2026-02-11},
  year         = {2013}
}

@inproceedings{swersky2013multitask,
  author    = {Swersky, Kevin and Snoek, Jasper and Adams, Ryan P.},
  title     = {Multi-Task {B}ayesian Optimization},
  booktitle = {Advances in Neural Information Processing Systems},
  volume    = {26},
  year      = {2013}
}

@inproceedings{bergstra2013optim,
  author    = {Bergstra, James and Yamins, Daniel and Cox, David},
  title     = {Making a Science of Model Search: Hyperparameter Optimization in Hundreds of Dimensions for Vision Architectures},
  booktitle = {Proceedings of the 30th International Conference on Machine Learning},
  pages     = {115--123},
  year      = {2013}
}

@inproceedings{gonzalez2017pbo,
  author    = {González, Javier and Dai, Zhenwen and Damianou, Andreas and Lawrence, Neil D.},
  title     = {Preferential {B}ayesian Optimization},
  booktitle = {Proceedings of the 34th International Conference on Machine Learning},
  pages     = {1282--1291},
  year      = {2017}
}

@inproceedings{wang2017MES,
  author    = {Wang, Zi and Jegelka, Stefanie},
  title     = {{Max-Value} Entropy Search for Efficient {B}ayesian Optimization},
  booktitle = {Proceedings of the 34th International Conference on Machine Learning},
  pages     = {3627--3635},
  year      = {2017}
}

@book{garnett_bayesoptbook_2023,
  author    = {Garnett, Roman},
  title     = {{Bayesian Optimization}},
  year      = {2023},
  publisher = {Cambridge University Press}
}

@inproceedings{mikkola2020ppbo,
  author    = {Mikkola, Petrus and Todorović, Milica and Järvi, Jari and Rinke, Patrick and Kaski, Samuel},
  title     = {Projective Preferential {B}ayesian Optimization},
  booktitle = {Proceedings of the 37th International Conference on Machine Learning},
  pages     = {6884--6892},
  year      = {2020}
}

@misc{lofgren2020boss,
  author = {L{\"o}fgren, Joakim and Todorovic, Milica and Rinke, Patrick and Paulam{\"a}ki, Henri and Tolvanen, Arttu and Parkkinen, Ville and Remes, Ulpu and Sten, Nuutti},
  title  = {{B}ayesian Optimization Structure Search ({BOSS})},
  year   = {2020}
}

@misc{lee2020costawarebo,
  author       = {Lee, Eric Hans and Perrone, Valerio and Archambeau, Cedric and Seeger, Matthias},
  title        = {Cost-Aware {B}ayesian Optimization},
  howpublished = {arXiv preprint arXiv:2003.10870},
  year         = {2020}
}

@article{nguyen2021top,
  author    = {Nguyen, Quoc Phong and Tay, Sebastian and Low, Bryan Kian Hsiang and Jaillet, Patrick},
  title     = {Top-\textit{k} Ranking {B}ayesian Optimization},
  journal = {Proceedings of the AAAI Conference on Artificial Intelligence},
  volume    = {35},
  pages     = {9135--9143},
  year      = {2021}
}

@article{luong2021costawarebo,
  author  = {Luong, Phuc and Nguyen, Dang and Gupta, Sunil and Rana, Santu and Venkatesh, Svetha},
  title   = {Adaptive Cost-Aware {B}ayesian Optimization},
  journal = {Knowledge-Based Systems},
  volume  = {232},
  pages   = {107481},
  year    = {2021}
}

@inproceedings{agarwal2022duel,
  author    = {Agarwal, Arpit and Ghuge, Rohan and Nagarajan, Viswanath},
  title     = {Batched Dueling Bandits},
  booktitle = {Proceedings of the 39th International Conference on Machine Learning},
  pages     = {89--110},
  year      = {2022}
}

@inproceedings{lin2022eubo,
  author    = {Lin, Zhiyuan Jerry and Astudillo, Raul and Frazier, Peter I. and Bakshy, Eytan},
  title     = {Preference Exploration for Efficient {B}ayesian Optimization with Multiple Outcomes},
  booktitle = {Proceedings of the 25th International Conference on Artificial Intelligence and Statistics},
  year      = {2022}
}

@inproceedings{astudillo2023qeubo,
  author    = {Astudillo, Raul and Lin, Zhiyuan Jerry and Bakshy, Eytan and Frazier, Peter},
  title     = {{qEUBO}: A Decision-Theoretic Acquisition Function for Preferential {B}ayesian Optimization},
  booktitle = {Proceedings of the 26th International Conference on Artificial Intelligence and Statistics},
  pages     = {1093--1114},
  year      = {2023}
}

@misc{liu2024reward,
  author       = {Liu, Jinsong and Ge, Dongdong and Zhu, Ruihao},
  title        = {Reward Learning from Preference with Ties},
  howpublished = {arXiv preprint arXiv:2410.05328},
  year         = {2024}
}

@article{rainforth2024modern,
  author  = {Rainforth, Tom and Foster, Adam and Ivanova, Desi R. and Smith, Freddie Bickford},
  title   = {Modern {B}ayesian Experimental Design},
  journal = {Statistical Science},
  volume  = {39},
  number  = {1},
  pages   = {100--114},
  year    = {2024}
}

@article{catkin2023haptic,
  author  = {Catkin, Bilal and Patoglu, Volkan},
  title   = {Preference-Based Human-in-the-Loop Optimization for Perceived Realism of Haptic Rendering},
  journal = {IEEE Transactions on Haptics},
  volume  = {16},
  number  = {4},
  pages   = {470--476},
  year    = {2023}
}

@inproceedings{xu2024popbo,
  author    = {Xu, Wenjie and Wang, Wenbin and Jiang, Yuning and Svetozarevic, Bratislav and Jones, Colin N.},
  title     = {Principled Preferential {B}ayesian Optimization},
  booktitle = {Proceedings of the 41st International Conference on Machine Learning},
  year      = {2024}
}

@misc{benavoli2024tutorial,
      title={A tutorial on learning from preferences and choices with Gaussian Processes}, 
      author={Alessio Benavoli and Dario Azzimonti},
      year={2024},
      eprint={2403.11782},
      archivePrefix={arXiv},
}

@inproceedings{adachi2024looping,
  author    = {Adachi, Masaki and Planden, Brady and Howey, David and Osborne, Michael A. and Orbell, Sebastian and Ares, Natalia and Muandet, Krikamol and Lun Chau, Siu},
  title     = {Looping in the Human: Collaborative and Explainable {B}ayesian Optimization},
  booktitle = {Proceedings of the 27th International Conference on Artificial Intelligence and Statistics},
  pages     = {505--513},
  year      = {2024}
}

@article{klami2024virtual,
  author  = {Klami, Arto and Damoulas, Theo and Engkvist, Ola and Rinke, Patrick and Kaski, Samuel},
  title   = {Virtual Laboratories: Transforming Research with {AI}},
  journal = {Data-Centric Engineering},
  volume  = {5},
  pages   = {e19},
  year    = {2024}
}

@inproceedings{kochanski2017dessert,
  author    = {Kochanski, Greg and Golovin, Daniel and Karro, John and Solnik, Benjamin and Moitra, Subhodeep and Sculley, D.},
  title     = {Bayesian Optimization for a Better Dessert},
  booktitle = {Proceedings of the NeurIPS 2017 Workshop on Bayesian Optimization},
  year      = {2017}
}

@inproceedings{swersky2020amortized,
  author    = {Swersky, Kevin and Rubanova, Yulia and Dohan, David and Murphy, Kevin},
  title     = {Amortized {B}ayesian Optimization over Discrete Spaces},
  booktitle = {Proceedings of the 36th Conference on Uncertainty in Artificial Intelligence},
  pages     = {769--778},
  year      = {2020}
}

@inproceedings{volpp2020meta,
  author    = {Volpp, Michael and Fr{\"o}hlich, Lukas P. and Fischer, Kirsten and Doerr, Andreas and Falkner, Stefan and Hutter, Frank and Daniel, Christian},
  title     = {Meta-Learning Acquisition Functions for Transfer Learning in {B}ayesian Optimization},
  booktitle = {Proceedings of the International Conference on Learning Representations},
  year      = {2020}
}

@book{lawless2010sensory,
  author    = {Lawless, Harry T. and Heymann, Hildegarde},
  title     = {Sensory Evaluation of Food: Principles and Practices},
  publisher = {Springer Science \& Business Media},
  year      = {2010}
}

@article{allred2016memory,
  author  = {Allred, Sarah R. and Crawford, L. Elizabeth and Duffy, Sean and Smith, John},
  title   = {Working Memory and Spatial Judgments: Cognitive Load Increases the Central Tendency Bias},
  journal = {Psychonomic Bulletin \& Review},
  volume  = {23},
  number  = {6},
  pages   = {1825--1831},
  year    = {2016}
}

@article{dekkers2018meat,
  author  = {Dekkers, Birgit L. and Boom, Remko M. and van der Goot, Atze Jan},
  title   = {Structuring Processes for Meat Analogues},
  journal = {Trends in Food Science \& Technology},
  volume  = {81},
  pages   = {25--36},
  year    = {2018}
}

@article{mertens2021difficulty,
  author  = {Mertens, Alica and Mertens, Ulf K. and Lerche, Veronika},
  title   = {On the Difficulty to Think in Ratios: A Methodological Bias in Stevens’ Magnitude Estimation Procedure},
  journal = {Attention, Perception, \& Psychophysics},
  volume  = {83},
  number  = {5},
  pages   = {2347--2365},
  year    = {2021}
}

@article{nisov2024peptide,
  author  = {Nisov, Anni and Valtonen, Anniina and Aisala, Heikki and Spaccasassi, Andrea and Walser, Christoph and Dawid, Corinna and Sozer, Nesli},
  title   = {Effect of Peptide Formation During Rapeseed Fermentation on Meat Analogue Structure and Sensory Properties at Different {pH} Conditions},
  journal = {Food Research International},
  volume  = {180},
  pages   = {114070},
  year    = {2024}
}

@article{johnson2002neural,
  title={Neural coding and the basic law of psychophysics},
  author={Johnson, Kenneth O. and Hsiao, Steven S. and Yoshioka, Takashi},
  journal={The Neuroscientist},
  volume={8},
  number={2},
  pages={111--121},
  year={2002},
  publisher={SAGE Publications Sage CA: Los Angeles, CA}
}

@inproceedings{brochu2010bayesian,
  title={A Bayesian interactive optimization approach to procedural animation design},
  author={Brochu, Eric and Brochu, Tyson and De Freitas, Nando},
  booktitle={Proceedings of the 2010 ACM SIGGRAPH/Eurographics Symposium on Computer Animation},
  pages={103--112},
  year={2010}
}

@article{balandat2020botorch,
  title={{BoTorch}: A framework for efficient Monte-Carlo Bayesian optimization},
  author={Balandat, Maximilian and Karrer, Brian and Jiang, Daniel and Daulton, Samuel and Letham, Ben and Wilson, Andrew G and Bakshy, Eytan},
  journal={Advances in neural information processing systems},
  volume={33},
  pages={21524--21538},
  year={2020}
}

@article{bradley1952rank,
  title={Rank analysis of incomplete block designs: I. The method of paired comparisons},
  author={Bradley, Ralph Allan and Terry, Milton},
  journal={Biometrika},
  volume={39},
  number={3/4},
  pages={324--345},
  year={1952},
  publisher={JSTOR}
}

@article{plackett1975analysis,
  title={The analysis of permutations},
  author={Plackett, Robin L},
  journal={Journal of the Royal Statistical Society Series C: Applied Statistics},
  volume={24},
  number={2},
  pages={193--202},
  year={1975},
  publisher={Oxford University Press}
}

@book{luce1959individual,
  title     = {Individual Choice Behavior},
  author    = {Luce, R. Duncan},
  year      = {1959},
  publisher = {John Wiley}
}
\bibliographystyle{tmlr}


\clearpage


\appendix

\setcounter{figure}{0}
\setcounter{table}{0}
\setcounter{equation}{0}

\renewcommand{\thefigure}{S\arabic{figure}}
\renewcommand{\thetable}{S\arabic{table}}
\renewcommand{\theequation}{S\arabic{equation}}

\renewcommand{\theHfigure}{S\arabic{figure}}
\renewcommand{\theHtable}{S\arabic{table}}
\renewcommand{\theHequation}{S\arabic{equation}}

\section{Appendix}
The appendix complements the main manuscript, providing additional method details (Section~\ref{app:method_details}), details of the empirical experiments on the benchmark datasets with additional illustrations and numerical results (Section~\ref{app:experiment_details}), and description of the high-moisture extrusion experiment (Section~\ref{app:extrusion_optimization}).

All figures and tables in this Appendix are numbered separately from the main paper, using the format Figure S1, Table S1, and similar.


\section{Method Details}
\label{app:method_details}

In this section, we provide high-level pseudocode for the complete CPBO procedure, followed by details of the surrogate GP model, three-outcome likelihood, and acquisition-function implementation, together with the corresponding hyperparameter choices.


\subsection{CPBO Algorithm}
\label{app:cpbo_algorithm}

Algorithm~\ref{alg:cpbo} provides high-level pseudo-code for the complete CPBO procedure, while lower-level implementation and numerical details are described in the subsequent subsections. The procedure starts from preference data $\mathcal{D}_{t_0}$ and the currently available reference candidate $\candidate_{t_0}=\candidate(\configuration_{t_0})$. The response returned by \textsc{Query} follows Eq.~\eqref{eq:preference_model}.

\begin{algorithm}[t]
\caption{Consecutive Preferential Bayesian Optimization (CPBO)}
\label{alg:cpbo}
\begin{algorithmic}[1]
\setlength{\itemsep}{0.2em}

\For{$t=t_0,\ldots,T-1$}

    \State $(q_t(\utility),\hat{\gamma}_t)
    \gets \Call{FitModel}{\mathcal{D}_t}$

    \State $\hat{\alpha}_t(\configuration)
    \gets
    \Call{Acquisition}
    {q_t(\utility),\hat{\gamma}_t,\configuration_t,\configuration},
    \quad \configuration\in\mathcal{X}$

    \State $\configuration_{t+1}
    \gets
    \arg\max_{\configuration\in\mathcal{X}}
    \hat{\alpha}_t(\configuration)$

    \State $\candidate_{t+1}
    \gets \Call{Produce}{\configuration_{t+1}}$

    \State $\response_{t+1}
    \gets \Call{Query}{\candidate_t,\candidate_{t+1}}$

    \State $\mathcal{D}_{t+1}
    \gets
    \mathcal{D}_t
    \cup
    \{([\configuration_t,\configuration_{t+1}],
    \response_{t+1})\}$

\EndFor

\State $(q_T(\utility),\hat{\gamma}_T)
\gets \Call{FitModel}{\mathcal{D}_T}$

\State \Return
$\hat{\configuration}_T
=
\arg\max_{\configuration\in\mathcal{X}}
\mu_T(\configuration)$

\Statex

\Procedure{Acquisition}
{$q_t(\utility),\hat{\gamma}_t,\configuration_t,\configuration$}

    \State $\hat{\mathcal{X}}
    \gets \Call{Discretize}{\mathcal{X}}$

    \State $G
    \gets
    \Call{FitGumbel}
    {q_t(\utility),\hat{\mathcal{X}}}$

    \State $\{\utility_k^*\}_{k=1}^{M}
    \gets \Call{Sample}{G}$

    \State $\{(\bar{\utility}_j^*,n_j)\}_{j=1}^{J}
    \gets
    \Call{Bin}{\{\utility_k^*\}_{k=1}^{M}}$

    \State $H
    \gets
    \Call{ResponseEntropy}
    {q_t(\utility),\configuration_t,\configuration,\hat{\gamma}_t}$

    \For{$j=1,\ldots,J$}

        \State $p_{\leq\bar{\utility}_j^*}
        \gets
        q_t(\utility)
        \mid
        \left\{
        \utility(\configuration_t)\leq\bar{\utility}_j^*,
        \;
        \utility(\configuration)\leq\bar{\utility}_j^*
        \right\}$

        \State $H_j
        \gets
        \Call{ResponseEntropy}
        {p_{\leq\bar{\utility}_j^*},
        \configuration_t,\configuration,\hat{\gamma}_t}$

    \EndFor

    \State $\hat{\alpha}_t(\configuration)
    \gets
    H-\frac{1}{M}
    \sum_{j=1}^{J}n_jH_j$

    \State \Return $\hat{\alpha}_t(\configuration)$

\EndProcedure

\Statex

\Procedure{ResponseEntropy}
{$p(\utility),\configuration_t,\configuration,\gamma$}

    \State $\utility^{(s)}
    \sim p(\utility),
    \quad s=1,\ldots,S$

    \State $\Delta\utility^{(s)}
    \gets
    \utility^{(s)}(\configuration)
    -
    \utility^{(s)}(\configuration_t)$

    \State $\widehat{P}(\responseRND_{t+1}=r)
    \gets
    \frac{1}{S}
    \sum_{s=1}^{S}
    P(r\mid\Delta\utility^{(s)}),
    \quad r\in\mathcal{R}$

    \State \Return
    $-\sum_{r\in\mathcal{R}}
    \widehat{P}(\responseRND_{t+1}=r)
    \log
    \widehat{P}(\responseRND_{t+1}=r)$

\EndProcedure

\end{algorithmic}
\end{algorithm}


\subsection{Surrogate GP Model and Likelihood}
\label{app:gp_model_details}

To remove the scale ambiguity in Eq.~\ref{eq:gp_kernel}, we fix $\sigma_f^2=10.0$.
Inference is performed using a variational approximation to the posterior $p(\utility \mid \mathcal{D}_t)$ over the latent utility function $\utility(\configuration)$. At iteration $t$, we define the variational distribution over the latent utility values $\utility(\configuration_i)$ corresponding to all observed configurations ${\configuration_i}_{i=1}^t$. The corresponding latent utility values are denoted as
$ 
  \mathbf{u} = [\utility(\configuration_1), \dots, \utility(\configuration_t)]^\top.
$

A diagonal jitter of $10^{-4}$ is added to all kernel covariance matrices to ensure numerical stability during Cholesky decompositions. We implement the model using GPyTorch's variational GP framework \citep{gardner2018gpytorch}, specifically the \texttt{ApproximateGP} class with a Cholesky-parameterized variational distribution and a learned inducing set. While we use all training points as inducing inputs in our experiments (effectively removing the sparsity), our model follows the variational GP framework introduced by \citet{titsias2009variational}, as implemented in GPyTorch's \texttt{VariationalStrategy}.

While we do not apply any prior on the ARD lengthscales $\{\ell_d\}$ in the benchmark experiments, we found it occasionally helpful in practice to encourage moderate smoothness early in training. In particular, for the real-world extrusion experiment, we placed independent Gamma priors $\ell_d \sim \mathrm{Gamma}(1.0, 0.05)$ (shape, rate) on each lengthscale dimension.


\paragraph{Likelihood Implementation.}

The expected log-likelihood in the ELBO is approximated by averaging over $S = 50$ samples $\utility^{(s)}$ drawn from the variational posterior $q(\utility)$
\begin{equation}
\mathbb{E}_{q(\utility)} [\log P(\mathcal{D}_t \mid \utility)] \approx \frac{1}{S} \sum_{s=1}^S \log P(\mathcal{D}_t \mid \utility^{(s)}).
\end{equation}
We found $S = 50$ to be sufficiently accurate in practice and did not explore smaller values, as the overhead was acceptable relative to overall runtime. To prevent numerical issues, we add a small constant $\varepsilon = 10^{-5}$ to the computed probabilities before applying the logarithm.

The perceptual noise parameter is fixed to $\sigma = 0.04$ to eliminate scale ambiguity between utility magnitude and noise, and sensitivity to this choice is evaluated in Section~\ref{cpbo_benchmarks}.

\paragraph{Optimization.}
We jointly optimize:
\begin{itemize}
    \item the variational parameters $\mathbf{m}_u$ and $\mathbf{L}$,
    \item the ARD lengthscales $\{\ell_d\}$ and the prior mean constant $m$,
    \item and the JND threshold $\gamma$.
\end{itemize}

Optimization is performed by maximizing the ELBO using full-batch Adam (learning rate $10^{-2}$) for a fixed number of 2000 iterations, which we found sufficient for the approximation to converge in all experiments. During training, we retain the model parameters that achieve the lowest loss across all iterations and return them for evaluation.


\subsection{Acquisition Implementation Details}
\label{app:acquisition_details}
We describe key implementation choices that improve the numerical stability and computational efficiency of our acquisition function based on MES adapted for CPBO.

\paragraph{Zero-Mean Normalization of Samples.}
To ensure numerical stability and comparability during acquisition, we apply zero-mean normalization to every posterior sample $\utility^{(s)}$ drawn when computing entropy terms or estimating the Gumbel distribution. Given a batch $\mathcal{X}_\text{eval} = \{\configuration^{(1)}, \dots, \configuration^{(N)}\}$, we define:
\begin{equation}
\tilde{\utility}^{(s)}(\configuration^{(i)}) = \utility^{(s)}(\configuration^{(i)}) - \frac{1}{N} \sum_{j=1}^{N} \utility^{(s)}(\configuration^{(j)}), \quad \text{for } i = 1, \dots, N.
\end{equation}
This transformation preserves preference structure (based on utility differences) while eliminating arbitrary shifts in the sample-wise utility scale. It improves stability and alignment across Monte Carlo samples used in acquisition.

\paragraph{Unconditional Entropy.}
We estimate $H(\responseRND_{t+1} \mid \mathcal{D}_t)$, using 1000 Monte Carlo samples from the zero-centered GP posterior. Predictive probabilities are computed from the three-outcome likelihood with fixed $\sigma$ and learned $\gamma$. As above, a small constant $\varepsilon = 10^{-5}$ is added to all probabilities before taking logarithms for numerical stability.

\paragraph{Conditional Entropy Approximation.}
The main computational bottleneck in our acquisition is evaluating the expectation over utility maxima $\utility^* \sim p(\utility^*)$ in the conditional entropy term. A naive solution would require drawing many $\utility^*$ samples and computing the full conditional preference distribution for each, which is prohibitively expensive due to the need for repeated truncated sampling. Instead, we use an efficient binning strategy
\begin{itemize}
    \item We first sample 25,000 utility maxima $\utility^*$ from the Gumbel distribution $G$ fitted to the GP posterior. Drawing these samples is computationally inexpensive relative to the subsequent truncated-posterior computations, so we use a relatively large sample count to obtain a stable approximation of the distribution of $\utility^*$.
    \item The samples are grouped into 20 equally spaced bins, and the midpoint $\bar{\utility}^*$ of each bin is used as the representative truncation level. Each representative is weighted by the fraction of sampled maximum values falling in its bin.
\end{itemize}
This reduces the number of truncation conditions to just 20. For each $\bar{\utility}^*$, we draw 1000 posterior samples at the two compared configurations and retain samples satisfying $\utility(\configuration_t)\leq\bar{\utility}^*$ and $\utility(\configuration)\leq\bar{\utility}^*$. We then apply zero-centering and compute the conditional preference quantities.

\paragraph{Gumbel Fitting and Sampling.}
We approximate the distribution of utility maxima $\utility^*$ by fitting a Gumbel distribution to posterior samples from the GP, following the approach of \citet{wang2017MES}. Specifically, we estimate the 25th, 50th, and 75th percentiles ($y_1$, $y_2$, $y_3$) of the utility samples corresponding to ranks $r_1 = 0.25$, $r_2 = 0.75$, and $r_3 = 0.50$, and compute the scale and location parameters as
\begin{equation}
b = \frac{y_1 - y_2}{\log(-\log r_2) - \log(-\log r_1)}, \quad
a = y_3 + b \cdot \log(-\log r_3).
\end{equation}
While the original MES formulation uses only the 25th and 75th percentiles, we also include the median percentile ($r_3=0.50$) to better center the fit, which we found to yield slightly improved empirical accuracy and stability. This choice also aligns with the convention used in the existing implementation of MES in BoTorch. To improve numerical stability and reduce computation time, we restrict Gumbel rank sampling to the interval $[\varepsilon_{\text{r}}, 1 - \varepsilon_{\text{r}}]$ rather than the full $[0, 1]$, which avoids extreme tail values that are difficult to truncate and slow to evaluate. We use $\varepsilon_{\text{r}} = 0.01$ in all experiments.

\begin{figure}[htbp]
    \centering
    \includegraphics[width=\textwidth]{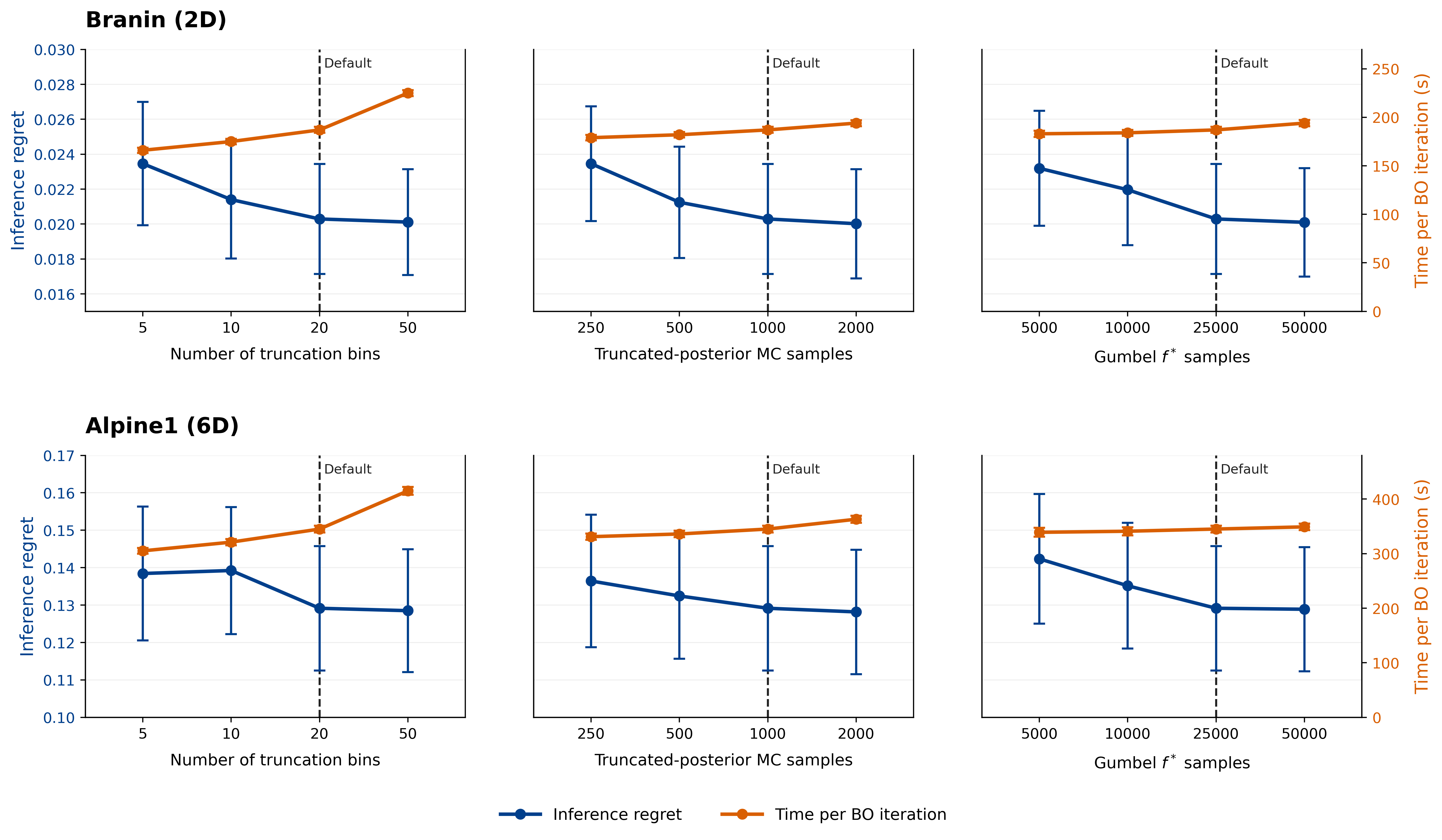}
    \caption{Sensitivity of the acquisition approximation parameters on Branin (2D) and Alpine1 (6D). Columns vary the number of truncation bins, truncated-posterior Monte Carlo samples, and Gumbel $\utility^*$ samples, respectively. For each ablation, all other algorithmic settings are held fixed at their default values, while the initial configurations and run seeds are matched across settings. Results are aggregated over 30 runs. Blue curves show mean inference regret after 30 BO iterations, while orange bars show mean wall-clock time (seconds) for one BO iteration. Error bars show $\pm$ one standard error of the mean (SEM) across runs. Dashed vertical lines indicate the default values used in the main experiments.}
    \label{fig:acq_parameter_ablation}
\end{figure}

\paragraph{Sensitivity of the Acquisition Approximation.}
We additionally evaluate sensitivity to the main approximation parameters used by the acquisition function. Figure~\ref{fig:acq_parameter_ablation} varies the number of truncation bins, the number of truncated-posterior Monte Carlo samples, and the number of Gumbel samples used for $\utility^*$, while keeping the remaining parameters at their default values. We consider both Branin (2D) and Alpine1 (6D). Across all three parameters, increasing the approximation budget generally gives diminishing improvements in inference regret while increasing computation time. The default values of 20 bins, 1000 truncated-posterior samples, and 25,000 Gumbel samples provide a practical trade-off between approximation accuracy and runtime. These values were chosen partly to keep the acquisition computation compatible with the natural 5--10 minute interval between successive runs in the real-world extrusion process, rather than by tuning them separately for synthetic benchmark performance. We therefore use the same acquisition configuration throughout the synthetic experiments and the real-world study.

\paragraph{Acquisition Maximization.}
Rather than evaluating the information gain acquisition function $\alpha(\configuration)$ exhaustively over $\configuration \in \mathcal{X}$, we optimize it using a global black-box optimization routine to find informative candidates. Specifically, we use Bayesian optimization as the optimizer, since it does not require gradient information and can handle noisy function evaluations resulting from the conditional entropy approximation. Let $\hat{\configuration} = \argmax_{\configuration \in \mathcal{X}} \alpha(\configuration)$. We approximate $\hat{\configuration}$ by running an internal Bayesian optimization loop using GPyOpt, which employs a Gaussian process surrogate with a Matérn 5/2 kernel and ARD. The procedure starts with 10 random initial evaluations, followed by 25 optimization steps using expected improvement as the acquisition strategy. This procedure yields a computationally efficient approximation of $\hat{\configuration}$ without resorting to full grid evaluation.



\section{Extended Experiments And Details}
\label{app:experiment_details}

This section provides full experimental details and additional results supporting the main paper. We organize the content into seven parts:
\begin{enumerate}
    \item \textbf{Benchmark functions:} Additional details on the benchmark functions used in the experiments.
    \item \textbf{Details for “When is CPBO needed?”:} Detailed explanation of the production and evaluation cost modeling used, including cost budgets and behavior under different regimes.
    \item \textbf{Baselines and Adaptation to CPBO:} Description of how EUBO and POP-BO were adapted to fit the CPBO setting.
    \item \textbf{Additional details for the UCB acquisition baselines:} Technical details of the standard UCB and consecutive UCB baselines used in the main-text acquisition comparison, including the posterior variance construction and experimental settings.
    \item \textbf{Statistical analysis of the main benchmark results:} Paired statistical comparisons supporting Table~\ref{tab:combined_results}, including two-sided Wilcoxon signed-rank tests, benchmark-level intersection-union tests, effect sizes, and confidence intervals for inference regret and ordinal accuracy.
    \item \textbf{Additional CPBO benchmarks:} Additional CPBO benchmarks, reporting results for all seven standard 2D functions under varying JND thresholds $\gamma_{\text{true}}$. This includes the utility recovery and choice accuracy metric analysis, ablations on the perceptual-noise scale $\sigma$, higher-dimensional benchmarks (Alpine1 6D and Levy 6D), runtime scaling with dimensionality, and a supplementary comparison against qEUBO.
    \item \textbf{Additional visualizations:} Complementary plots such as learned JND threshold, latent utility surfaces, and JND band across different functions.
\end{enumerate}

Synthetic experiments were run on the Puhti supercomputer provided by CSC - IT Center for Science, Finland, configured to use 4 CPU cores and 10 GB RAM per job. Real-world optimization was performed on a laptop with an Intel i7-9750H processor and 16 GB RAM. Acquisition parameters that impact runtime were configured so each BO step completed in approximately 5 minutes, aligning with the 5-10 minute interval required between real-life extrusion runs.


\subsection{Benchmark Functions}
\label{sec:benchmark_functions}
We consider a set of standard continuous benchmark functions \citep{surjanovic2013virtual} commonly used in the Bayesian optimization literature. For readability in tables and figures, the \emph{Six-Hump Camelback} function is referred to as \textbf{Six-hump}, and the \emph{Cross-in-Tray} function as \textbf{Cross-Tray}. 

All benchmark functions are normalized to the $[0, 1]$ range to ensure consistent interpretation of the JND thresholds $\gamma_{\text{true}}$ and to allow fair comparison across different functions. An overview of the normalized utility surfaces is provided in Figure~\ref{fig:gamma_band_visualisation}, which illustrates the function landscapes and their corresponding indifference regions.


\subsection{Details for ‘When Is CPBO Needed?’}
\label{additional_cost_results}
Section~\ref{sec:experiment_cost_aware} introduced three preferential BO strategies that differ in how candidates are generated and compared, under varying assumptions about production and evaluation cost. Here, we provide additional results to support that analysis.

The cost regimes used in our experiments are summarized in Table~\ref{tab:cost_regime}, covering settings where either production, evaluation, or both are costly.

\vspace{-0.5em}
\begin{table}[H]
\centering
\caption{Cost regimes used in our experiments}
\label{tab:cost_regime}
\begin{tabular}{lcc}
\toprule
\textbf{Regime} & \textbf{Production cost} $\cost_p$ & \textbf{Evaluation cost} $\cost_e$ \\
\midrule
$\cost_p \ll \cost_e$ & 0 & 1 \\
$\cost_p \approx \cost_e$ & 1 & 1 \\
$\cost_p \gg \cost_e$ & 1 & 0 \\
\bottomrule
\end{tabular}
\end{table}
\vspace{-0.75em}

Each strategy incurs a different per-iteration cost that depends on the relative weights of production and evaluation, denoted by $w_p$ and $w_e$. The total cost per iteration is computed as:
\vspace{-0.5em}
\begin{equation}
\text{Iteration cost} = \cost_p \cdot w_p + \cost_e \cdot w_e.
\end{equation}
\vspace{-2.0em}

The weighting terms $(w_p, w_e)$ vary by comparison strategy as follows:
\begingroup
\setlength{\itemsep}{0pt}
\setlength{\parsep}{0pt}
\setlength{\topsep}{2pt}
\setlength{\partopsep}{0pt}
\begin{itemize}
    \item \textbf{Standard:} Compares two new candidates; $w_p = 2$, $w_e = 1$.
    \item \textbf{Consecutive:} Compares the current candidate to the previous one; $w_p = 1$, $w_e = 1$.
    \item \textbf{Multiple:} Uses the same acquisition strategy as the consecutive case (conditioning on the previous candidate) but compares the new candidate against $L = 5$ previously produced candidates, including the latest; $w_p = 1$, $w_e = 5$.
\end{itemize}
\endgroup

\vspace{-0.75em}
\begin{figure}[H]
    \centering
    \includegraphics[width=0.95\textwidth]{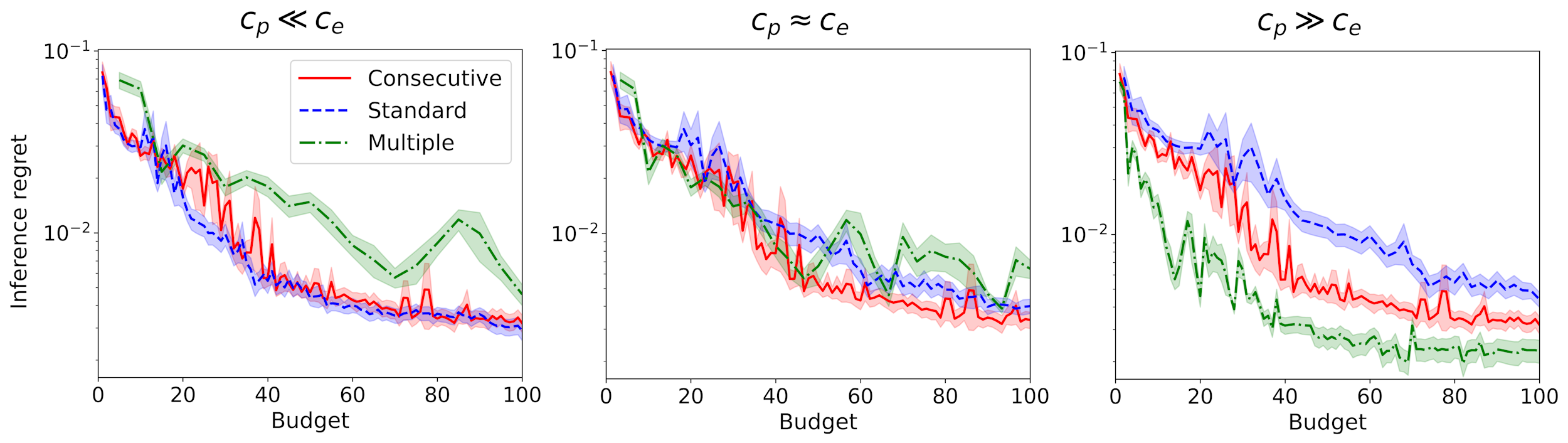}
    \vspace{-0.5em}
    \caption{Convergence of PBO variants on Branin for three cost-balance scenarios, averaged over 30 runs.}
    \label{fig:branin_cost_analysis_inference_regret}
\end{figure}
\vspace{-0.75em}

To complement the simple regret results shown in Figure~\ref{fig:branin_cost_analysis} in the main paper, Figure~\ref{fig:branin_cost_analysis_inference_regret} shows inference regret for Branin under a fixed budget of 100 units. Tables~\ref{tab:cost_simple-regret-30} and~\ref{tab:cost_inference-regret-30} provide corresponding numerical results under a fixed cost budget of 30 units. The smaller budget is used in tables for readability, as regret values become very small at 100 units.

\setlength{\tabcolsep}{4.7pt}
\renewcommand{\arraystretch}{1.1}
\begin{table}[!htbp]
  \centering
  \Large
  \caption{Simple regret under a fixed budget of 30 cost-units ($B=30$). Values are mean $\pm$ one standard deviation over 30 runs. Bold indicates the lowest mean regret within each function and cost scenario.}
  \vspace{0.5em}
  \label{tab:cost_simple-regret-30}
  \resizebox{\textwidth}{!}{%
    \begin{tabular}{l ccc|ccc|ccc}
      \toprule
      \multirow{2}{*}{\textbf{$\utility(\candidate)$}}
        & \multicolumn{3}{c|}{\textbf{$c_p\ll c_e$}}
        & \multicolumn{3}{c|}{\textbf{$c_p\approx c_e$}}
        & \multicolumn{3}{c}{\textbf{$c_p\gg c_e$}} \\
      \cmidrule(lr){2-4} \cmidrule(lr){5-7} \cmidrule(lr){8-10}
        & \textbf{Standard} & \textbf{Consecutive} & \textbf{Multiple}
        & \textbf{Standard} & \textbf{Consecutive} & \textbf{Multiple}
        & \textbf{Standard} & \textbf{Consecutive} & \textbf{Multiple} \\
      \midrule
      Branin       & $\mathbf{.006 \pm {\scriptstyle .003}}$ & $.007 \pm {\scriptstyle .003}$ & $.015 \pm {\scriptstyle .012}$
                   & $.012 \pm {\scriptstyle .008}$ & $\mathbf{.011 \pm {\scriptstyle .007}}$ & $.019 \pm {\scriptstyle .015}$
                   & $.009 \pm {\scriptstyle .006}$ & $.007 \pm {\scriptstyle .003}$ & $\mathbf{.003 \pm {\scriptstyle .002}}$ \\
      Six-hump     & $\mathbf{.0004 \pm {\scriptstyle .0003}}$ & $.0007 \pm {\scriptstyle .0008}$ & $.096 \pm {\scriptstyle .079}$
                   & $.0014 \pm {\scriptstyle .0011}$ & $\mathbf{.0009 \pm {\scriptstyle .0005}}$ & $.0110 \pm {\scriptstyle .0091}$
                   & $.0021 \pm {\scriptstyle .0019}$ & $.0015 \pm {\scriptstyle .0012}$ & $\mathbf{.0002 \pm {\scriptstyle .0001}}$ \\
      Bohachevsky  & $\mathbf{.0001 \pm {\scriptstyle .0002}}$ & $.002 \pm {\scriptstyle .004}$ & $.086 \pm {\scriptstyle .082}$
                   & $.005 \pm {\scriptstyle .006}$ & $\mathbf{.003 \pm {\scriptstyle .003}}$ & $.131 \pm {\scriptstyle .106}$
                   & $.002 \pm {\scriptstyle .003}$ & $.002 \pm {\scriptstyle .003}$ & $\mathbf{.0001 \pm {\scriptstyle .001}}$ \\
      Levy13       & $\mathbf{.001 \pm {\scriptstyle .001}}$ & $.014 \pm {\scriptstyle .043}$ & $.051 \pm {\scriptstyle .046}$
                   & $\mathbf{.006 \pm {\scriptstyle .004}}$ & $.007 \pm {\scriptstyle .005}$ & $.061 \pm {\scriptstyle .054}$
                   & $.003 \pm {\scriptstyle .001}$ & $.014 \pm {\scriptstyle .043}$ & $\mathbf{.001 \pm {\scriptstyle .001}}$ \\
      Bukin6       & $\mathbf{.053 \pm {\scriptstyle .042}}$ & $.078 \pm {\scriptstyle .076}$ & $.148 \pm {\scriptstyle .070}$
                   & $.170 \pm {\scriptstyle .091}$ & $\mathbf{.137 \pm {\scriptstyle .082}}$ & $.156 \pm {\scriptstyle .076}$
                   & $.107 \pm {\scriptstyle .092}$ & $.067 \pm {\scriptstyle .066}$ & $\mathbf{.033 \pm {\scriptstyle .023}}$ \\
      Cross-Tray   & $.040 \pm {\scriptstyle .043}$ & $\mathbf{.037 \pm {\scriptstyle .035}}$ & $.131 \pm {\scriptstyle .078}$
                   & $.084 \pm {\scriptstyle .079}$ & $\mathbf{.074 \pm {\scriptstyle .052}}$ & $.134 \pm {\scriptstyle .082}$
                   & $.051 \pm {\scriptstyle .067}$ & $.038 \pm {\scriptstyle .037}$ & $\mathbf{.029 \pm {\scriptstyle .035}}$ \\
      Ackley       & $\mathbf{.017 \pm {\scriptstyle .041}}$ & $.040 \pm {\scriptstyle .036}$ & $.227 \pm {\scriptstyle .103}$
                   & $.112 \pm {\scriptstyle .089}$ & $\mathbf{.112 \pm {\scriptstyle .054}}$ & $.248 \pm {\scriptstyle .099}$
                   & $.086 \pm {\scriptstyle .093}$ & $.039 \pm {\scriptstyle .034}$ & $\mathbf{.005 \pm {\scriptstyle .004}}$ \\
      \bottomrule
    \end{tabular}%
  }
\end{table}

\setlength{\tabcolsep}{4.7pt}
\renewcommand{\arraystretch}{1.1}
\begin{table}[!htbp]
  \centering
  \Large
  \caption{Inference regret under a fixed budget of 30 cost-units ($B=30$). Values are mean $\pm$ one standard deviation over 30 runs. Bold indicates the lowest mean regret within each function and cost scenario.}
  \vspace{0.5em}
  \label{tab:cost_inference-regret-30}
  \resizebox{\textwidth}{!}{%
    \begin{tabular}{l ccc|ccc|ccc}
      \toprule
      \multirow{2}{*}{\textbf{$\utility(\candidate)$}}
        & \multicolumn{3}{c|}{\textbf{$c_p\ll c_e$}}
        & \multicolumn{3}{c|}{\textbf{$c_p\approx c_e$}}
        & \multicolumn{3}{c}{\textbf{$c_p\gg c_e$}} \\
      \cmidrule(lr){2-4} \cmidrule(lr){5-7} \cmidrule(lr){8-10}
        & \textbf{Standard} & \textbf{Consecutive} & \textbf{Multiple}
        & \textbf{Standard} & \textbf{Consecutive} & \textbf{Multiple}
        & \textbf{Standard} & \textbf{Consecutive} & \textbf{Multiple} \\
      \midrule
      Branin        & $\mathbf{.009\pm{\scriptstyle .008}}$ & $.014\pm{\scriptstyle .015}$ & $.018\pm{\scriptstyle .013}$
                    & $.028\pm{\scriptstyle .015}$          & $\mathbf{.023\pm{\scriptstyle .017}}$ & $.026\pm{\scriptstyle .015}$
                    & $.024\pm{\scriptstyle .042}$          & $.015\pm{\scriptstyle .016}$          & $\mathbf{.005\pm{\scriptstyle .005}}$ \\
      Six-hump      & $\mathbf{.0016\pm{\scriptstyle .0015}}$ & $.0025\pm{\scriptstyle .0023}$ & $.0139\pm{\scriptstyle .0114}$
                    & $.0038\pm{\scriptstyle .0032}$ & $\mathbf{.0028\pm{\scriptstyle .0020}}$ & $.0143\pm{\scriptstyle .0127}$
                    & $.0044\pm{\scriptstyle .0040}$ & $.0032\pm{\scriptstyle .0025}$ & $\mathbf{.0009\pm{\scriptstyle .0006}}$ \\
      Bohachevsky   & $\mathbf{.002\pm{\scriptstyle .004}}$ & $.141\pm{\scriptstyle .135}$ & $.077\pm{\scriptstyle .089}$
                    & $.054\pm{\scriptstyle .067}$          & $\mathbf{.015\pm{\scriptstyle .012}}$ & $.109\pm{\scriptstyle .101}$
                    & $.117\pm{\scriptstyle .119}$          & $.172\pm{\scriptstyle .139}$          & $\mathbf{.001\pm{\scriptstyle .001}}$ \\
      Levy13        & $\mathbf{.002\pm{\scriptstyle .002}}$ & $.071\pm{\scriptstyle .142}$ & $.044\pm{\scriptstyle .052}$
                    & $\mathbf{.009\pm{\scriptstyle .007}}$ & $.011\pm{\scriptstyle .008}$          & $.052\pm{\scriptstyle .051}$
                    & $.061\pm{\scriptstyle .086}$          & $.056\pm{\scriptstyle .075}$          & $\mathbf{.024\pm{\scriptstyle .051}}$ \\
      Bukin6        & $\mathbf{.091\pm{\scriptstyle .044}}$ & $.232\pm{\scriptstyle .097}$ & $.192\pm{\scriptstyle .084}$
                    & $.282\pm{\scriptstyle .155}$          & $\mathbf{.251\pm{\scriptstyle .118}}$ & $.262\pm{\scriptstyle .114}$
                    & $.235\pm{\scriptstyle .115}$          & $.194\pm{\scriptstyle .110}$          & $\mathbf{.084\pm{\scriptstyle .029}}$ \\
      Cross-Tray & $\mathbf{.131\pm{\scriptstyle .097}}$ & $.142\pm{\scriptstyle .105}$ & $.196\pm{\scriptstyle .131}$
                    & $.161\pm{\scriptstyle .128}$          & $\mathbf{.156\pm{\scriptstyle .076}}$ & $.183\pm{\scriptstyle .080}$
                    & $.188\pm{\scriptstyle .102}$          & $.155\pm{\scriptstyle .110}$          & $\mathbf{.124\pm{\scriptstyle .107}}$ \\
      Ackley        & $\mathbf{.047\pm{\scriptstyle .044}}$ & $.152\pm{\scriptstyle .129}$ & $.203\pm{\scriptstyle .122}$
                    & $.163\pm{\scriptstyle .091}$          & $\mathbf{.132\pm{\scriptstyle .058}}$ & $.235\pm{\scriptstyle .102}$
                    & $.192\pm{\scriptstyle .141}$          & $.153\pm{\scriptstyle .131}$          & $\mathbf{.136\pm{\scriptstyle .114}}$ \\
      \bottomrule
    \end{tabular}%
  }
\end{table}


\subsection{Baselines and Adaptation to CPBO}

Here we explain how the comparison methods, EUBO and POP-BO, were used in the experiment reported in Section~\ref{sec:experiment_jnd}.


\paragraph{Oracle model.}
All baselines are trained using a common simulated oracle based on the three-outcome model, but adapted to produce binary labels. Specifically, if the noisy utility difference falls within the JND threshold $\gamma_{\text{true}}$, the oracle returns a random label in $\{-1, +1\}$:
\begin{equation}
\response_t =
\begin{cases}
+1 & \text{if } \Delta \utility_t + \delta_t > \gamma_{\text{true}}, \\
-1 & \text{if } \Delta \utility_t + \delta_t < -\gamma_{\text{true}}, \\
\sim \text{Uniform}(\{-1, +1\}) & \text{otherwise}.
\end{cases}  
\end{equation}
This ensures compatibility with baseline methods that assume binary feedback, while still capturing the ambiguity near the threshold.

\paragraph{EUBO Adaptation.}
We use the official \texttt{BoTorch} implementation of EUBO, modifying it to enforce the CPBO constraint of consecutive comparisons. Technically, we propose a single new candidate per iteration, and we condition the query on the candidate selected in the previous iteration. This restricts each query to compare the new candidate against the one produced in the prior step.

\paragraph{POP-BO Setup.}
POP-BO is inherently designed to operate with consecutive preference comparisons and requires no structural adaptation for the CPBO setting. However, its original implementation uses a binary preference oracle based on the Bradley-Terry model, where the probability of one configuration being preferred is given by $\sigma(f(\configuration) - f(\configuration'))$, with $\sigma(\cdot)$ denoting the logistic function. To ensure consistency across methods, we replace this with the Thurstone-style oracle described above.

In their original implementation, POP-BO assumes access to ground-truth function values in order to manually set kernel hyperparameters on a per-function basis. This is not applicable in real-world preference-based settings, where utility values are unobserved. As the original paper does not provide a method for estimating these parameters from preferences, and our own attempts proved unstable, we instead follow their published configuration and fix the hyperparameters for each benchmark.


\subsection{Additional Details for the UCB Acquisition Baselines}
\label{app:simple_acquisition_baselines}

Section~\ref{sec:jnd_acquisition_ablation} introduces the standard UCB and C-UCB baselines and reports their comparison with CPBO$_{\hat{\gamma}}$. Here we provide the remaining details of their construction.

For C-UCB, the posterior variance of the utility difference relative to the previous configuration $\configuration_t$ is
\begin{equation}
s_{\Delta,t}^2(\configuration)
=
s_t^2(\configuration)
+
s_t^2(\configuration_t)
-
2k_t(\configuration,\configuration_t),
\label{eq:cucb_variance}
\end{equation}
where $k_t(\configuration,\configuration_t)$ denotes the posterior covariance between the new and previous configurations. We use a fixed exploration coefficient $\beta=2$ for both UCB variants. In each case, the configuration maximizing the acquisition is produced and compared with the immediately previous configuration using the same consecutive evaluation protocol as CPBO$_{\hat{\gamma}}$.

Both baselines use the same GP surrogate construction, three-outcome JND likelihood, trainable threshold $\gamma$, and inference procedure as CPBO$_{\hat{\gamma}}$, while changing the acquisition rule. The experiments follow the same benchmark protocol with $\gamma_{\mathrm{true}}=0.04$, 30 optimization iterations, and matched initial configurations and random seeds across methods.

\subsection{Statistical Analysis of the Main Benchmark Results}
\label{app:statistical_analysis}

We compare CPBO$_{\hat{\gamma}}$ with each available comparison method using paired two-sided Wilcoxon signed-rank tests on the 30 matched runs. For each benchmark and metric, we then use an intersection-union test to assess whether CPBO$_{\hat{\gamma}}$ differs from all available comparison methods. The corresponding null hypothesis is that at least one comparison method does not differ from CPBO$_{\hat{\gamma}}$, and alternative is that CPBO$_{\hat{\gamma}}$ differs from every available comparison method. Since rejection requires all component pairwise hypotheses to be rejected, the benchmark-level $p$-value is $p_{\mathrm{IUT}}=\max_j p_j$, where $p_j$ denotes the two-sided Wilcoxon $p$-value for comparison method $j$. Tables~\ref{tab:statistical_regret} and~\ref{tab:statistical_ordinal} report pairwise $p$-values, matched-pairs rank-biserial correlations, and mean paired differences together with pointwise 95\% percentile confidence intervals computed from 100,000 paired bootstrap resamples. Statistical significance is assessed at $p_{\mathrm{IUT}}<0.05$. For both metrics, all mean paired differences and rank-biserial correlations favor CPBO$_{\hat{\gamma}}$, and intersection-union test is significant for every benchmark except Cross-Tray.

\begin{table*}[t]
\centering
\small
\caption{Paired statistical comparisons of inference regret between CPBO$_{\hat{\gamma}}$ and each available comparison method. $\Delta R$ is the mean paired regret difference, computed as comparison method minus CPBO$_{\hat{\gamma}}$, so positive values favor CPBO$_{\hat{\gamma}}$. Brackets report pointwise 95\% percentile paired-bootstrap confidence intervals, and positive $r_{\mathrm{rb}}$ values favor CPBO$_{\hat{\gamma}}$. The table reports the two-sided Wilcoxon signed-rank $p$-value for each pairwise comparison and the benchmark-level intersection-union $p$-value, $p_{\mathrm{IUT}}$, defined as the maximum pairwise $p$-value among the available comparison methods.}
\label{tab:statistical_regret}
\vspace{0.5em}

\setlength{\tabcolsep}{6.0pt}
\renewcommand{\arraystretch}{1.08}

\begin{tabular}{llcccc}
\toprule
\textbf{Benchmark} & \textbf{Comparison} &
$\boldsymbol{\Delta R}$ [95\% CI] &
$\boldsymbol{r_{\mathrm{rb}}}$ &
$\boldsymbol{p}$ &
$\boldsymbol{p_{\mathrm{IUT}}}$ \\
\midrule
Branin & POP-BO & .0100 [.0015, .0187] & .555 & .00711 & .00711 \\
Branin & EUBO & .0140 [.0047, .0248] & .572 & .00538 & .00711 \\
Branin & CPBO$_{\hat{\gamma}=0}$ & .0110 [.0048, .0176] & .618 & .00426 & .00711 \\
\addlinespace
Six-hump & EUBO & .0031 [.0009, .0055] & .450 & .04046 & .04046 \\
Six-hump & CPBO$_{\hat{\gamma}=0}$ & .0051 [.0011, .0094] & .473 & .03498 & .04046 \\
\addlinespace
Bohachevsky & EUBO & .0025 [.0009, .0043] & .570 & .01064 & .02244 \\
Bohachevsky & CPBO$_{\hat{\gamma}=0}$ & .0027 [.0009, .0047] & .497 & .02244 & .02244 \\
\addlinespace
Levy13 & POP-BO & .0110 [.0026, .0193] & .467 & .02816 & .03643 \\
Levy13 & EUBO & .0050 [-.0015, .0110] & .439 & .03643 & .03643 \\
Levy13 & CPBO$_{\hat{\gamma}=0}$ & .0180 [.0049, .0322] & .469 & .02479 & .03643 \\
\addlinespace
Bukin6 & POP-BO & .0670 [.0214, .1172] & .460 & .02774 & .03643 \\
Bukin6 & EUBO & .0110 [-.0114, .0305] & .451 & .03408 & .03643 \\
Bukin6 & CPBO$_{\hat{\gamma}=0}$ & .0400 [.0076, .0729] & .439 & .03643 & .03643 \\
\addlinespace
Cross-Tray & POP-BO & .0659 [.0224, .1115] & .454 & .02933 & .22065 \\
Cross-Tray & EUBO & .0252 [-.0097, .0607] & .329 & .11908 & .22065 \\
Cross-Tray & CPBO$_{\hat{\gamma}=0}$ & .0263 [-.0139, .0695] & .260 & .22065 & .22065 \\
\addlinespace
Ackley & EUBO & .0559 [.0172, .1009] & .542 & .00871 & .00871 \\
Ackley & CPBO$_{\hat{\gamma}=0}$ & .0699 [.0320, .1095] & .624 & .00219 & .00871 \\
\bottomrule
\end{tabular}
\end{table*}

\begin{table*}[t]
\centering
\small
\caption{Paired statistical comparisons of ordinal accuracy between CPBO$_{\hat{\gamma}}$ and each available comparison method. $\Delta A$ is the mean paired difference in percentage points, computed as CPBO$_{\hat{\gamma}}$ minus the comparison method, so positive values favor CPBO$_{\hat{\gamma}}$. Brackets report pointwise 95\% percentile paired-bootstrap confidence intervals, and positive $r_{\mathrm{rb}}$ values favor CPBO$_{\hat{\gamma}}$. The table reports the two-sided Wilcoxon signed-rank $p$-value for each pairwise comparison and the benchmark-level intersection-union $p$-value, $p_{\mathrm{IUT}}$, defined as the maximum pairwise $p$-value among the available comparison methods.}
\label{tab:statistical_ordinal}
\vspace{0.5em}

\setlength{\tabcolsep}{6.0pt}
\renewcommand{\arraystretch}{1.08}

\begin{tabular}{llcccc}
\toprule
\textbf{Benchmark} & \textbf{Comparison} &
$\boldsymbol{\Delta A}$ [95\% CI] &
$\boldsymbol{r_{\mathrm{rb}}}$ &
$\boldsymbol{p}$ &
$\boldsymbol{p_{\mathrm{IUT}}}$ \\
\midrule
Branin & POP-BO & 8.71 [6.94, 10.48] & .989 & $9.31\times10^{-9}$ & $3.51\times10^{-6}$ \\
Branin & EUBO & 21.10 [18.69, 23.54] & 1.000 & $1.86\times10^{-9}$ & $3.51\times10^{-6}$ \\
Branin & CPBO$_{\hat{\gamma}=0}$ & 8.21 [6.49, 9.87] & .986 & $3.51\times10^{-6}$ & $3.51\times10^{-6}$ \\
\addlinespace
Six-hump & EUBO & 11.89 [8.97, 14.89] & .966 & $4.66\times10^{-8}$ & $1.06\times10^{-5}$ \\
Six-hump & CPBO$_{\hat{\gamma}=0}$ & 6.19 [3.99, 8.43] & .841 & $1.06\times10^{-5}$ & $1.06\times10^{-5}$ \\
\addlinespace
Bohachevsky & EUBO & 21.61 [19.20, 24.09] & 1.000 & $1.86\times10^{-9}$ & $5.59\times10^{-9}$ \\
Bohachevsky & CPBO$_{\hat{\gamma}=0}$ & 7.10 [5.99, 8.14] & .991 & $5.59\times10^{-9}$ & $5.59\times10^{-9}$ \\
\addlinespace
Levy13 & POP-BO & 16.91 [14.57, 19.03] & .996 & $3.73\times10^{-9}$ & $7.06\times10^{-5}$ \\
Levy13 & EUBO & 10.40 [7.41, 13.42] & .927 & $3.86\times10^{-7}$ & $7.06\times10^{-5}$ \\
Levy13 & CPBO$_{\hat{\gamma}=0}$ & 5.50 [3.31, 7.63] & .776 & $7.06\times10^{-5}$ & $7.06\times10^{-5}$ \\
\addlinespace
Bukin6 & POP-BO & 17.59 [16.02, 19.22] & 1.000 & $1.86\times10^{-9}$ & $1.86\times10^{-8}$ \\
Bukin6 & EUBO & 18.69 [16.96, 20.33] & 1.000 & $1.86\times10^{-9}$ & $1.86\times10^{-8}$ \\
Bukin6 & CPBO$_{\hat{\gamma}=0}$ & 5.49 [4.41, 6.51] & .978 & $1.86\times10^{-8}$ & $1.86\times10^{-8}$ \\
\addlinespace
Cross-Tray & POP-BO & 1.40 [-.49, 3.10] & .432 & .04207 & .15795 \\
Cross-Tray & EUBO & 8.10 [3.56, 11.84] & .366 & .08407 & .15795 \\
Cross-Tray & CPBO$_{\hat{\gamma}=0}$ & 4.20 [-.69, 8.65] & .299 & .15795 & .15795 \\
\addlinespace
Ackley & EUBO & 15.80 [12.16, 19.53] & .991 & $3.17\times10^{-6}$ & $3.17\times10^{-6}$ \\
Ackley & CPBO$_{\hat{\gamma}=0}$ & 8.80 [6.73, 10.77] & .957 & $8.01\times10^{-8}$ & $3.17\times10^{-6}$ \\
\bottomrule
\end{tabular}
\end{table*}


\subsection{CPBO Benchmarks}
\label{cpbo_benchmarks}

We extend the main paper results by evaluating CPBO under a broader range of indifference levels to examine its robustness and generality. Specifically, we test how performance varies as the true JND threshold $\gamma_{\text{true}}$ and the expected proportion of indifferences $\mathbb{E}[\sim]$ change across benchmark functions.

We consider two complementary setups:
\begin{itemize}
    \setlength{\itemsep}{0pt}
    \setlength{\parsep}{0pt}
    \setlength{\topsep}{2pt}
    \setlength{\partopsep}{0pt}

    \item Fixed threshold $\gamma_{\text{true}} = 0.1$, corresponding to approximately 25--50\% expected indifferences $\mathbb{E}[\sim]$.
    \item Fixed $\mathbb{E}[\sim]$ values of 15\% and 30\%, with per-function $\gamma_{\text{true}}$ chosen accordingly.
\end{itemize}
\vspace{-0.5em}

Tables~\ref{tab:combined_results_gamma_0.1} to~\ref{tab:combined_results_fixed_tie_15} report inference regret and ordinal accuracy across these settings. CPBO with learned $\hat{\gamma}$ achieves the strongest overall performance and remains competitive across all benchmarks, including against a variant with fixed $\hat{\gamma}=0$, which highlights the benefit of explicitly modeling perceptual ambiguity. Overall, CPBO demonstrates stable optimization, accurate utility recovery, and robustness across a wide range of $\mathbb{E}[\sim]$ levels.

\begin{table}[!htbp]
\centering
\small
\caption{Optimization and exploration performance measured by inference regret (Regret) and global ordinal accuracy of the utility proxy (Ord.). Results (mean $\pm $ sd over 30 runs) after 30 iterations under $\gamma_{\text{true}}=0.1$. Bold indicates the best mean performance for each benchmark and metric.}
\vspace{0.5em}
\setlength{\tabcolsep}{5.7pt}
\renewcommand{\arraystretch}{1.1}
\begin{tabular}{lccccccccc}
\toprule
\multirow{2}{*}{$\utility(\candidate)$} 
  & \multirow{2}{*}{$\mathbb{E}[\sim]$} 
  & \multicolumn{2}{c}{\textbf{POP-BO}} 
  & \multicolumn{2}{c}{\textbf{EUBO}} 
  & \multicolumn{2}{c}{\textbf{CPBO$_{\hat \gamma=0}$}} 
  & \multicolumn{2}{c}{\textbf{CPBO$_{\hat \gamma}$}} \\
\cmidrule(lr){3-4} \cmidrule(lr){5-6} \cmidrule(lr){7-8} \cmidrule(lr){9-10}
 &  & Regret$\downarrow$ & Ord.$\uparrow$ & Regret$\downarrow$ & Ord.$\uparrow$ & Regret$\downarrow$ & Ord.$\uparrow$ & Regret$\downarrow$ & Ord.$\uparrow$ \\
\midrule
Branin         
 & $43\%$ & $.042 \pm {\scriptstyle .035}$ & 75.8 & $.044 \pm {\scriptstyle .023}$ & 68.2 & $.046 \pm {\scriptstyle .016}$ & 77.9 & $\mathbf{.021} \pm {\scriptstyle \mathbf{.017}}$ & \textbf{89.5} \\
Six-hump 
 & $54\%$ & NA & NA & $.029 \pm {\scriptstyle .017}$ & 68.2 & $.035 \pm {\scriptstyle .022}$ & 73.1 & $\mathbf{.012} \pm {\scriptstyle \mathbf{.007}}$ & \textbf{83.1} \\
Bohachevsky    
 & $25\%$ & NA & NA & $.021 \pm {\scriptstyle .018}$ & 73.6 & $.024 \pm {\scriptstyle .041}$ & 89.2 & $\mathbf{.001} \pm {\scriptstyle \mathbf{.002}}$ & \textbf{98.6} \\
\hdashline
Levy13         
 & $39\%$ & $.059 \pm {\scriptstyle .044}$ & 67.1 & $.018 \pm {\scriptstyle .019}$ & 75.7 & $.056 \pm {\scriptstyle .057}$ & 80.1 & $\mathbf{.007} \pm {\scriptstyle \mathbf{.007}}$ & \textbf{87.4} \\
Bukin6         
 & $25\%$ & $.204 \pm {\scriptstyle .086}$ & 77.9 & $\mathbf{.182} \pm {\scriptstyle \mathbf{.110}}$ & 76.4 & $.266 \pm {\scriptstyle .105}$ & 88.3 & $.184 \pm {\scriptstyle .101}$ & \textbf{95.3} \\
Cross-Tray  
 & $47\%$ & $.245 \pm {\scriptstyle .122}$ & 63.8 & $.198 \pm {\scriptstyle .129}$ & 60.7 & $.218 \pm {\scriptstyle .104}$ & 64.0 & $\mathbf{.152} \pm {\scriptstyle \mathbf{.074}}$ & \textbf{69.7} \\
Ackley         
 & $48\%$ & NA & NA & $.157 \pm {\scriptstyle .117}$ & 72.8 & $.238 \pm {\scriptstyle .128}$ & 77.3 & $\mathbf{.122} \pm {\scriptstyle \mathbf{.059}}$ & \textbf{87.9} \\
\bottomrule
\end{tabular}
\label{tab:combined_results_gamma_0.1}
\end{table}

\begin{table}[!htbp]
\centering
\small
\caption{Optimization and exploration performance measured by inference regret (Regret) and global ordinal accuracy of the utility proxy (Ord.). Results (mean $\pm $ sd over 30 runs) after 30 iterations with $\mathbb{E}[\sim]=30\%$. Bold indicates the best mean performance for each benchmark and metric.}
\vspace{0.5em}
\setlength{\tabcolsep}{5.7pt}
\renewcommand{\arraystretch}{1.1}
\begin{tabular}{lccccccccc}
\toprule
\multirow{2}{*}{$\utility(\candidate)$} 
  & \multirow{2}{*}{$\gamma_{\text{true}}$} 
  & \multicolumn{2}{c}{\textbf{POP-BO}} 
  & \multicolumn{2}{c}{\textbf{EUBO}} 
  & \multicolumn{2}{c}{\textbf{CPBO$_{\hat \gamma=0}$}} 
  & \multicolumn{2}{c}{\textbf{CPBO$_{\hat \gamma}$}} \\
\cmidrule(lr){3-4} \cmidrule(lr){5-6} \cmidrule(lr){7-8} \cmidrule(lr){9-10}
 &  & Regret$\downarrow$ & Ord.$\uparrow$ & Regret$\downarrow$ & Ord.$\uparrow$ & Regret$\downarrow$ & Ord.$\uparrow$ & Regret$\downarrow$ & Ord.$\uparrow$ \\
\midrule
Branin         
 & $.061$ & $.031 \pm {\scriptstyle .018}$ & 77.3 & $.042 \pm {\scriptstyle .020}$ & 67.8 & $.033 \pm {\scriptstyle .024}$ & 80.4 & $\mathbf{.020} \pm {\scriptstyle \mathbf{.014}}$ & \textbf{90.3} \\
Six-hump 
 & $.033$ & NA & NA & $.013 \pm {\scriptstyle .009}$ & 73.0 & $.014 \pm {\scriptstyle .010}$ & 78.0 & $\mathbf{.010} \pm {\scriptstyle \mathbf{.004}}$ & \textbf{84.0} \\
Bohachevsky    
 & $.120$ & NA & NA & $.022 \pm {\scriptstyle .018}$ & 74.0 & $.028 \pm {\scriptstyle .033}$ & 88.6 & $\mathbf{.001} \pm {\scriptstyle \mathbf{.002}}$ & \textbf{97.9} \\
\hdashline
Levy13         
 & $.074$ & $.035 \pm {\scriptstyle .022}$ & 69.1 & $.017 \pm {\scriptstyle .012}$ & 74.9 & $.046 \pm {\scriptstyle .053}$ & 80.4 & $\mathbf{.010} \pm {\scriptstyle \mathbf{.008}}$ & \textbf{88.6} \\
Bukin6         
 & $.122$ & $.208 \pm {\scriptstyle .103}$ & 76.4 & $.202 \pm {\scriptstyle .103}$ & 77.1 & $.272 \pm {\scriptstyle .104}$ & 87.7 & $\mathbf{.188} \pm {\scriptstyle \mathbf{.092}}$ & \textbf{94.7} \\
Cross-Tray  
 & $.062$ & $.219 \pm {\scriptstyle .103}$ & 66.1 & $.191 \pm {\scriptstyle .124}$ & 61.1 & $.194 \pm {\scriptstyle .096}$ & 63.9 & $\mathbf{.138} \pm {\scriptstyle \mathbf{.071}}$ & \textbf{67.7} \\
Ackley         
 & $.059$ & NA & NA & $.158 \pm {\scriptstyle .107}$ & 75.0 & $.174 \pm {\scriptstyle .072}$ & 77.9 & $\mathbf{.110} \pm {\scriptstyle \mathbf{.055}}$ & \textbf{89.1} \\
\bottomrule
\end{tabular}
\label{tab:combined_results_fixed_tie_30}
\end{table}

\begin{table}[!htbp]
\centering
\small
\caption{Optimization and exploration performance measured by inference regret (Regret) and global ordinal accuracy of the utility proxy (Ord.). Results (mean $\pm $ sd over 30 runs) after 30 iterations with $\mathbb{E}[\sim]=15\%$. Bold indicates the best mean performance for each benchmark and metric.}
\vspace{0.5em}
\setlength{\tabcolsep}{5.7pt}
\renewcommand{\arraystretch}{1.1}
\begin{tabular}{lccccccccc}
\toprule
\multirow{2}{*}{$\utility(\candidate)$} 
  & \multirow{2}{*}{$\gamma_{\text{true}}$} 
  & \multicolumn{2}{c}{\textbf{POP-BO}} 
  & \multicolumn{2}{c}{\textbf{EUBO}} 
  & \multicolumn{2}{c}{\textbf{CPBO$_{\hat \gamma=0}$}} 
  & \multicolumn{2}{c}{\textbf{CPBO$_{\hat \gamma}$}} \\
\cmidrule(lr){3-4} \cmidrule(lr){5-6} \cmidrule(lr){7-8} \cmidrule(lr){9-10}
 &  & Regret$\downarrow$ & Ord.$\uparrow$ & Regret$\downarrow$ & Ord.$\uparrow$ & Regret$\downarrow$ & Ord.$\uparrow$ & Regret$\downarrow$ & Ord.$\uparrow$ \\
\midrule
Branin         
 & $.0279$ & $.025 \pm {\scriptstyle .019}$ & 81.3 & $\mathbf{.018} \pm {\scriptstyle \mathbf{.013}}$ & 70.2 & $.026 \pm {\scriptstyle .017}$ & 81.2 & $.020 \pm {\scriptstyle .017}$ & \textbf{90.1} \\
Six-hump 
 & $.0103$ & NA & NA & $\mathbf{.006} \pm {\scriptstyle \mathbf{.004}}$ & 73.2 & $.012 \pm {\scriptstyle .008}$ & 78.3 & $.008 \pm {\scriptstyle .004}$ & \textbf{86.2} \\
Bohachevsky    
 & $.0578$ & NA & NA & $.011 \pm {\scriptstyle .012}$ & 77.5 & $.009 \pm {\scriptstyle .018}$ & 90.6 & $\mathbf{.001} \pm {\scriptstyle \mathbf{.001}}$ & \textbf{98.7} \\
\hdashline
Levy13         
 & $.0355$ & $.016 \pm {\scriptstyle .012}$ & 70.9 & $.015 \pm {\scriptstyle .011}$ & 75.8 & $.028 \pm {\scriptstyle .029}$ & 82.5 & $\mathbf{.008} \pm {\scriptstyle \mathbf{.009}}$ & \textbf{87.1} \\
Bukin6         
 & $.0596$ & $.195 \pm {\scriptstyle .115}$ & 78.7 & $.171 \pm {\scriptstyle .114}$ & 75.3 & $.166 \pm {\scriptstyle .080}$ & 90.1 & $\mathbf{.159} \pm {\scriptstyle \mathbf{.085}}$ & \textbf{96.3} \\
Cross-Tray  
 & $.0302$ & $.196 \pm {\scriptstyle .088}$ & 67.6 & $.160 \pm {\scriptstyle .091}$ & 63.1 & $.166 \pm {\scriptstyle .081}$ & 64.3 & $\mathbf{.153} \pm {\scriptstyle \mathbf{.076}}$ & \textbf{68.3} \\
Ackley         
 & $.0287$ & NA & NA & $.122 \pm {\scriptstyle .058}$ & 78.1 & $.126 \pm {\scriptstyle .068}$ & 81.2 & $\mathbf{.106} \pm {\scriptstyle \mathbf{.047}}$ & \textbf{85.8} \\
\bottomrule
\end{tabular}
\label{tab:combined_results_fixed_tie_15}
\end{table}


\paragraph{Utility Recovery and Choice Accuracy.}

In addition to ordinal accuracy, we also propose a \emph{choice accuracy} metric for evaluating the exploration performance of the PBO method. It measures how often the model correctly predicts the outcome of a uniformly random pairwise comparison. Each comparison can have three possible outcomes: better, worse, or indifferent, depending on whether the true utility difference exceeds the JND threshold $\gamma_{\text{true}}$. While ordinal accuracy ignores indifference and only assesses the relative ranking between configurations, choice accuracy evaluates all three outcomes by incorporating the learned JND threshold $\hat{\gamma}$.

For baseline models that do not account for perceptual uncertainty, predictions falling within the true JND band are always treated as incorrect, since such models cannot represent indifference. Table~\ref{tab:accuracy_comparison_fixed_gamma_0.04_full} reports both ordinal and choice accuracy under $\gamma_{\text{true}} = 0.04$, complementing Table~\ref{tab:combined_results} in the main paper by including the corresponding standard deviations. Together, these results show how accurately each method recovers the underlying utility and predicts human-like preference behavior.

\begin{table}[H]
\setlength{\tabcolsep}{3.7pt}
\renewcommand{\arraystretch}{1.1}
\centering
\caption{Latent-space prediction ordinal (Ord.) and choice accuracy (\%) at iteration 30 under $\gamma_{\text{true}} = 0.04$, reported as mean $\pm$ sd over 30 runs. Bold indicates the best mean performance for each benchmark and metric.}
\vspace{0.5em}
\begin{tabular}{lcccccccc}
\toprule
\multirow{2}{*}{$\utility(\candidate)$}
  & \multicolumn{2}{c}{\textbf{POP-BO}}
  & \multicolumn{2}{c}{\textbf{EUBO}}
  & \multicolumn{2}{c}{\textbf{CPBO$_{\hat \gamma=0}$}}
  & \multicolumn{2}{c}{\textbf{CPBO$_{\hat \gamma}$}} \\
\cmidrule(lr){2-3} \cmidrule(lr){4-5} \cmidrule(lr){6-7} \cmidrule(lr){8-9}
 & \textbf{Ord.} & \textbf{Choice}
 & \textbf{Ord.} & \textbf{Choice}
 & \textbf{Ord.} & \textbf{Choice}
 & \textbf{Ord.} & \textbf{Choice} \\
\midrule
Branin        & 81.5 $\pm$ {\scriptsize 4.2} & 72.2 $\pm$ {\scriptsize 4.3} & 69.1 $\pm$ {\scriptsize 6.3} & 58.9 $\pm$ {\scriptsize 6.3} & 82.0 $\pm$ {\scriptsize 3.6} & 71.3 $\pm$ {\scriptsize 3.7} & \textbf{90.2 $\pm$ {\scriptsize 2.3}} & \textbf{77.9 $\pm$ {\scriptsize 2.1}} \\
Six-hump      & NA & NA & 72.4 $\pm$ {\scriptsize 5.9} & 63.8 $\pm$ {\scriptsize 6.0} & 78.1 $\pm$ {\scriptsize 3.8} & 68.7 $\pm$ {\scriptsize 3.9} & \textbf{84.3 $\pm$ {\scriptsize 2.5}} & \textbf{73.8 $\pm$ {\scriptsize 2.7}} \\
Bohachevsky   & NA & NA & 77.2 $\pm$ {\scriptsize 7.3} & 69.5 $\pm$ {\scriptsize 6.8} & 91.7 $\pm$ {\scriptsize 2.1} & 83.2 $\pm$ {\scriptsize 2.5} & \textbf{98.8 $\pm$ {\scriptsize 1.0}} & \textbf{91.5 $\pm$ {\scriptsize 1.0}} \\
\hdashline
Levy13        & 70.5 $\pm$ {\scriptsize 4.1} & 64.3 $\pm$ {\scriptsize 4.1} & 77.0 $\pm$ {\scriptsize 6.1} & 66.6 $\pm$ {\scriptsize 6.3} & 81.9 $\pm$ {\scriptsize 3.7} & 70.3 $\pm$ {\scriptsize 3.9} & \textbf{87.4 $\pm$ {\scriptsize 2.5}} & \textbf{76.9 $\pm$ {\scriptsize 2.2}} \\
Bukin6        & 79.4 $\pm$ {\scriptsize 4.5} & 74.8 $\pm$ {\scriptsize 4.4} & 78.3 $\pm$ {\scriptsize 5.1} & 70.9 $\pm$ {\scriptsize 5.2} & 91.5 $\pm$ {\scriptsize 2.0} & 82.8 $\pm$ {\scriptsize 2.3} & \textbf{97.0 $\pm$ {\scriptsize 1.0}} & \textbf{90.0 $\pm$ {\scriptsize 1.0}} \\
Cross-Tray & 66.9 $\pm$ {\scriptsize 3.4} & \textbf{60.0 $\pm$ {\scriptsize 3.5}} & 60.2 $\pm$ {\scriptsize 5.9} & 50.1 $\pm$ {\scriptsize 6.1} & 64.1 $\pm$ {\scriptsize 6.9} & 53.9 $\pm$ {\scriptsize 6.8} & \textbf{68.3 $\pm$ {\scriptsize 6.4}} & 56.6 $\pm$ {\scriptsize 6.5} \\
Ackley        & NA & NA & 72.9 $\pm$ {\scriptsize 8.2} & 62.4 $\pm$ {\scriptsize 7.6} & 79.9 $\pm$ {\scriptsize 3.4} & 70.3 $\pm$ {\scriptsize 3.8} & \textbf{88.7 $\pm$ {\scriptsize 2.4}} & \textbf{74.9 $\pm$ {\scriptsize 2.0}} \\
\bottomrule
\end{tabular}
\label{tab:accuracy_comparison_fixed_gamma_0.04_full}
\end{table}

\begin{figure}[H]
    \centering
    \includegraphics[width=1.00\textwidth]{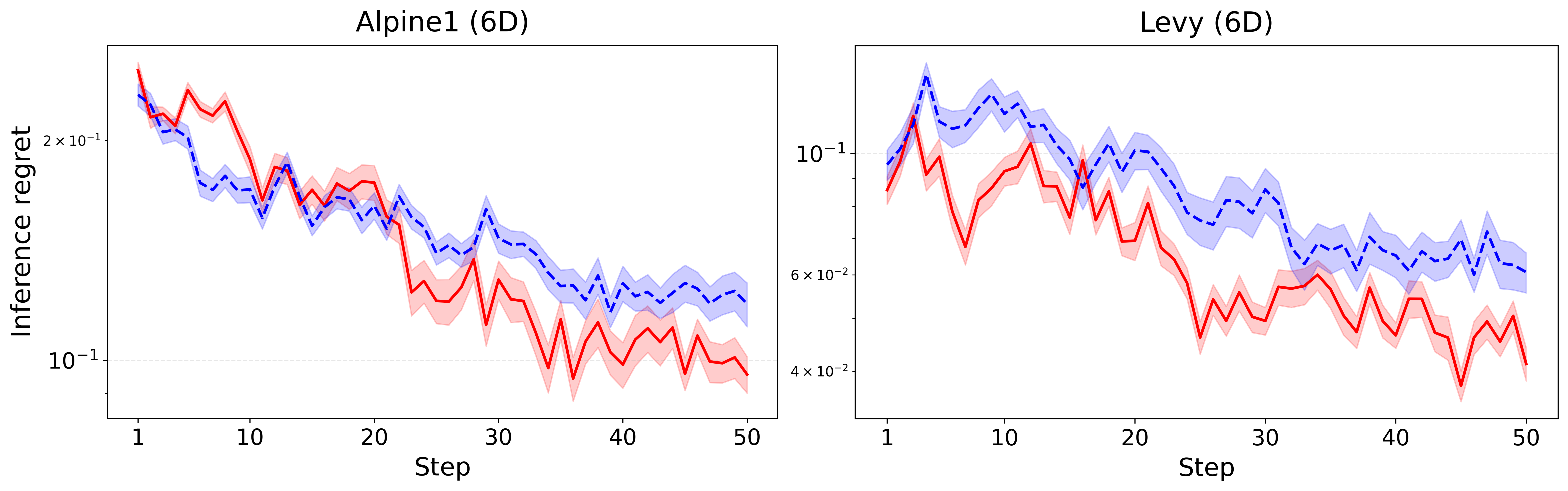}
    \caption{\emph{Inference regret} as a function of steps under $\gamma_{\text{true}} = 0.04$. (Left) 6D Alpine1 function; (Right) 6D Levy function. Solid line shows the mean and shaded regions the standard error of the mean over 30 runs.
    }
    \vspace{0.5em}
    \label{fig:6d_benchmarks}
\end{figure}

\paragraph{Higher-Dimensional Benchmarks (6D and 12D).} 
All synthetic experiments in the main paper were for 2-dimensional problems, simply for ease of visualization. The method itself naturally extends to higher dimensionalities.
Figure~\ref{fig:6d_benchmarks} shows this in a problem of moderate dimensionality, using two 6-dimensional standard optimization functions: Alpine1 6D and Levy 6D, in comparison to EUBO. All experimental protocols, including the oracle model, CPBO configuration, and EUBO adaptation (run in a consecutive comparison mode), follow the same setup as in the 2D benchmark experiments. The utility values are normalized to $[0, 1]$, and the perceptual noise used in the oracle is fixed to $\gamma_{\text{true}} = 0.04$. The results are consistent with those reported for the 2D experiments in the main paper, with CPBO clearly outperforming the comparison methods.

To further assess scalability, we additionally consider a higher-dimensional Alpine1 benchmark in 12D and extend the evaluation horizon up to 100 iterations. The experimental setup is identical to the 2D benchmarks, with the only difference being a slightly increased computational budget for acquisition optimization (35 optimization steps instead of 25) to account for the higher-dimensional search space. The results, reported in Table~\ref{tab:12d_alpine_long_horizon}, show that the advantage of CPBO persists both in higher dimensions and over longer optimization horizons.

\begin{table}[tb]
\centering
\caption{Inference regret (mean $\pm$ sd over 30 runs) on Alpine1 in 12D for increasing optimization horizons.}
\vspace{0.4em}
\label{tab:12d_alpine_long_horizon}
\begin{tabular}{lcccc}
\toprule
\textbf{Method} & \textbf{30 steps} & \textbf{50 steps} & \textbf{75 steps} & \textbf{100 steps} \\
\midrule
CPBO & \textbf{.158 $\pm$ .081} & \textbf{.077 $\pm$ .053} & \textbf{.047 $\pm$ .039} & \textbf{.022 $\pm$ .019} \\
EUBO & .262 $\pm$ .076 & .251 $\pm$ .075 & .236 $\pm$ .068 & .214 $\pm$ .060 \\
\bottomrule
\end{tabular}
\end{table}

\paragraph{Scaling with Dimensionality.} 
The computational complexity of CPBO scales with dimensionality in the same way as the original MES, which has been demonstrated to extend to tens of dimensions. For practical human-in-the-loop scenarios, wall-clock time per iteration is the relevant measure. In the real-world extrusion experiment (Section~\ref{sec:experiment_extrusion}), our configuration resulted in an average of about five minutes per iteration, including visualization and plotting overhead, which conveniently matches the time required to produce a new sample with the machine. 

To illustrate scalability, we report per-iteration wall-clock times (mean $\pm$ sd over 15 runs) for the Alpine1 benchmark across varying dimensionalities, keeping the same Gumbel sampling and training parameters, with slightly more optimization trials for the acquisition maximization. Results are shown in Table~\ref{tab:wallclock_scaling}, indicating feasible runtimes up to $d=12$.

\begin{table}[htbp]
\centering
\caption{Wall-clock time per BO iteration (seconds) as a function of dimensionality, averaged over 15 runs on Alpine1.}
\vspace{0.5em}
\label{tab:wallclock_scaling}
\begin{tabular}{lccccc}
\toprule
\textbf{Dimension} & d=2 & d=3 & d=6 & d=10 & d=12 \\
\midrule
Wall-clock (s) & 193$\pm$19 & 229$\pm$22 & 345$\pm$34 & 510$\pm$43 & 626$\pm$50 \\
\bottomrule
\end{tabular}
\end{table}

\paragraph{Sensitivity to perceptual-noise scale $\sigma$.}
To assess how misspecifying the perceptual-noise scale affects CPBO, we re-ran all seven benchmarks while fixing $\sigma$ to $0.25\times$ and $4\times$ its default value. Table~\ref{tab:sigma_sensitivity} shows that inference regret after 30 iterations is essentially unchanged across these settings. The middle column ($\sigma$) corresponds to the default setting reported in Table~\ref{tab:combined_results} in the main paper; the other two columns replicate that table under misspecified noise.

To quantify this sensitivity, for each benchmark we compared the default setting against $0.25\sigma$ and $4\sigma$ using paired two-sided Wilcoxon signed-rank tests on matched runs. The raw $p$-values ranged from $0.245$ to $0.716$, with none below $0.05$. Together with the similar mean regrets in Table~\ref{tab:sigma_sensitivity}, these results suggest that CPBO is relatively robust to the choice of $\sigma$ over the tested range. When $\sigma$ is under- or overestimated, other trainable model quantities, notably the learned JND threshold $\hat\gamma$, can partially adapt to the mismatch, while the resulting acquisition behavior and final performance remain similar over the tested range.

\begin{table}[htbp]
\centering
\caption{Sensitivity of CPBO$_{\hat\gamma}$ to the perceptual-noise scale $\sigma$: inference regret (mean $\pm$ sd over 30 runs) after 30 iterations with $\sigma$ fixed to $0.25\times$, $1\times$ (original, as in Table~\ref{tab:combined_results}), and $4\times$.}
\vspace{0.5em}
\label{tab:sigma_sensitivity}
\begin{tabular}{lccc}
\toprule
\textbf{$\utility(\candidate)$} & \textbf{$0.25\sigma$} & \textbf{$\sigma$} & \textbf{$4\sigma$} \\
\midrule
Branin        & .022 $\pm$ {\scriptsize .021} & .020 $\pm$ {\scriptsize .017} & .019 $\pm$ {\scriptsize .017} \\
Six-hump      & .008 $\pm$ {\scriptsize .006} & .009 $\pm$ {\scriptsize .004} & .012 $\pm$ {\scriptsize .007} \\
Bohachevsky   & .001 $\pm$ {\scriptsize .002} & .001 $\pm$ {\scriptsize .001} & .001 $\pm$ {\scriptsize .001} \\
Levy13        & .012 $\pm$ {\scriptsize .014} & .011 $\pm$ {\scriptsize .016} & .009 $\pm$ {\scriptsize .012} \\
Bukin6        & .131 $\pm$ {\scriptsize .048} & .115 $\pm$ {\scriptsize .053} & .119 $\pm$ {\scriptsize .052} \\
Cross-Tray    & .143 $\pm$ {\scriptsize .069} & .150 $\pm$ {\scriptsize .064} & .146 $\pm$ {\scriptsize .067} \\
Ackley        & .113 $\pm$ {\scriptsize .048} & .093 $\pm$ {\scriptsize .052} & .115 $\pm$ {\scriptsize .048} \\
\bottomrule
\end{tabular}
\end{table}

\paragraph{Sensitivity to oracle-feedback noise.}

\begin{figure}[htbp]
    \centering
    \includegraphics[width=0.65\textwidth]{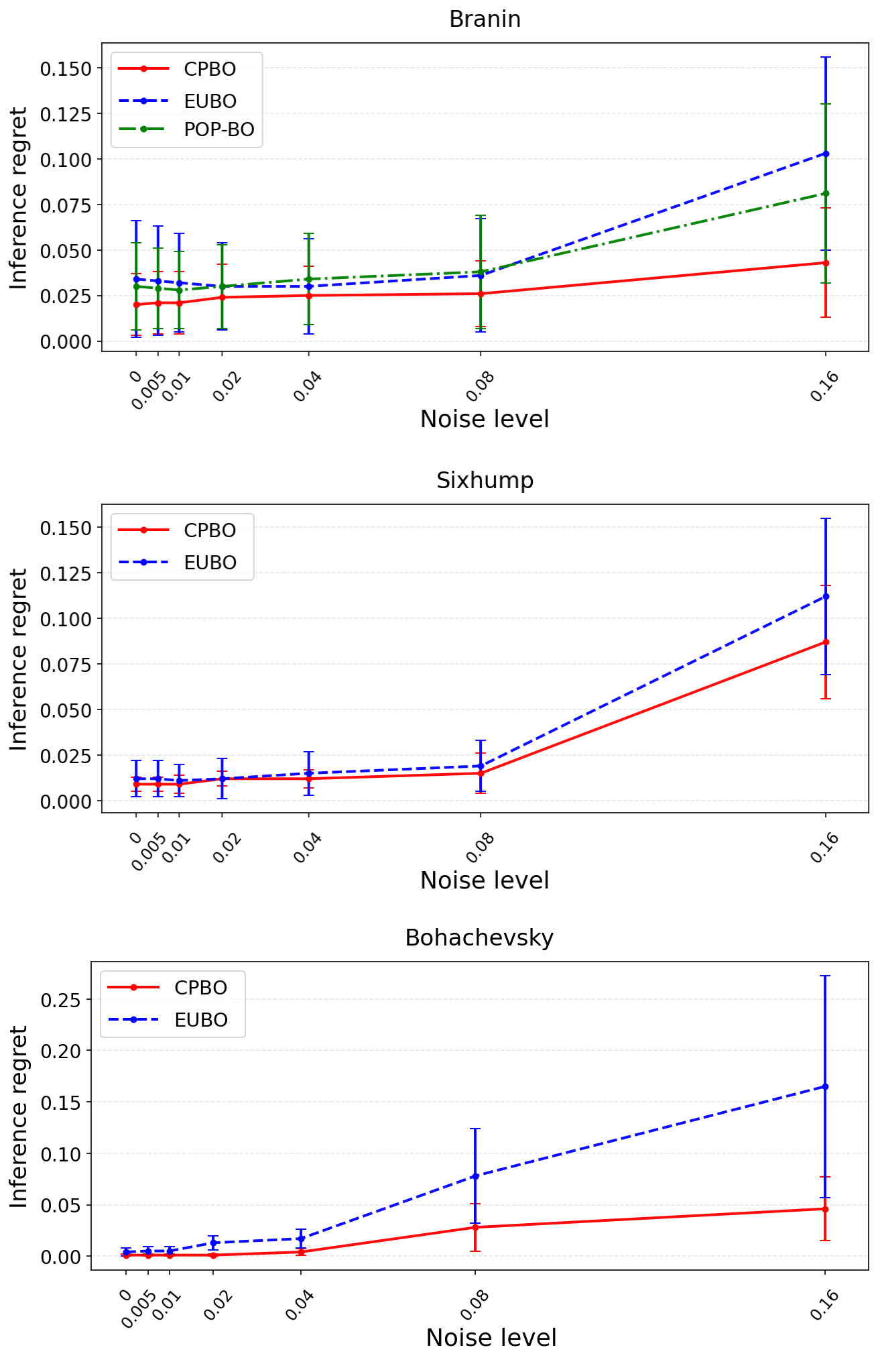}
    \caption{
        Sensitivity to oracle-feedback noise (RUM noise) on the three representative benchmark functions. 
        We vary the RUM noise used to generate pairwise preferences while keeping the rest of the Table~\ref{tab:combined_results} setup fixed.
        The plots report inference regret after 30 iterations for CPBO, EUBO, and, where available, POP-BO.
        Markers show the mean across runs and vertical bars show $\pm$ standard deviation over 30 runs.
    }
    \label{fig:rum_noise_sensitivity}
\end{figure}

We also evaluate robustness to noise in the preference oracle itself. Unlike the sensitivity analysis for the model noise scale $\sigma$, this experiment changes the RUM noise used to generate pairwise feedback while keeping the experimental setup otherwise identical to Table~\ref{tab:combined_results}. Figure~\ref{fig:rum_noise_sensitivity} shows that CPBO$_{\hat{\gamma}}$ is consistently less affected by increasing oracle-feedback noise than the baselines.

We summarize this trend by computing the area under the inference-regret curve across noise levels. On Branin, CPBO has 42.1\% lower AUC than EUBO and 36.6\% lower AUC than POP-BO. On Bohachevsky and Six-hump, CPBO has 69.6\% and 21.3\% lower AUC than EUBO, respectively. These results suggest that modeling indifference through the learned threshold $\hat{\gamma}$ improves robustness when pairwise preferences become increasingly noisy.

\paragraph{Comparison with qEUBO.}
In our consecutive-comparison setup, each iteration evaluates one new candidate against a fixed reference, which effectively corresponds to a batch of size $q=2$ in the original batch formulation. In this regime, qEUBO \citep{astudillo2023qeubo} reduces to a Monte Carlo estimator of the same expected-utility-of-best-outcome (EUBO) objective, for which closed-form expressions are available. Table~\ref{tab:eubo_mc_vs_analytic} provides a direct comparison between qEUBO ($q=2$), the analytic EUBO implementation, and CPBO$_{\hat\gamma}$, mirroring the Table~\ref{tab:combined_results} setup. As shown, both EUBO formulations yield nearly identical results, confirming their equivalence under this regime, while CPBO consistently achieves lower inference regret across all benchmark functions.

For completeness, we further evaluate qEUBO under a Gumbel-noise oracle calibrated to the “moderate logistic’’ level, as described in \citet{astudillo2023qeubo}. This setup mirrors the conditions of Table~\ref{tab:combined_results} but employs a different noise model. As summarized in Table~\ref{tab:gumbel_noise}, the overall conclusions remain unchanged: CPBO$_{\hat\gamma}$ achieves lower inference regret and higher ordinal accuracy than the baseline, under this alternative noise model.

\begin{table}[t]
\centering
\caption{Inference regret after 30 iterations (mean $\pm$ sd over 30 runs) for qEUBO ($q=2$), EUBO, and CPBO$_{\hat\gamma}$, mirroring the format of Table~\ref{tab:combined_results}. Bold indicates the lowest mean regret for each benchmark.}
\vspace{0.5em}
\label{tab:eubo_mc_vs_analytic}
\begin{tabular}{lccc}
\toprule
\textbf{$\utility(\candidate)$} & \textbf{qEUBO} & \textbf{EUBO} & \textbf{CPBO$_{\hat\gamma}$} \\
\midrule
Branin        & .036 $\pm$ {\scriptsize .031} & .034 $\pm$ {\scriptsize .032} & \textbf{.020 $\pm$ {\scriptsize .017}} \\
Six-hump      & .014 $\pm$ {\scriptsize .011} & .012 $\pm$ {\scriptsize .010} & \textbf{.009 $\pm$ {\scriptsize .004}} \\
Bohachevsky   & .004 $\pm$ {\scriptsize .004} & .004 $\pm$ {\scriptsize .004} & \textbf{.001 $\pm$ {\scriptsize .001}} \\
\hdashline
Levy13        & .018 $\pm$ {\scriptsize .013} & .016 $\pm$ {\scriptsize .012} & \textbf{.011 $\pm$ {\scriptsize .016}} \\
Bukin6        & .129 $\pm$ {\scriptsize .081} & .126 $\pm$ {\scriptsize .078} & \textbf{.115 $\pm$ {\scriptsize .053}} \\
Cross-Tray    & .171 $\pm$ {\scriptsize .108} & .177 $\pm$ {\scriptsize .112} & \textbf{.150 $\pm$ {\scriptsize .064}} \\
Ackley        & .151 $\pm$ {\scriptsize .122} & .149 $\pm$ {\scriptsize .118} & \textbf{.093 $\pm$ {\scriptsize .052}} \\
\bottomrule
\end{tabular}
\end{table}

\begin{table}[t]
\centering
\caption{Optimization and exploration performance measured by inference regret (Regret) and global ordinal accuracy of the utility proxy (Ord.) under a Gumbel-noise oracle calibrated to the “moderate logistic’’ level, mirroring the Table~\ref{tab:combined_results} setup but with qEUBO. Values are mean $\pm$ sd over 30 runs, bold denotes the better mean performance for each benchmark and metric, with lower Regret and higher Ord. being preferable.}
\vspace{0.5em}
\setlength{\tabcolsep}{5.7pt}
\renewcommand{\arraystretch}{1.1}
\begin{tabular}{lcccc}
\toprule
\multirow{2}{*}{$\utility(\candidate)$}
  & \multicolumn{2}{c}{\textbf{qEUBO}}
  & \multicolumn{2}{c}{\textbf{CPBO$_{\hat \gamma}$}} \\
\cmidrule(lr){2-3} \cmidrule(lr){4-5}
 & Regret$\downarrow$ & Ord.$\uparrow$ & Regret$\downarrow$ & Ord.$\uparrow$ \\
\midrule
Branin        & $0.044 \pm {\scriptstyle 0.035}$ & 64.4 $\pm$ {\scriptsize 6.9} & $\mathbf{0.023 \pm {\scriptstyle 0.021}}$ & \textbf{85.4 $\pm$ {\scriptsize 3.4}} \\
Six-hump      & $0.023 \pm {\scriptstyle 0.019}$ & 68.9 $\pm$ {\scriptsize 6.6} & $\mathbf{0.010 \pm {\scriptstyle 0.006}}$ & \textbf{80.8 $\pm$ {\scriptsize 3.6}} \\
Bohachevsky   & $0.009 \pm {\scriptstyle 0.007}$ & 75.6 $\pm$ {\scriptsize 7.6} & $\mathbf{0.001 \pm {\scriptstyle 0.001}}$ & \textbf{93.3 $\pm$ {\scriptsize 2.0}} \\
\hdashline
Levy13        & $0.025 \pm {\scriptstyle 0.018}$ & 72.5 $\pm$ {\scriptsize 6.8} & $\mathbf{0.015 \pm {\scriptstyle 0.016}}$ & \textbf{83.0 $\pm$ {\scriptsize 3.9}} \\
Bukin6        & $0.141 \pm {\scriptstyle 0.096}$ & 73.5 $\pm$ {\scriptsize 6.1} & $\mathbf{0.124 \pm {\scriptstyle 0.061}}$ & \textbf{92.8 $\pm$ {\scriptsize 2.1}} \\
Cross-Tray    & $0.175 \pm {\scriptstyle 0.115}$ & 58.4 $\pm$ {\scriptsize 6.8} & $\mathbf{0.154 \pm {\scriptstyle 0.069}}$ & \textbf{64.9 $\pm$ {\scriptsize 7.4}} \\
Ackley        & $0.167 \pm {\scriptstyle 0.128}$ & 69.2 $\pm$ {\scriptsize 8.2} & $\mathbf{0.101 \pm {\scriptstyle 0.057}}$ & \textbf{83.1 $\pm$ {\scriptsize 3.8}} \\
\bottomrule
\end{tabular}
\label{tab:gumbel_noise}
\end{table}


\subsection{Additional Visualizations}

This section provides visual illustrations to help interpret model behavior beyond the quantitative results. We focus on three aspects: (i) the estimated probability of indifference conditioned on a reference point, (ii) how indifference regions vary across utility functions and perceptual thresholds, and (iii) differences in the utility surfaces recovered by CPBO and competing baselines.


\paragraph{Recovered Indifference Probability.}

\begin{figure}[htbp]
    \centering
    \includegraphics[width=0.80\textwidth]{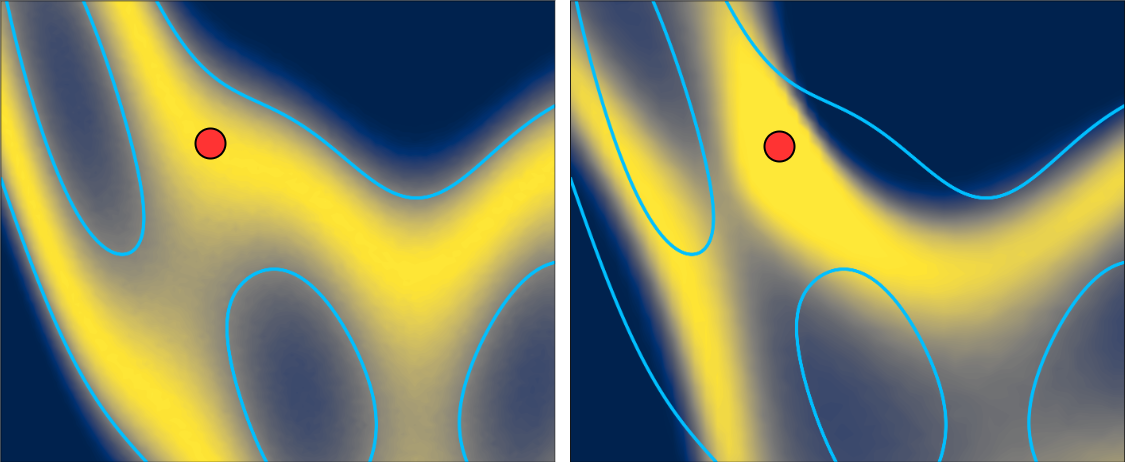}
    \caption{Indifference probability heatmaps conditioned on a candidate (red circle). (Left) Ground-truth probability based on $\gamma_{\text{true}} = 0.1$ and known utility. (Right) Estimated probability from CPBO after 30 iterations. The contour indicates the boundary of JND region.
    }
    \label{fig:recovered_indifference_heatmap}
\end{figure}

Figure~\ref{fig:recovered_indifference_heatmap} illustrates the probability of indifference under a fixed reference point (marked in red). The left image visualizes the ground-truth probability that a new candidate will be judged indifferent to the reference, based on the true utility function with perceptual noise and $\gamma_{\text{true}} = 0.1$. The right image shows the same quantity estimated from a trained CPBO model after 30 iterations. The similarity of the patterns indicates that CPBO successfully captures the shape and extent of the indifference region with limited observations.

\paragraph{Indifference Region.}

Figure~\ref{fig:gamma_band_visualisation} illustrates how indifference regions vary across utility functions, perceptual thresholds, and reference points. Since all utility functions are normalized to the $[0, 1]$ range, the global maximum corresponds to a reference utility of $1.0$, while the mid-level reference is set to $0.5$. These two points illustrate how the JND threshold $\gamma_{\text{true}}$ translates into perceptual indifference depending on both the reference location and the underlying utility landscape. We compare a fixed threshold $\gamma_{\text{true}} = 0.1$ with a $\gamma_{\text{true}}$ yielding $\mathbb{E}[\sim] = 30\%$. For ease of interpretation, we indicate the $\gamma$ band directly, rather than the probability of an indifferent response ($r_t=0$) as shown in Figure~\ref{fig:recovered_indifference_heatmap}; the yellow areas here correspond to regions within that contour.

\newcommand{\rowW}{0.95cm}  
\newcommand{\colW}{2.15cm}  
\setlength{\tabcolsep}{3.5pt}
\renewcommand{\arraystretch}{1.02}
\begin{figure}[htbp]
  \centering
  \begin{tabular}{@{} >{\centering\arraybackslash}m{\rowW}
                  >{\centering\arraybackslash}m{\colW}
                  >{\centering\arraybackslash}m{\colW}
                  >{\centering\arraybackslash}m{\colW}
                  >{\centering\arraybackslash}m{\colW}
                  >{\centering\arraybackslash}m{\colW} @{}}
\toprule
& \multirow{2}{*}{\makebox[\colW][c]{\textbf{Utility surface}}}
& \multicolumn{2}{c}{\textbf{Global maximum}}
& \multicolumn{2}{c}{\textbf{Mid-level}} \\
\cmidrule(lr){3-4}\cmidrule(lr){5-6}
&
& $\gamma_{\text{true}} = 0.1$ &
  $\mathbb{E}[\sim] = 30\%$   &
  $\gamma_{\text{true}} = 0.1$ &
  $\mathbb{E}[\sim] = 30\%$ \\
\midrule

\rname{Branin}
  & \includegraphics[width=\colW]{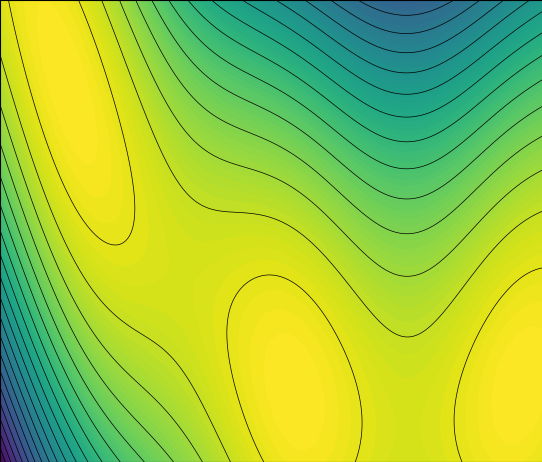}
  & \includegraphics[width=\colW]{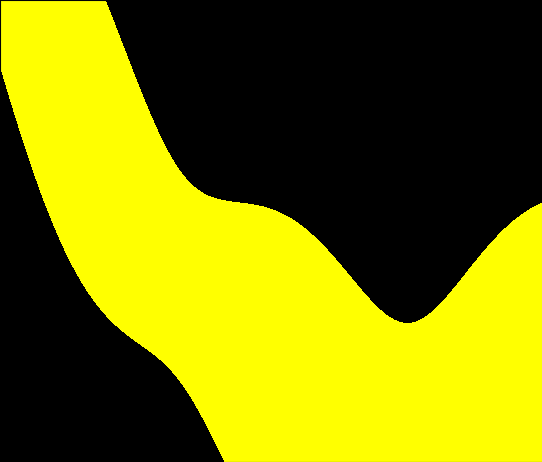}
  & \includegraphics[width=\colW]{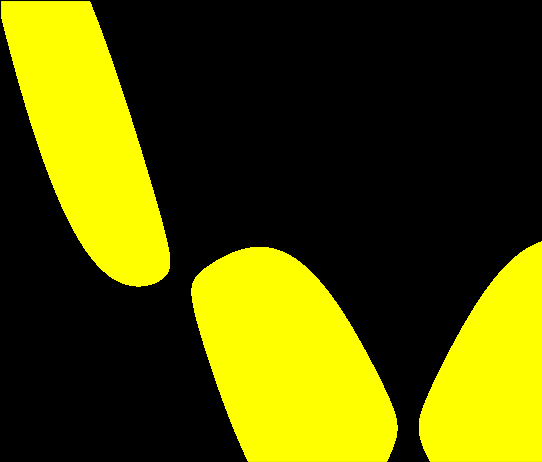}
  & \includegraphics[width=\colW]{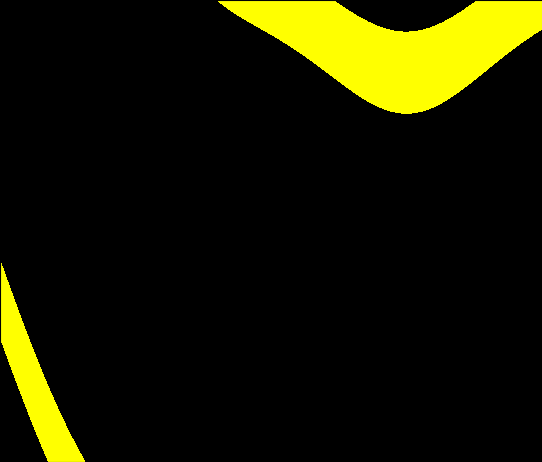}
  & \includegraphics[width=\colW]{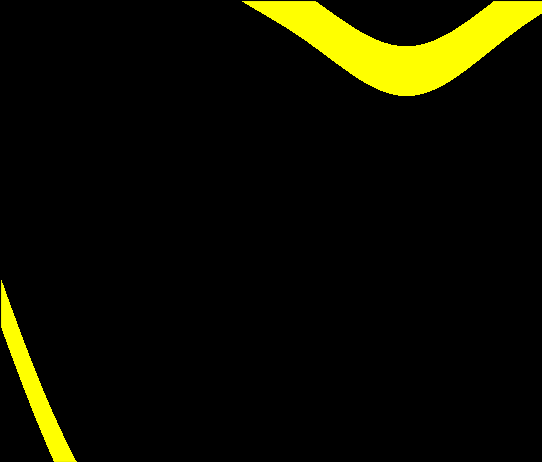} \\

\rname{Six-hump}
  & \includegraphics[width=\colW]{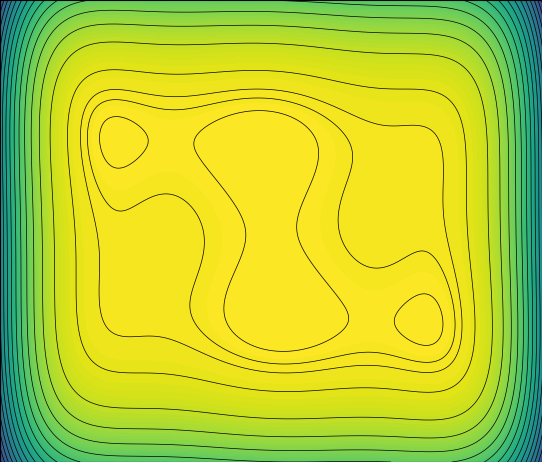}
  & \includegraphics[width=\colW]{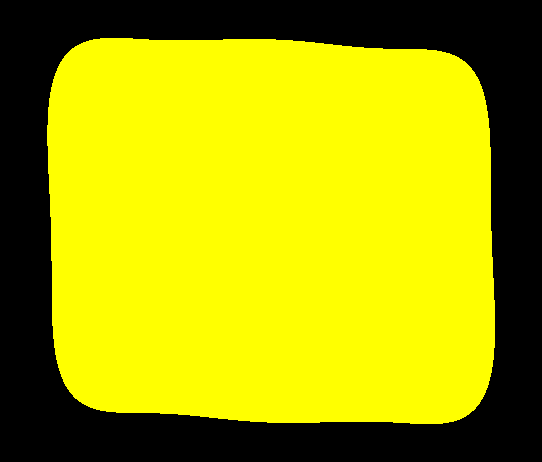}
  & \includegraphics[width=\colW]{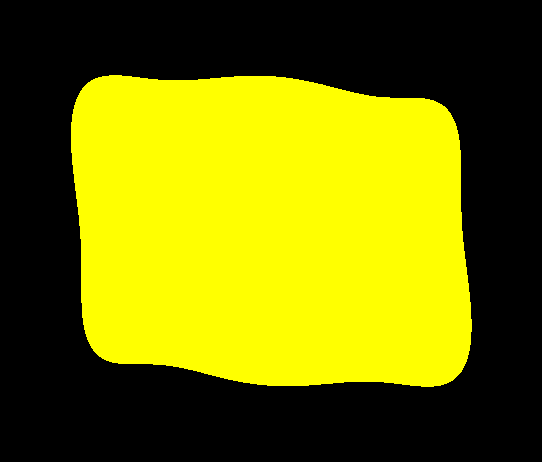}
  & \includegraphics[width=\colW]{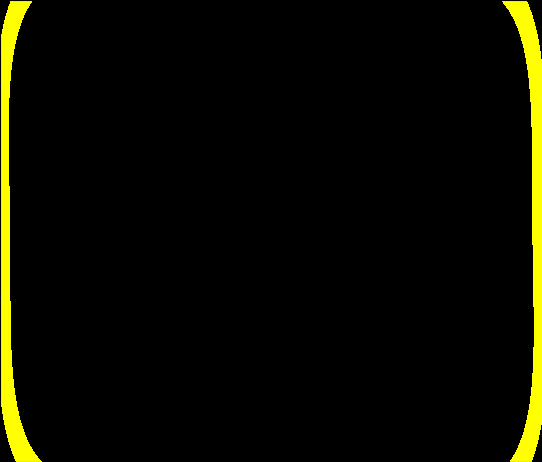}
  & \includegraphics[width=\colW]{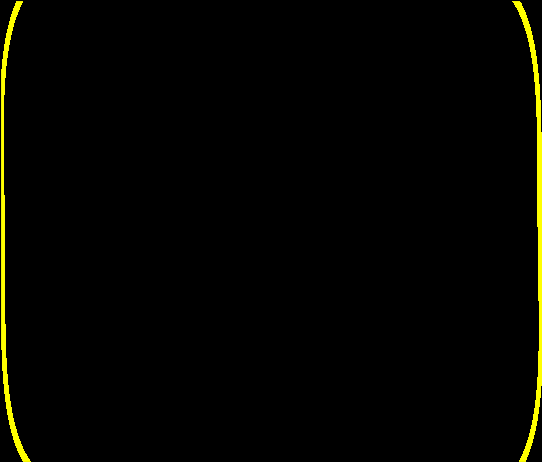} \\

\rname{Bohachevsky}
  & \includegraphics[width=\colW]{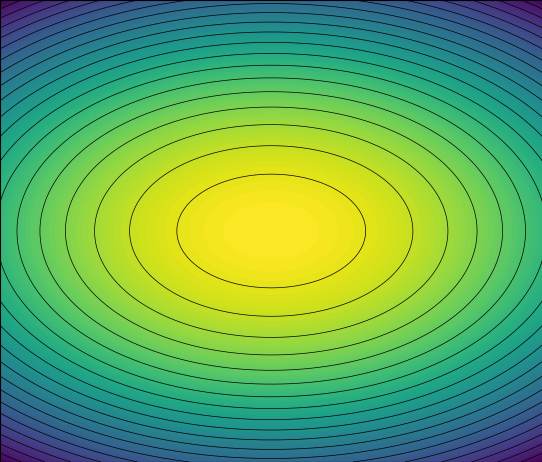}
  & \includegraphics[width=\colW]{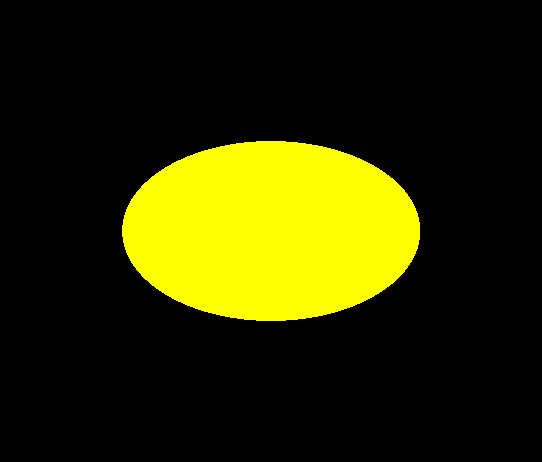}
  & \includegraphics[width=\colW]{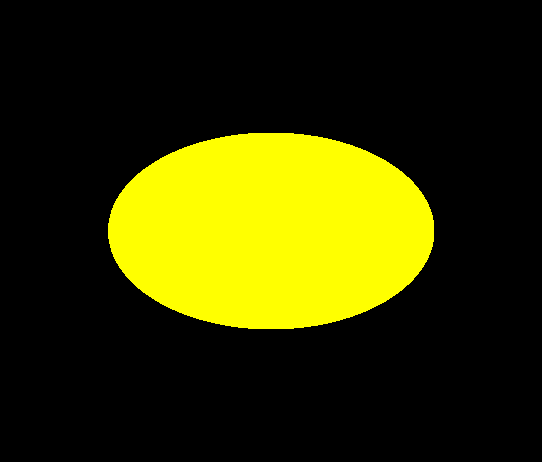}
  & \includegraphics[width=\colW]{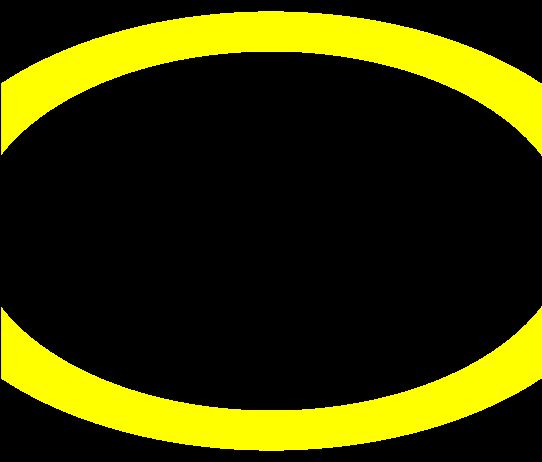}
  & \includegraphics[width=\colW]{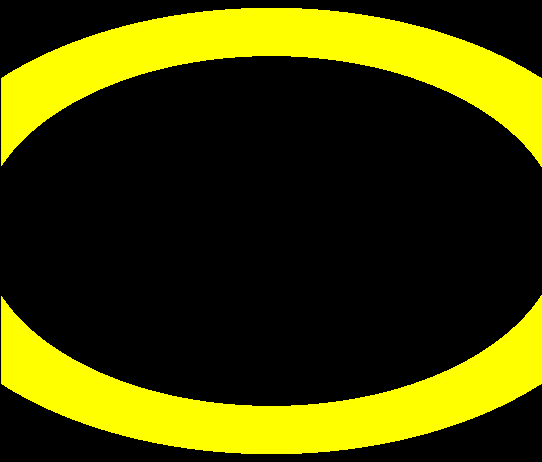} \\

\hdashline
\addlinespace[4pt]

\rname{Levy13}
  & \includegraphics[width=\colW]{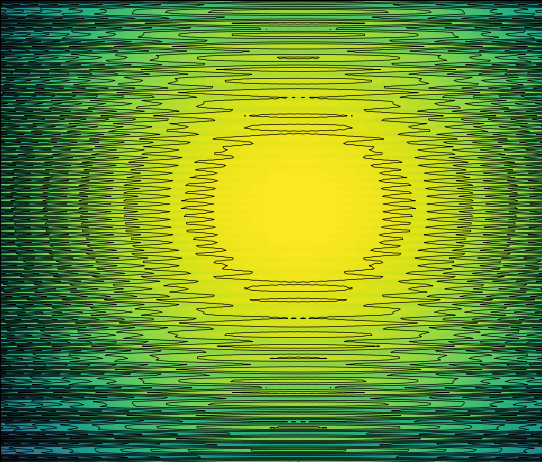}
  & \includegraphics[width=\colW]{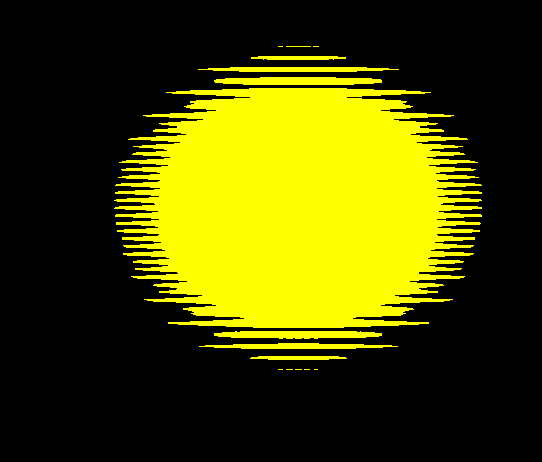}
  & \includegraphics[width=\colW]{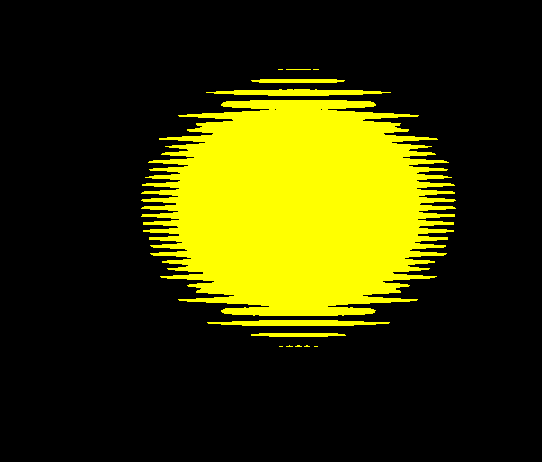}
  & \includegraphics[width=\colW]{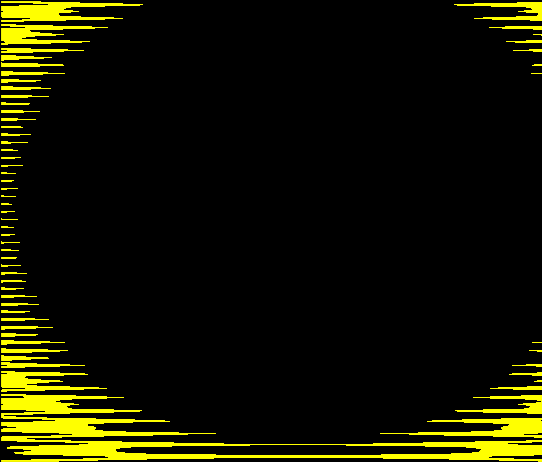}
  & \includegraphics[width=\colW]{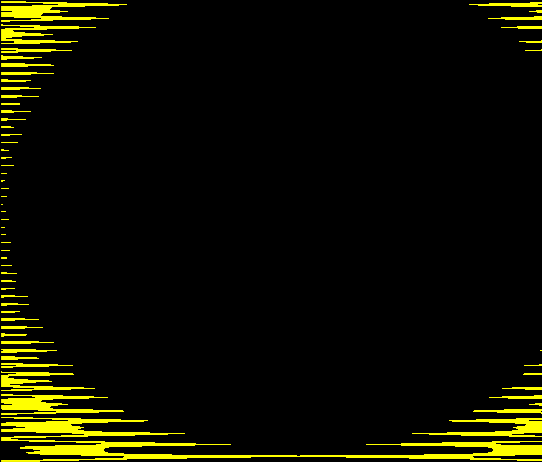} \\

\rname{Bukin6}
  & \includegraphics[width=\colW]{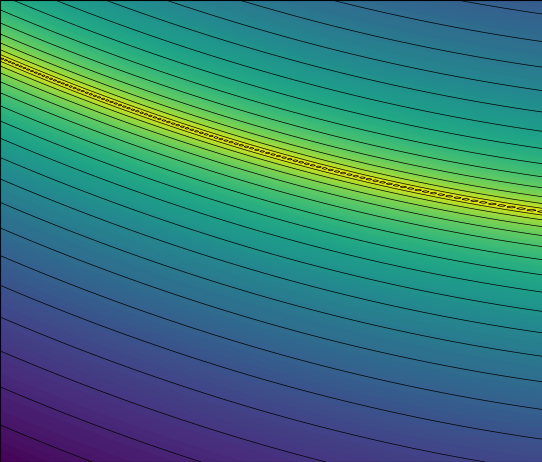}
  & \includegraphics[width=\colW]{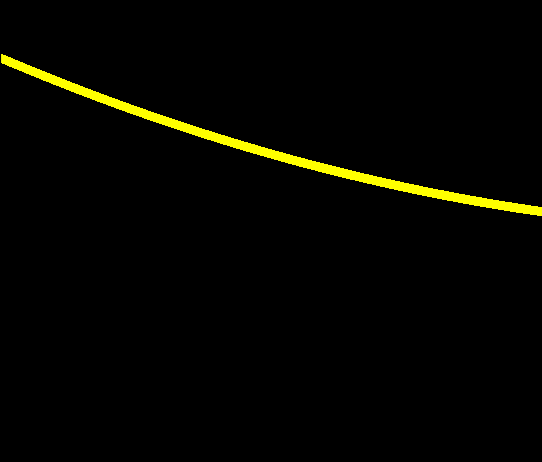}
  & \includegraphics[width=\colW]{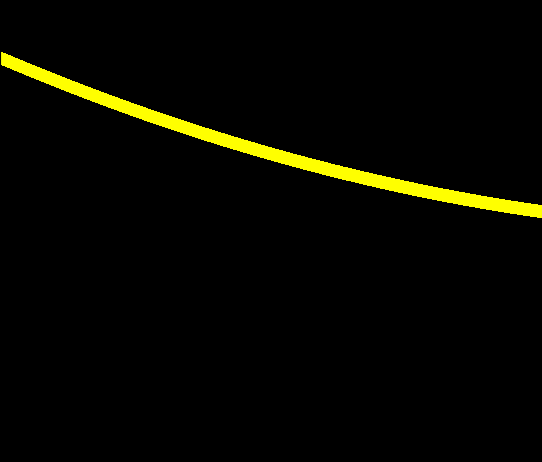}
  & \includegraphics[width=\colW]{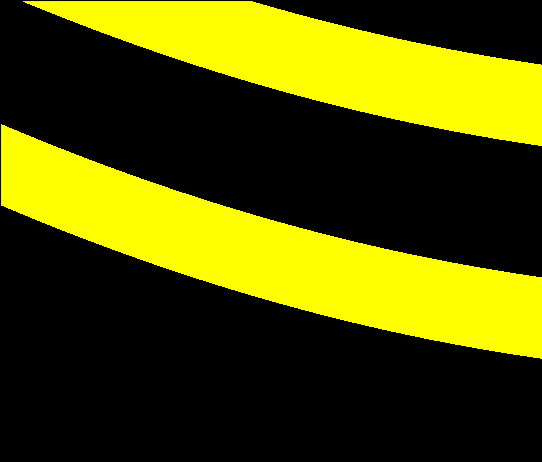}
  & \includegraphics[width=\colW]{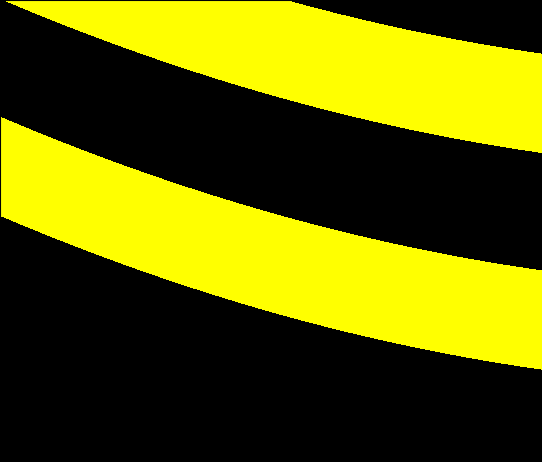} \\

\rname{Cross-Tray}
  & \includegraphics[width=\colW]{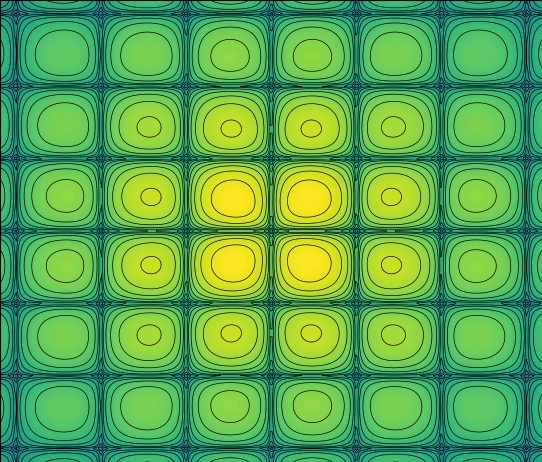}
  & \includegraphics[width=\colW]{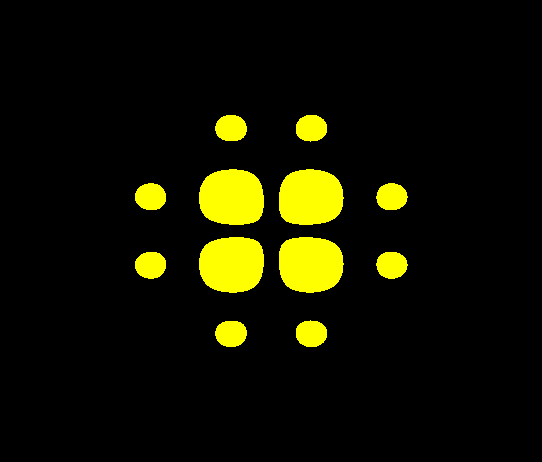}
  & \includegraphics[width=\colW]{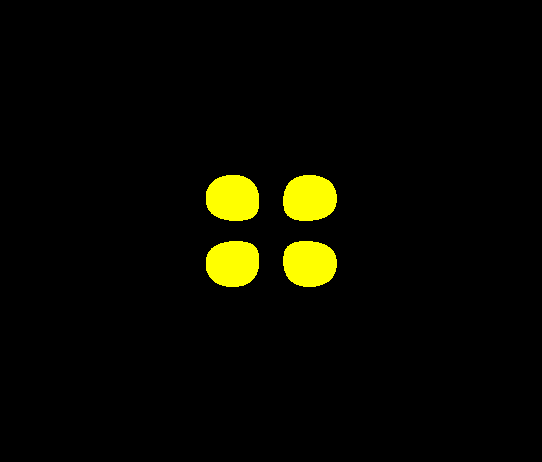}
  & \includegraphics[width=\colW]{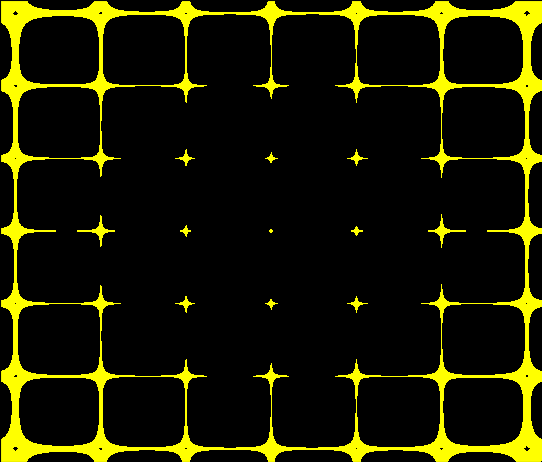}
  & \includegraphics[width=\colW]{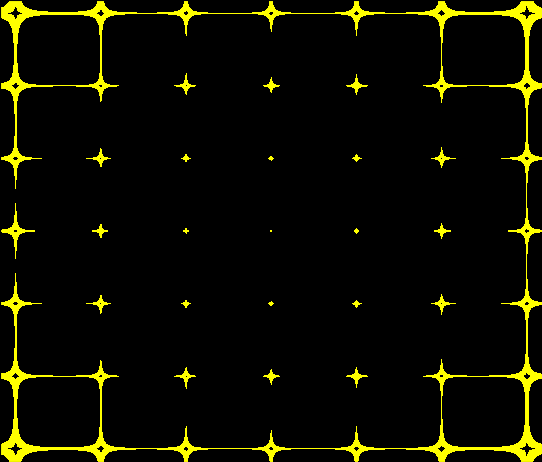} \\

\rname{Ackley}
  & \includegraphics[width=\colW]{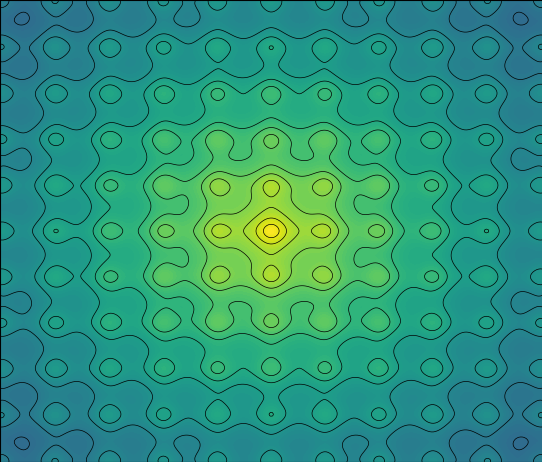}
  & \includegraphics[width=\colW]{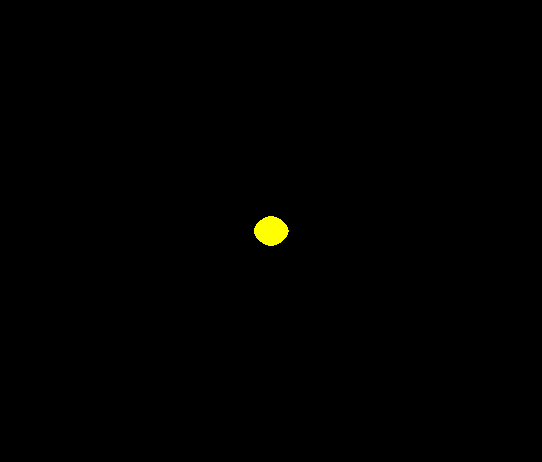}
  & \includegraphics[width=\colW]{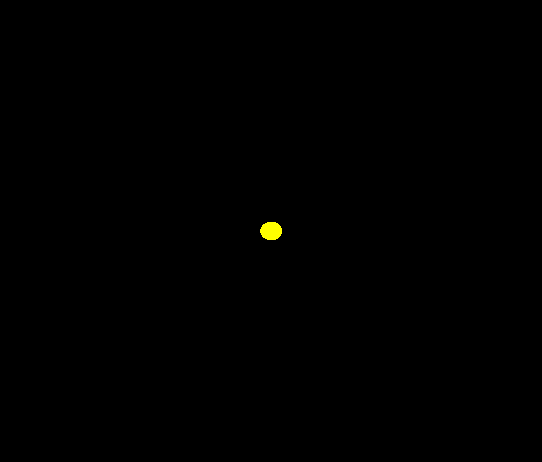}
  & \includegraphics[width=\colW]{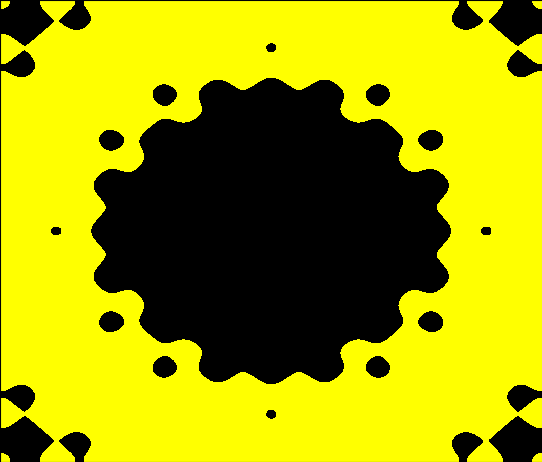}
  & \includegraphics[width=\colW]{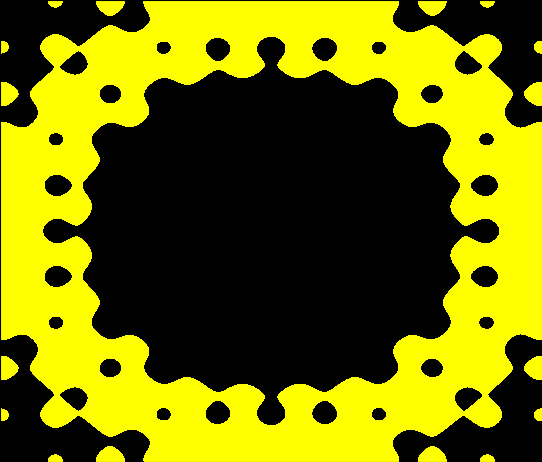} \\
\bottomrule
\end{tabular}
\caption{Indifference regions for seven benchmark utility functions. Each column shows areas within $\gamma_{\text{true}}$ of a fixed reference point, using either a fixed threshold ($\gamma_{\text{true}} = 0.1$) or a threshold chosen to yield $\mathbb{E}[\sim] = 30\%$. Reference points correspond to the global maximum ($1.0$) and the mid-level utility ($0.5$) on the normalized $[0, 1]$ surface. The dashed line separates smoother, preferentially relevant functions (top) from non-smooth, spiky functions (bottom).}
\label{fig:gamma_band_visualisation}
\end{figure}


\paragraph{Utility Surface.}

\begin{figure}[htbp]
  \centering
  \begin{tabular}{@{} >{\centering\arraybackslash}p{0.1cm} 
                     c  c  c  c  c @{}}
    \toprule
      & \textbf{Utility surface}
      & \textbf{CPBO$_{\hat\gamma}$}
      & \textbf{CPBO$_{\hat\gamma=0}$}
      & \textbf{EUBO}
      & \textbf{POP-BO} \\

    \rotlabel{Branin}
      & \includegraphics[width=0.15\textwidth]{figures/surfaces/Branin_utility_2d.png}
      & \includegraphics[width=0.15\textwidth]{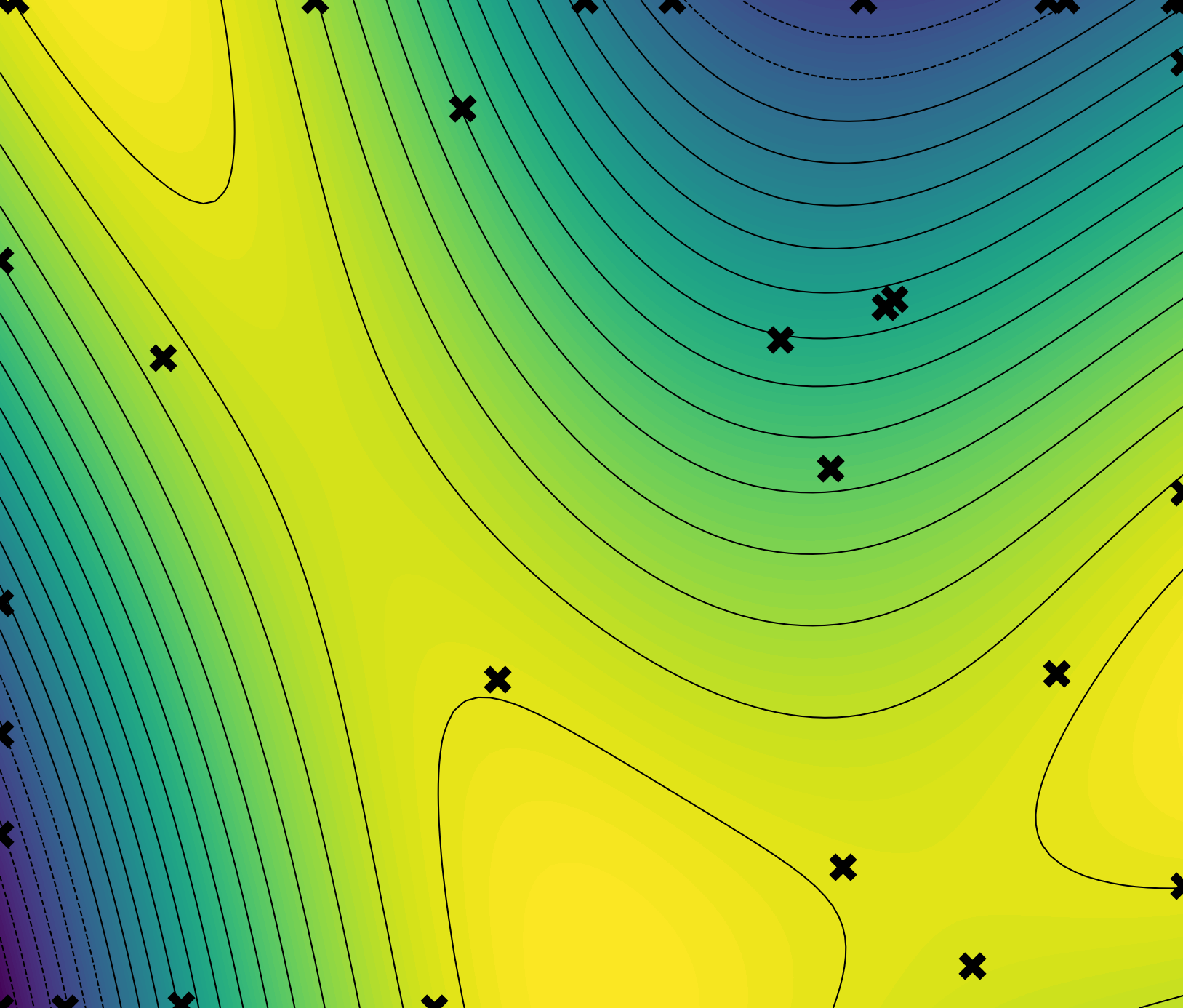}
      & \includegraphics[width=0.15\textwidth]{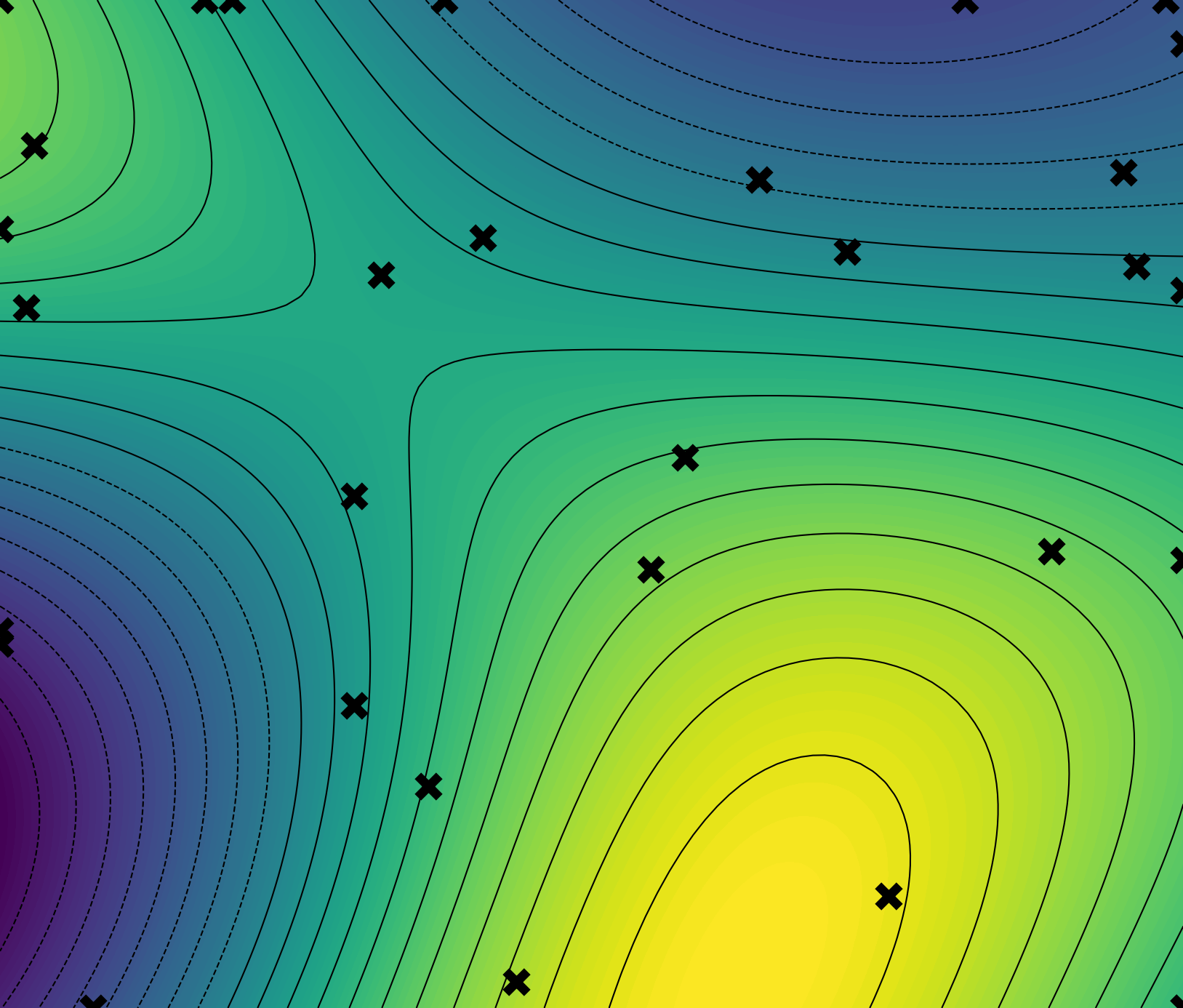}
      & \includegraphics[width=0.15\textwidth]{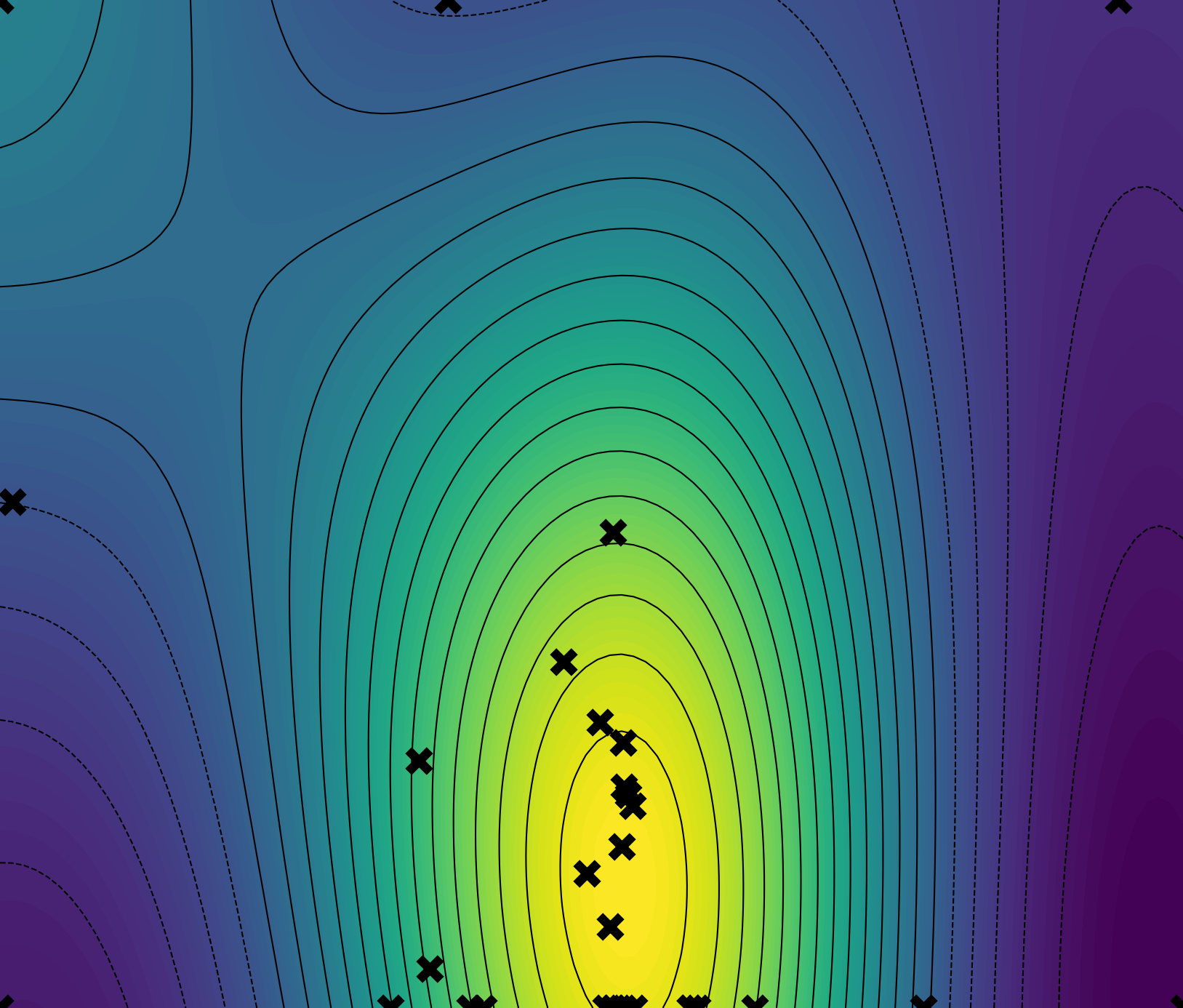}
      & \includegraphics[width=0.15\textwidth]{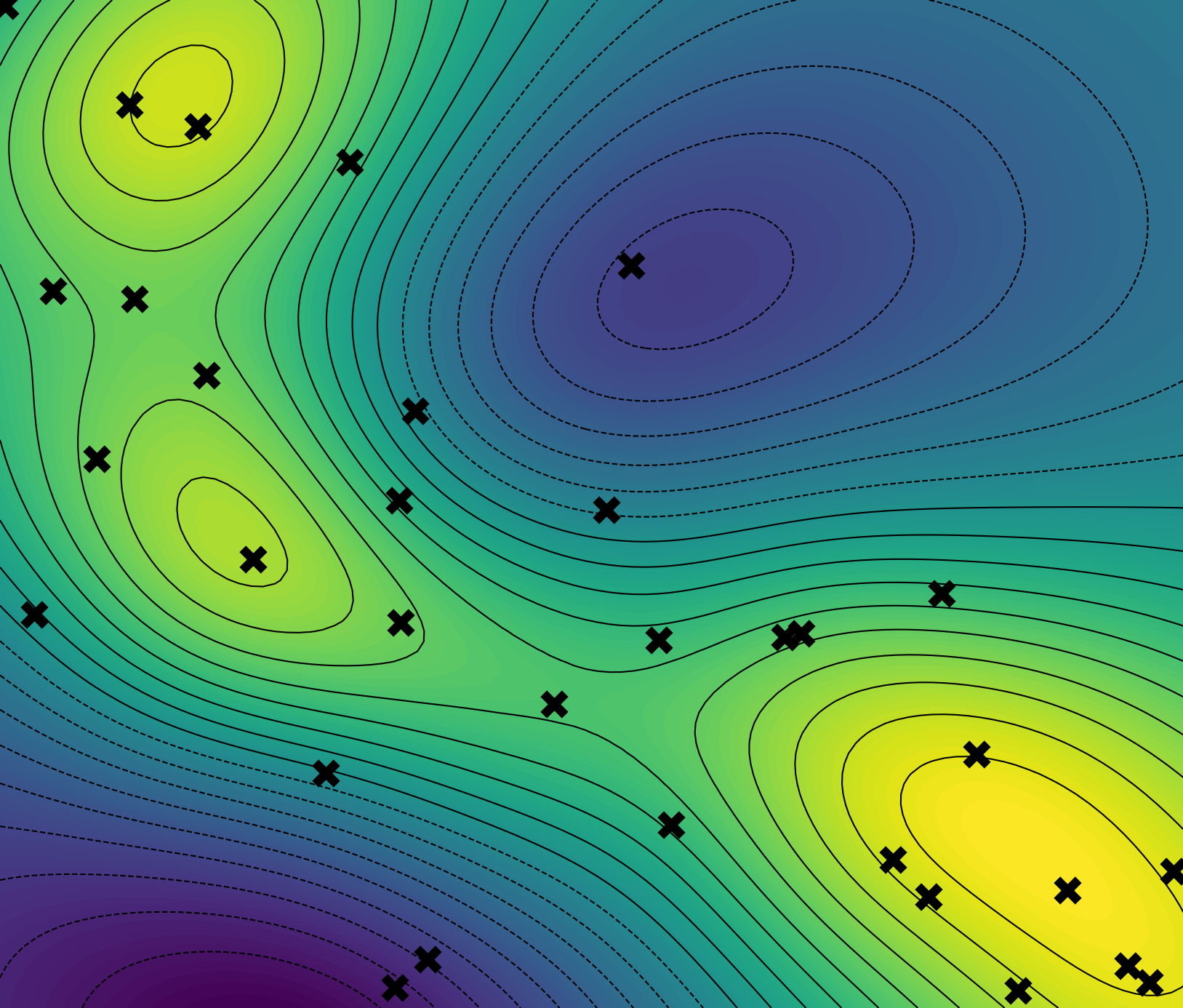} \\

    \rotlabel{Bukin6}
      & \includegraphics[width=0.15\textwidth]{figures/surfaces/Bukin_utility_2d.png}
      & \includegraphics[width=0.15\textwidth]{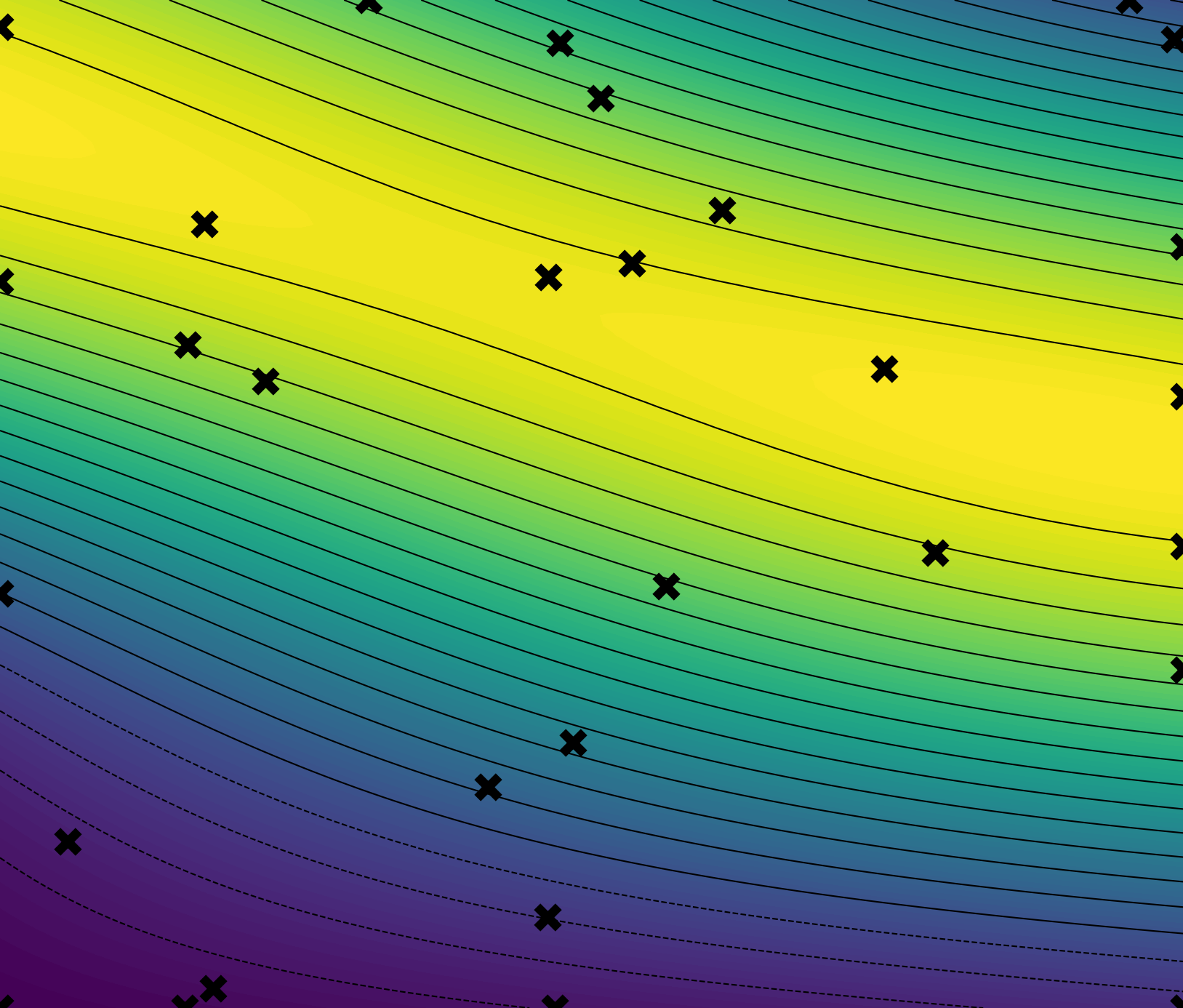}
      & \includegraphics[width=0.15\textwidth]{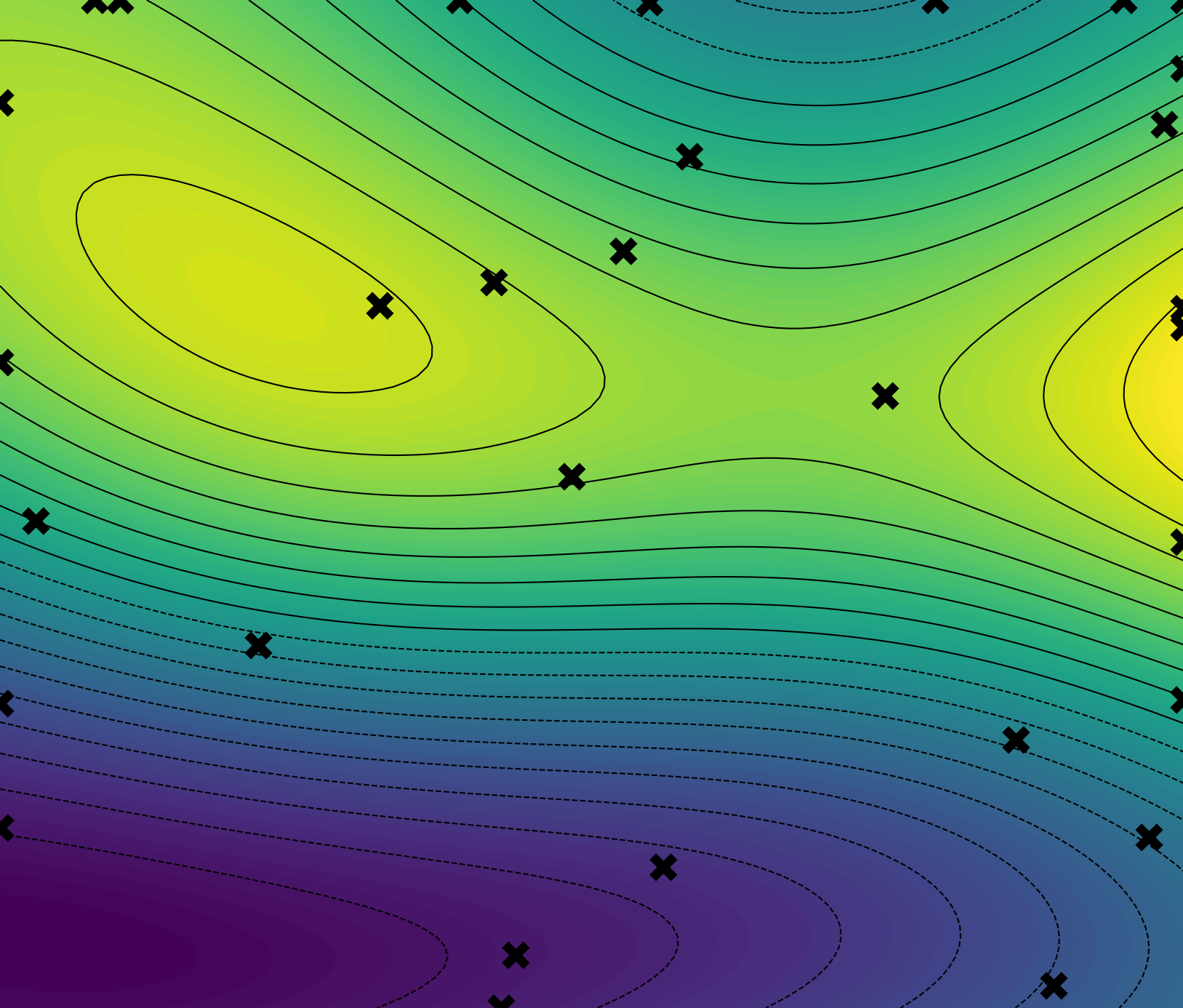}
      & \includegraphics[width=0.15\textwidth]{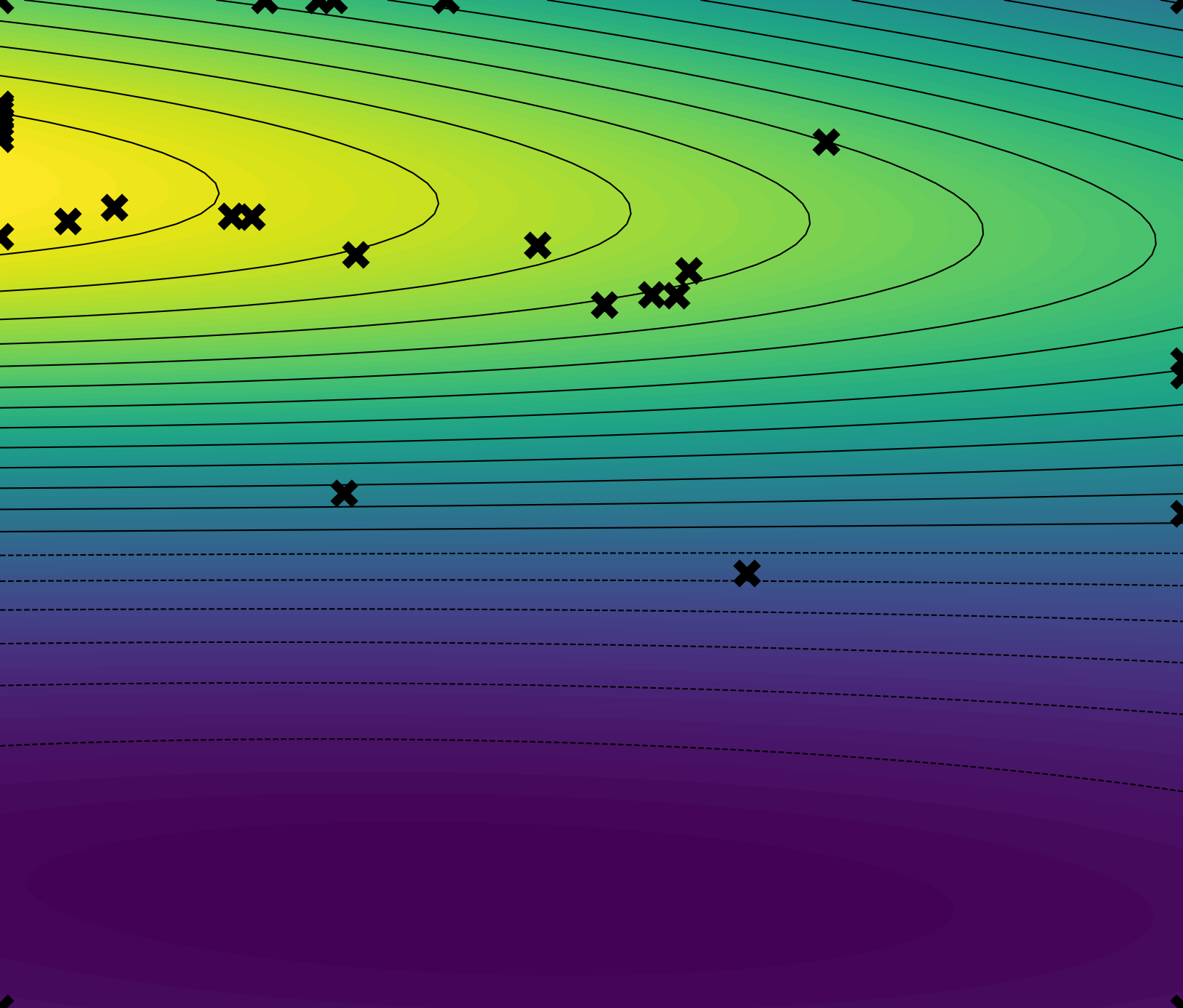}
      & \includegraphics[width=0.15\textwidth]{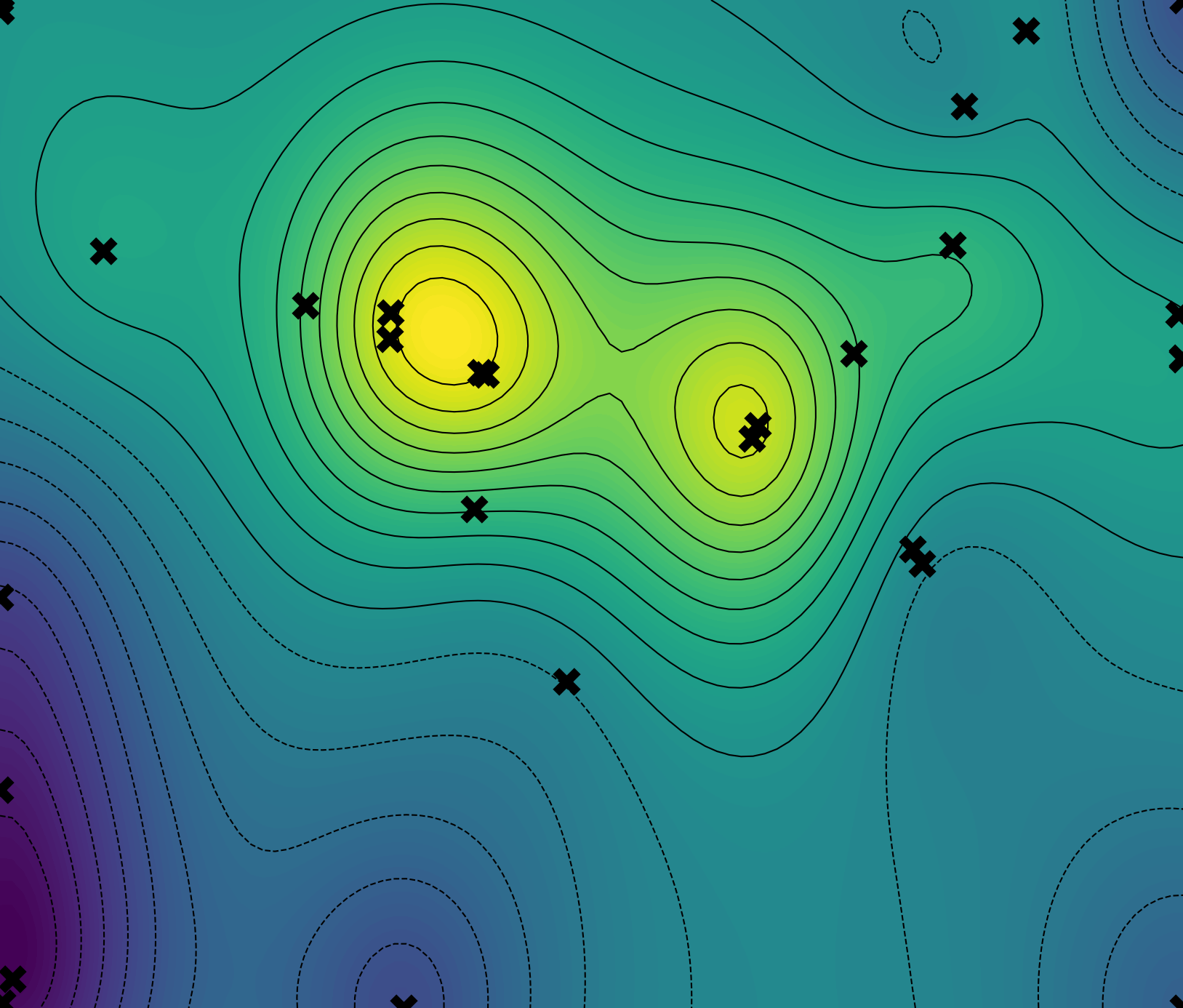} \\

    \rotlabel{Levy13}
      & \includegraphics[width=0.15\textwidth]{figures/surfaces/Levy13_utility_2d.png}
      & \includegraphics[width=0.15\textwidth]{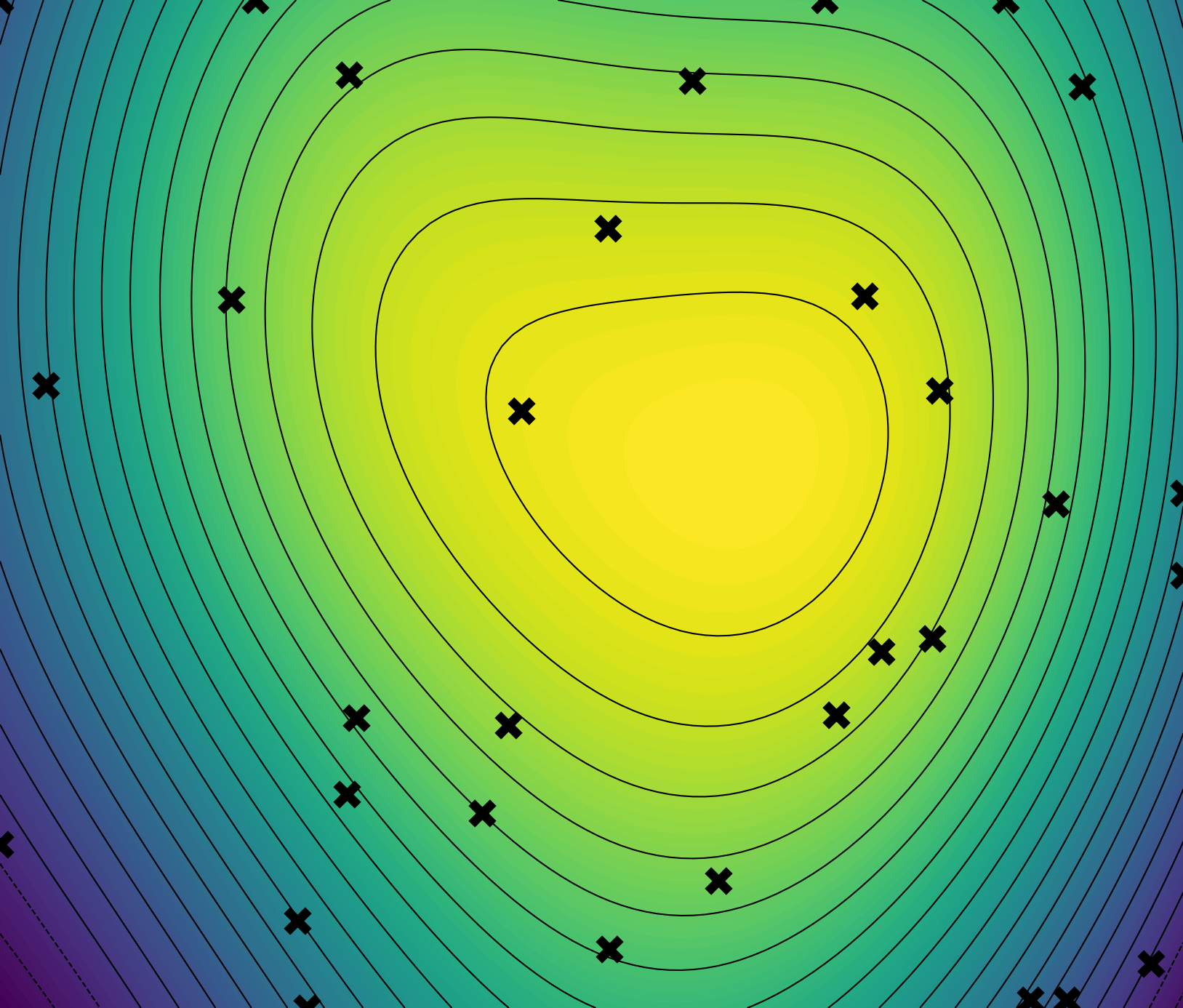}
      & \includegraphics[width=0.15\textwidth]{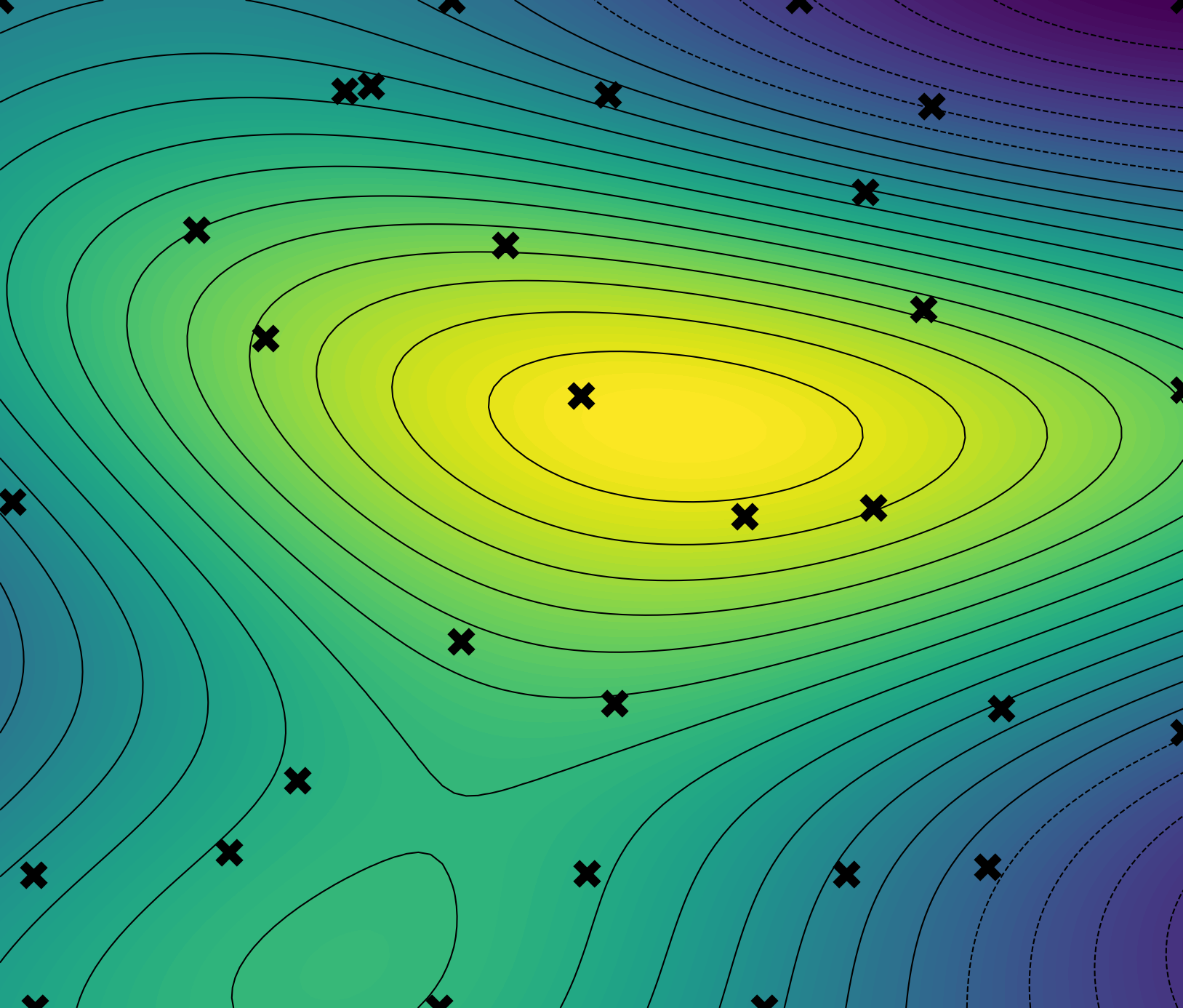}
      & \includegraphics[width=0.15\textwidth]{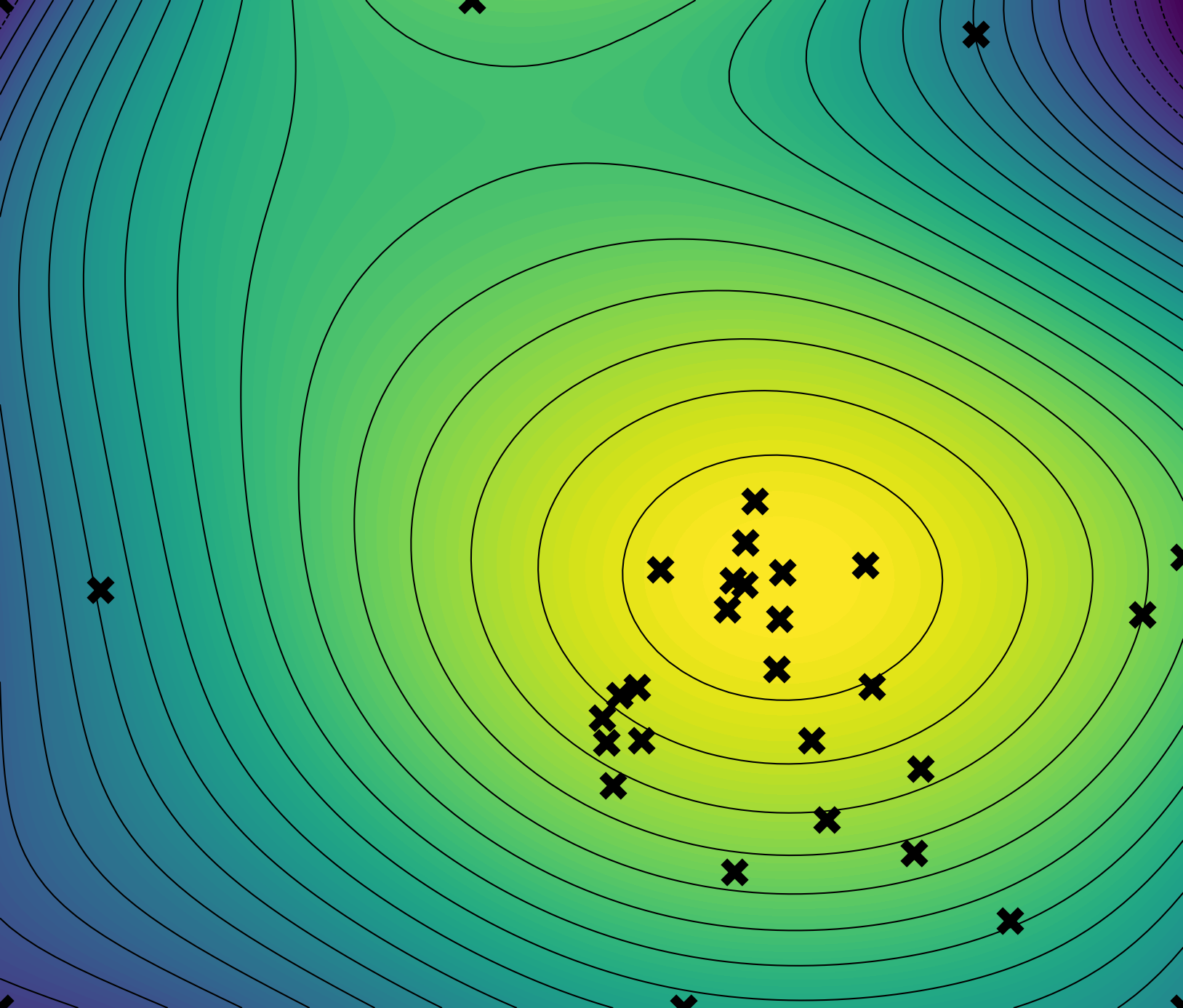}
      & \includegraphics[width=0.15\textwidth]{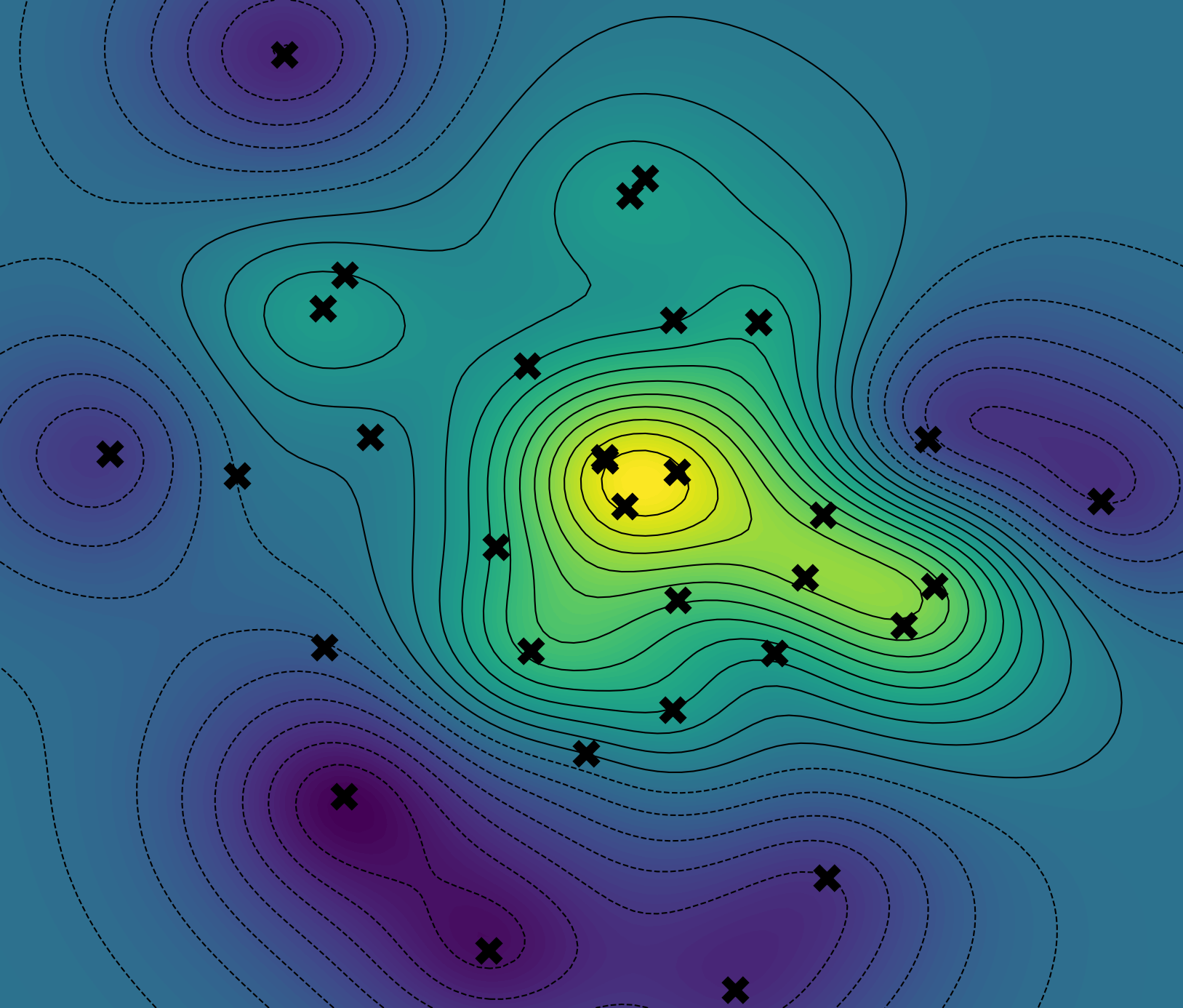} \\
    \bottomrule
  \end{tabular}
  \caption{Latent surface examples for the three benchmark utilities under $\gamma_{\text{true}}=0.1$ and 30 iteration training, comparing CPBO with estimated $\hat\gamma$, CPBO at $\hat\gamma=0$, EUBO, and POP-BO.}
  \label{fig:latent_surface_comparison}
\end{figure}

Figure~\ref{fig:latent_surface_comparison} shows example latent utility surfaces after 30 iterations on three benchmarks. While results may vary between runs, CPBO$_{\hat\gamma}$ generally provides more accurate reconstructions of the true utility compared to the baselines. The adapted MES acquisition promotes broader exploration, and incorporating a learnable JND threshold improves reconstruction in regions with flat or ambiguous preferences.

\section{EXTRUSION OPTIMIZATION}
\label{app:extrusion_optimization}

This section describes the real-world optimization study used to evaluate CPBO in a practical setting. The task involves optimizing a physical process using human preferences as feedback. We provide an overview of the experimental setup, including operator instructions, the optimization task, and supporting tools for real-time interaction.

\subsection{Overview}

In this experiment, three qualified operators with varying levels of experience were tasked with solving a high-moisture extrusion optimization challenge using recommendations generated by the CPBO algorithm. A technical facilitator (one of the authors) assisted the process by recording verbal preference responses $\response_t$ from each operator, inputting them into the algorithm, and communicating the recommended configuration. Recommendations were typically displayed on a screen, and operators recorded them on their individual tracking sheets. A fourth operator was responsible for the physical preparation of the 15 initial configurations (see Section~\ref{app:vtt_experiment_details} for details), which were then evaluated by the decision-making operators before proceeding to CPBO runs.

\subsection{Operator Instructions}

All participating operators were employed at the institute where the experiment was conducted and were already familiar with extrusion processing. They varied in their experience with optimization tasks and use of the specific extruder, ranging from novice/intermediate to advanced and expert levels. The advanced and expert operators operated the extruder independently. In contrast, the novice operator, who lacked hands-on experience with this particular extruder model, received technical support from an independent operator responsible for the initial configurations, but made all operational decisions independently.

For this study, operators received a shared set of task-specific instructions to ensure consistency. These included general guidance on how to interact with the CPBO system, evaluate samples, and provide preference feedback. The individual steps and instructions are paraphrased below, in a condensed format:
\begin{itemize}
    \item The operator should compare each newly produced sample to the one produced in the previous step.
    \item After each production, the operator packages the produced sample, putting it into a zip-lock bag, for future evaluations.
    \item The preferences are to be based on the overall quality of the extrudate, considering characteristics such as fibril length, toughness, and tearing force. These characteristics are commonly used in the field and the operators are familiar with their meaning. A high-quality extrudate is defined as one that exhibits a fibrous structure resembling meat-like texture, but beyond this the operators were judged to provide their subjective evaluation on the overall quality.
    \item If the operator cannot distinguish a difference or considers both samples equally good, they report them as “the same”, corresponding to the indifference option in CPBO.
    \item If the recommended configuration is expected to cause an extruder blockage or produce clearly flawed material, the operator stops the run and marks the new candidate as not preferred.
    \item Preferences are communicated verbally to the facilitator, who records and enters them into the optimization system.
    \item After each acquisition run, the newly recommended configuration is displayed on a screen for the operator to input into the extruder.
\end{itemize}

\subsection{Optimization Task}

The experiment was conducted using a twin-screw extruder (Process 11 Hygienic, screw diameter 11 mm, L/D 40:1) equipped with a long cooling die designed specifically for high-moisture extrusion of fibrous meat-analogue products. The goal was to identify optimal values for three parameters: temperature, screw speed, and water feed, jointly denoted as $\configuration \in \mathbb{R}^3$, that yield a high-quality extrudate $\candidate(\configuration)$.  

The task represents a subcomponent of a general R\&D workflow, where the aim is to identify a sufficiently good configuration to proceed to downstream analysis and product evaluation. For each feed mixture candidate, we first need to find the optimal configuration before it makes sense to continue further.

In this study, the feed mixture consisted of 50\% pea protein isolate and 50\% hemp protein concentrate, and as explained above a high-quality extrudate is defined as one that exhibits a fibrous structure resembling meat-like texture.

\subsection{Experiment Details}
\label{app:vtt_experiment_details}
The study comprised one initial session and three operator-specific optimization sessions. In the initial phase, an independent operator produced $N_0 = 15$ initial extrudates by executing candidate configurations, selected via Latin hypercube sampling \citep{mckay1979lhs} over the 3D parameter space of temperature, screw speed, and water feed. Each configuration $\configuration_i$ was used to produce a physical sample $\candidate(\configuration_i)$.

All initial samples were frozen immediately after production. During the operator sessions, each operator received three 5 cm segments of each sample, which were reheated in sealed plastic cups at 40 °C for 10 minutes and then left at room temperature for 20 minutes before the evaluation.

Each operator evaluated the initial samples through pairwise comparisons, generating preference data $ 
\mathcal{D}_t = \left\{ \left( \left[ \configuration_{i-1}, \configuration_i \right], \response_i \right) \right\}_{i=2}^{t}
$. These responses were used to fit a surrogate model ${\utility}_0(\configuration)$, serving as the starting point for the CPBO algorithm.

During the experiment, the surrogate model ${\utility}_t(\configuration)$ was iteratively updated from the operator's feedback. At each step $t$, the operator produced the candidate $\candidate(\configuration_t)$ using the configuration $\configuration_t$, and compared it to the previous candidate. The process continued for 30 total steps (15 initial + 15 interactive).

At steps $t = 24$ and $t = 29$, the current surrogate model's maximizer $\hat{\configuration}^* = \argmax_{\configuration} {\utility}_t(\configuration)$ was selected for production, resulting in the 25th and 30th candidates, respectively. Unlike the other candidates, which were selected based on the acquisition function to maximize information gain, these were explicitly recommended by the model as the best configurations. Those configurations were retained for potential reproduction in sensory panel testing and further qualitative assessment.

\begin{figure}[tbp]
    \centering
    \includegraphics[width=1.00\textwidth]{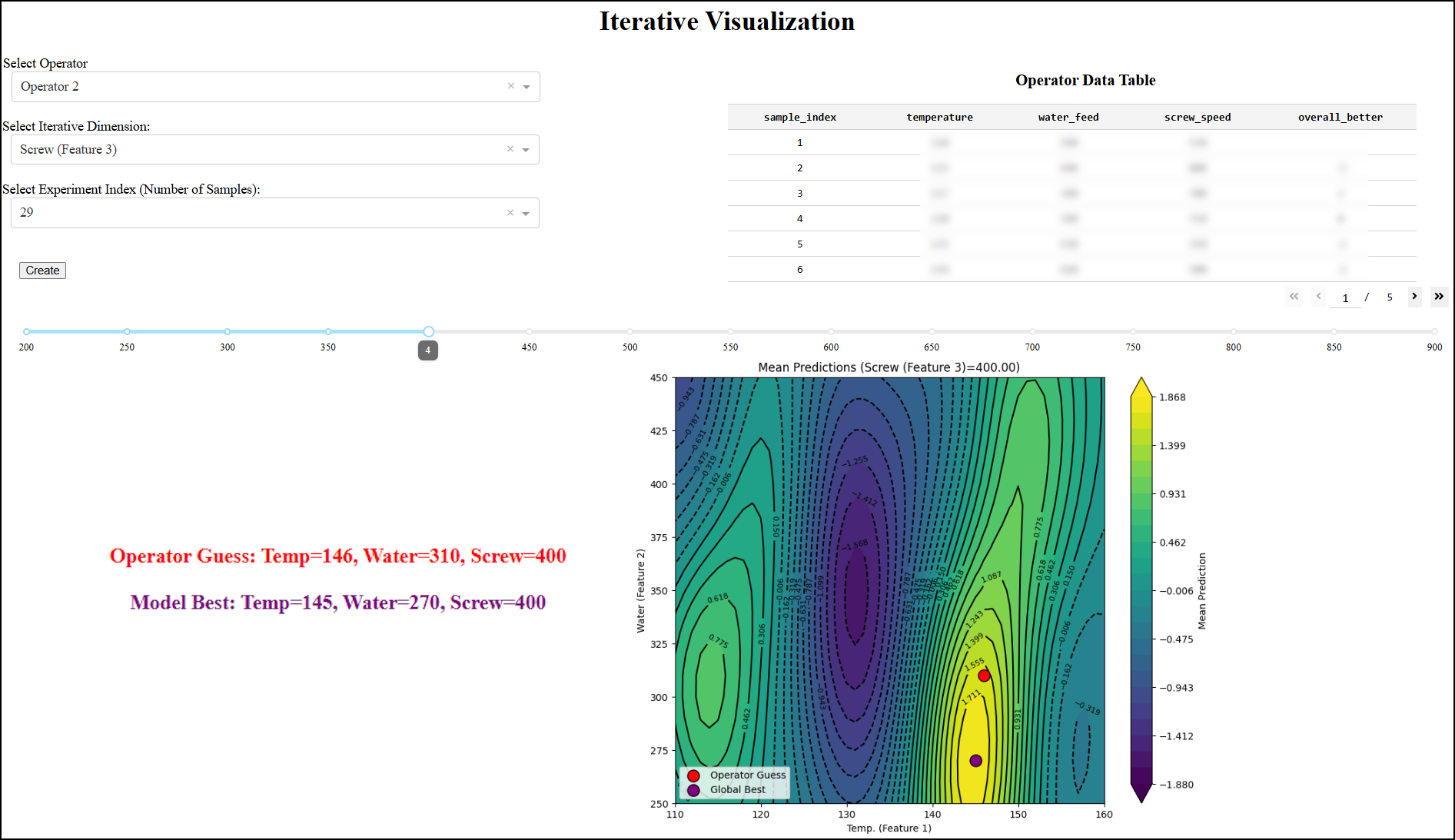}
    \caption{Web-based GUI developed with Plotly Dash for monitoring the optimization process. Tabular data is blurred for privacy.
    }
    \label{fig:vtt_iterative_visualization}
\end{figure}

Figure~\ref{fig:vtt_iterative_visualization} shows a web-based interactive visualization tool developed using Plotly Dash. The interface allows tracking the model's predicted optimum over time and visualizing how the estimated utility evolves as more feedback is collected, providing insights into operator preferences.

\subsection{Technical Details}
\label{app:vtt_technical_details}
The real-world extrusion optimization followed the same GP surrogate and acquisition configurations used in the synthetic benchmark experiments, with one exception: we found it beneficial to impose a prior over the GP kernel lengthscales. Specifically, we placed independent Gamma priors $\ell_d \sim \mathrm{Gamma}(1.0, 0.05)$ on each lengthscale dimension. 

\paragraph{Fixed Parameters.}
A constant flour feed of 300 g/h was maintained throughout the experiment. The temperature of the cooling die was set to 45 °C. The first four adjustable barrel zones were fixed at 60, 70, 80, and 100 °C, respectively.

\paragraph{Optimized Parameters.}
Three parameters were optimized during the experiment:
\begin{itemize}
    \item \textbf{Barrel temperature} in the last four extruder zones: 110–160 °C, with 1 °C resolution
    \item \textbf{Water feed:} 250-450 ml/h, with 10 ml/h resolution
    \item \textbf{Screw speed:}: 200-900 rpm, with 50 rpm resolution
\end{itemize}
These ranges were selected based on prior empirical knowledge of stable process windows and known safe operating boundaries for the twin-screw extruder and the feed mixture. The temperature limits were set to enable the production of reasonable extrudates while avoiding a high risk of extruder blockage. The water feed limits ensured that the resulting moisture content was appropriate, given the fixed flour feed of 300 g/h. The chosen screw speed range corresponded to the practical operating range of the extruder.

\subsection{Evaluation}

After the experiment, the operators were informally interviewed to assess their satisfaction with the final recommendation produced by CPBO. They were also given the opportunity to continue the experiment by testing additional configurations of their own choosing, allowing them to explore whether they could identify a better solution through manual iteration. In addition, the operators were asked to comment on the fixed parameters and the selected boundaries for the optimized variables, to confirm their practical relevance for the task. Although we do not report individual responses, the interviews served to verify that the recommendations were satisfactory and that the experimental setup aligned with the operators' expectations. The raw iteration-level data from the extrusion study are subject to confidentiality restrictions and therefore cannot be released. A controlled repetition of the extrusion study with additional optimization algorithms was not practically feasible within the scope of this work. Each experimental campaign requires substantial machine time, raw materials, operator involvement, cleaning, and repeated sensory evaluation. We therefore treat the extrusion study as a demonstration of the practical feasibility of CPBO under real operational constraints rather than as a controlled comparison with alternative optimization methods.

\subsection{Results}

\begin{figure}[tbp]
    \centering
    \includegraphics[width=0.90\textwidth]{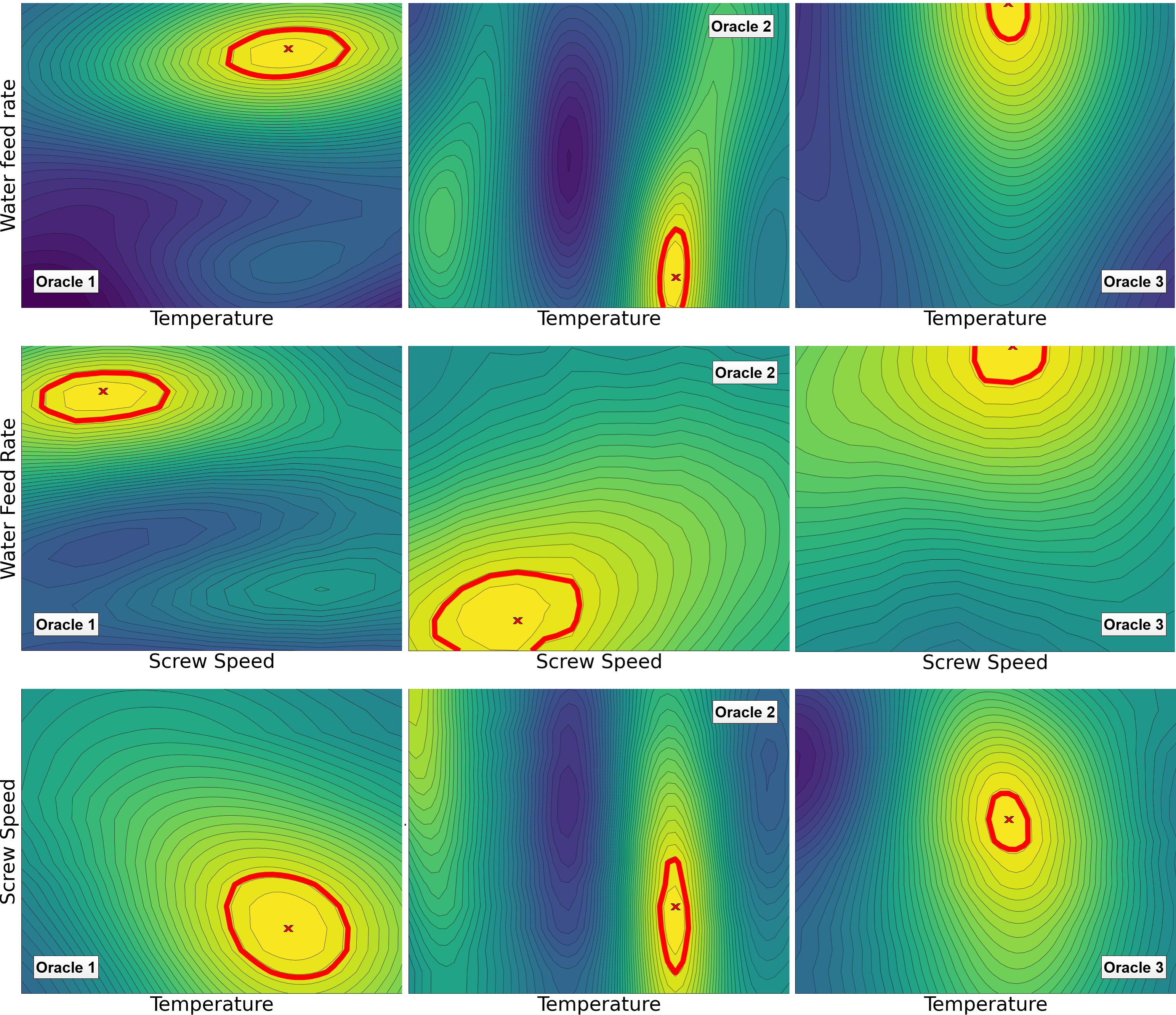}
    \caption{2D slices of the learned utility surfaces for the three human operators, with the third parameter fixed to its estimated optimal value. Red ‘x’ marks the estimated optimum, and the bold red contour shows the learned JND region conditioned on the estimated optimum.
    }
    \label{fig:vtt_oracle_surfaces}
\end{figure}

The three human operators demonstrated notably different preferences, resulting in three distinct final recommendations. Figure~\ref{fig:vtt_oracle_surfaces} illustrates 2D slices of the utility surface, always fixing one setting to its estimated optimal value and visualizing the estimated utility over the remaining two. Operator 1 preferred a balanced setting across temperature and water feed rate. Operator 2 found two distinct configurations acceptable, indicating flexibility in the desired product. In contrast, operator 3 strongly favored high water content with a specific temperature range. Despite reaching different solutions, all operators were satisfied with the final extrudate, highlighting the subjective nature of the task. Furthermore, the three operators indicated that the fixed parameters and the selected parameter boundaries were appropriate and consistent with their domain knowledge and expectations for stable operation.

The plot also shows the learned threshold $\gamma$ indicating the configurations that the model considers likely to produce candidates indistinguishable from the estimated optimal one for this operator. This information helps in validating sensitivity of the optimal solution.

\end{document}